\documentclass[10pt]{article} 
\usepackage[preprint]{tmlr}

\usepackage{amsmath,amsfonts,bm}

\def\eqref#1{equation~\ref{#1}}

\def\1{\bm{1}}

\DeclareMathAlphabet{\mathsfit}{\encodingdefault}{\sfdefault}{m}{sl}
\SetMathAlphabet{\mathsfit}{bold}{\encodingdefault}{\sfdefault}{bx}{n}

\usepackage{float}
\usepackage{graphicx}
\usepackage[hidelinks]{hyperref}
\usepackage{url}
\usepackage{cleveref}
\usepackage[utf8]{inputenc}
\usepackage[T1]{fontenc}
\usepackage{amsmath}
\usepackage{textcomp}
\usepackage{array}
\usepackage[numbers]{natbib}
\usepackage[edges]{forest}
\usepackage{microtype}
\usepackage{xcolor}
\usepackage{titlesec}
\useforestlibrary{edges}

\newcolumntype{P}[1]{>{\centering\arraybackslash}p{#1}}

\title{Generative AI for Character Animation: A Comprehensive Survey of Techniques, Applications, and Future Directions}

\author{%
\begin{center}
Mohammad Mahdi Abootorabi{\normalfont\textsuperscript{\textdagger}},
Omid Ghahroodi{\normalfont\textsuperscript{\textdagger}},  
Pardis Sadat Zahraei{\normalfont\textsuperscript{\textdagger}}, \\
Hossein Behzadasl{\normalfont\textsuperscript{$\diamond$}},
Alireza Mirrokni{\normalfont\textsuperscript{$\diamond$}},
Mobina Salimipanah{\normalfont\textsuperscript{$\diamond$}},
Arash Rasouli{\normalfont\textsuperscript{$\diamond$}}, \\
Bahar Behzadipour{\normalfont\textsuperscript{$\diamond$}},
Sara Azarnoush{\normalfont\textsuperscript{$\diamond$}},
Benyamin Maleki{\normalfont\textsuperscript{$\diamond$}}, Erfan Sadraiye{\normalfont\textsuperscript{$\diamond$}},   
Kiarash Kiani Feriz{\normalfont\textsuperscript{$\diamond$}}, Mahdi Teymouri Nahad{\normalfont\textsuperscript{$\diamond$}}, Ali Moghadasi{\normalfont\textsuperscript{$\diamond$}},
Abolfazl Eshagh Abianeh{\normalfont\textsuperscript{$\diamond$}}, 
Nizi Nazar{\normalfont\textsuperscript{\textdagger}},\\
Hamid R. Rabiee{\normalfont\textsuperscript{$\diamond$}},
Mahdieh Soleymani Baghshah{\normalfont\textsuperscript{$\diamond$}},
Meisam Ahmadi{\normalfont\textsuperscript{$\S$}},
Ehsaneddin Asgari{\normalfont\textsuperscript{\textdagger}}
\\[5pt]
    \addr {\normalfont\textsuperscript{$\diamond$}}Computer Engineering Department, Sharif University of Technology, Tehran, Iran\\
    \addr {\normalfont\textsuperscript{$\S$}}Iran University of Science and Technology\\
    \addr{\normalfont\textsuperscript{\textdagger}} Qatar Computing Research Institute, Doha, Qatar\\[6pt]
  \small{
    \textbf{Correspondence:}
    \href{mailto:easgari@hbku.edu.qa}{easgari@hbku.edu.qa}
  }
\\[4pt]
  \normalfont\href{https://github.com/llm-lab-org/Generative-AI-for-Character-Animation-Survey}{\textcolor{blue}{https://github.com/llm-lab-org/Generative-AI-for-Character-Animation-Survey}}
\end{center}
}

\def\month{MM}  
\def\year{YYYY} 
\def\openreview{\url{https://openreview.net/forum?id=XXXX}} 

\titlespacing*{\paragraph}{0pt}{0.5ex plus 0.1ex minus 0.2ex}{0.5em}

\begin{document}

\maketitle

\begin{abstract}
Generative AI is reshaping art, gaming, and most notably animation. Recent breakthroughs in foundation and diffusion models have reduced the time and cost of producing animated content. Characters are central animation components, involving motion, emotions, gestures, and facial expressions. The pace and breadth of advances in recent months make it difficult to maintain a coherent view of the field, motivating the need for an integrative review. Unlike earlier overviews that treat avatars, gestures, or facial animation in isolation, this survey offers a single, comprehensive perspective on all the main generative AI applications for character animation. We begin by examining the state-of-the-art in facial animation, expression rendering, image synthesis, avatar creation, gesture modeling, motion synthesis, object generation, and texture synthesis. We highlight leading research, practical deployments, commonly used datasets, and emerging trends for each area. To support newcomers, we also provide a comprehensive background section that introduces foundational models and evaluation metrics, equipping readers with the knowledge needed to enter the field.
We discuss open challenges and map future research directions, providing a roadmap to advance AI-driven character-animation technologies. This survey is intended as a resource for researchers and developers entering the field of generative AI animation or adjacent fields.  Resources are available at: \\
\href{https://github.com/llm-lab-org/Generative-AI-for-Character-Animation-Survey}{\textcolor{blue}{https://github.com/llm-lab-org/Generative-AI-for-Character-Animation-Survey}}\textcolor{black}.
\end{abstract}

\section{Introduction}
Generative artificial intelligence (AI) has rapidly progressed in recent years, revolutionizing fields such as computer vision, natural language processing, and human-computer interaction. A significant frontier in this evolution is generating human-related content, including realistic facial synthesis, expressive gestures, and complex motion sequences. These advancements have profound implications for virtual avatars, gaming, animation, assistive technologies, and beyond. At the heart of this innovation is character animation, which combines intricate visual, temporal, and multimodal elements to achieve lifelike representations. This paper comprehensively surveys generative AI techniques for character animation, unifying developments across traditionally fragmented domains. We examine the critical components of this field, starting with fundamental topics such as SMPL \cite{SMPL}, which underpin the training and evaluation of generative systems. 

We then delve into model architectures that have transformed generative AI, from convolutional neural networks (CNNs) and temporal convolutional networks (TCNs) \cite{tcn_main_paper} to state-of-the-art techniques like Generative Adversarial Networks (GANs) \cite{goodfellow2014generative}, Variational Autoencoders (VAEs) \cite{kingma2022autoencodingvariationalbayes}, Transformers \cite{vaswani2023attentionneed}, and Denoising Diffusion Probabilistic Models (DDPMs) \cite{ho-denoising}. Each architecture contributes unique capabilities for addressing challenges in animation, such as generating realistic motion dynamics, modeling multimodal dependencies, and capturing long-range temporal correlations. The discussion extends to cutting-edge methods for integrating generative AI into multimodal systems. Technologies like CLIP \cite{DBLP:conf/icml/RadfordKHRGASAM21} and ControlNet \cite{zhang2023adding} demonstrate how text, image, and audio inputs can be harmonized to produce cohesive and context-aware character animations. Similarly, 3D modeling advancements, including Neural Radiance Fields (NeRFs) \cite{NERF} and 3D Gaussian Splatting \cite{3DGS}, offer novel approaches to representing and synthesizing 3D content for immersive applications.

Additionally, this survey explores and categorizes evaluation metrics that gauge the performance of these models. We focus on metrics that measure computational efficiency, perceptual realism, and the contextual appropriateness of generated animations. These metrics ensure that models meet technical and user-centered standards crucial for real-world deployment. Each subsequent section delves into recent advances in a key aspect of character animation.

\begin{figure*}[t]
    \centering
    \includegraphics[width=\textwidth]{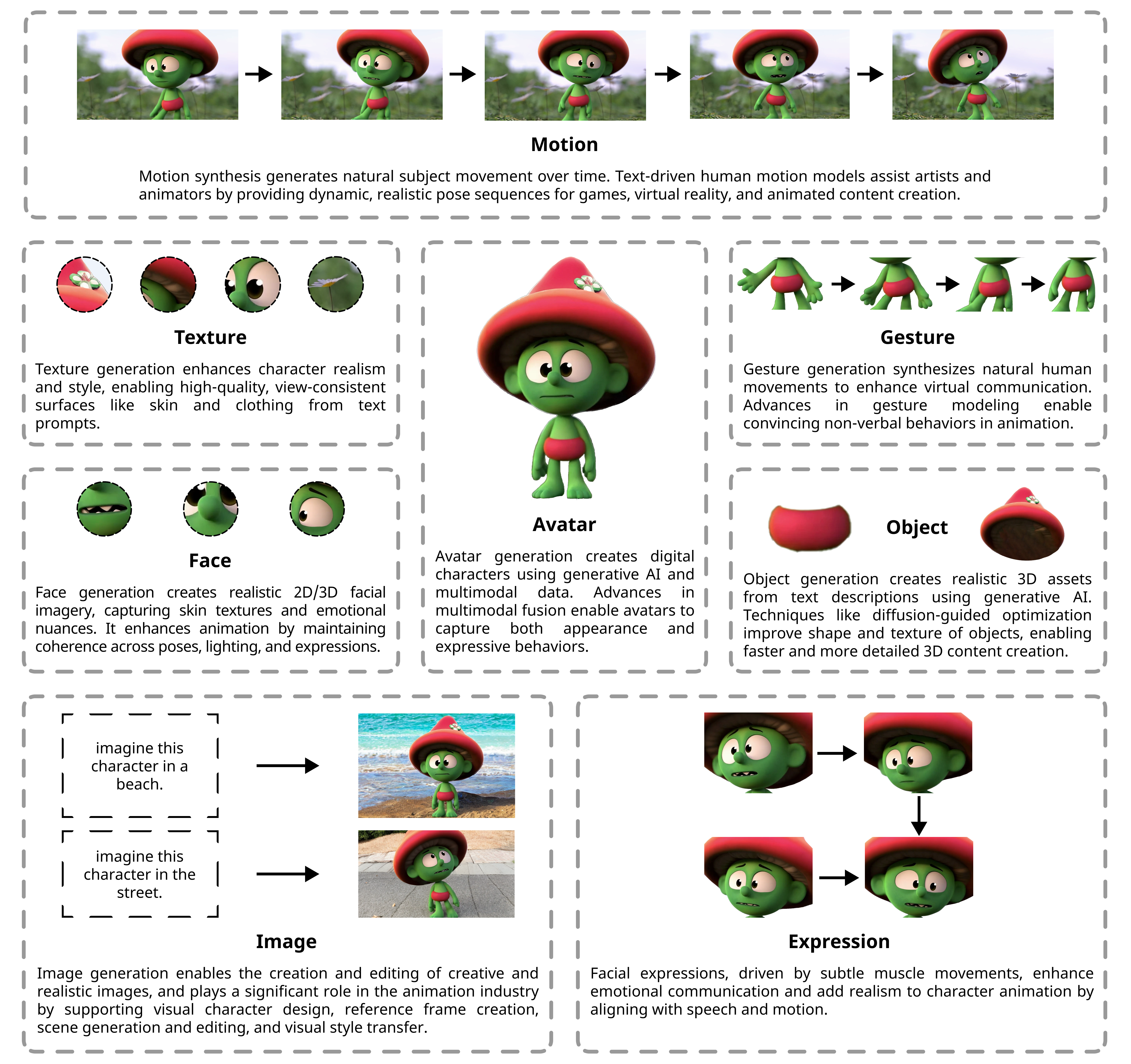}
    \caption{
    Overview of different components in animated character generation. Each aspect, including face, expression, image, avatar, gesture, motion, object, and texture, enhances realism and expressiveness within digital animation environments. Generative AI techniques, such as transformer-based and diffusion-based models, contribute to these components by improving quality, streamlining content creation, and enabling more sophisticated character animation. 
    Generative AI techniques, such as transformer-based and diffusion-based models, contribute to these components, significantly enhancing quality and streamlining content creation.
    }
    \label{fig:character_generation}
\end{figure*}

In the Face section, we discuss the role of generative AI in tasks such as realistic face generation, facial reenactment, and attribute editing. Models like StyleGAN \cite{FFHQ} and its derivatives excel in producing high-resolution, photorealistic faces, enabling detailed control over facial features, expressions, and poses. Recent advances include emotion-conditioned synthesis and speech-driven facial animation, which enhance realism and interactivity. 

In the Expression section, we explore the transformative impact of generative AI in modeling and synthesizing facial expressions, a cornerstone of nonverbal communication and character animation. Techniques such as emotion-driven animation, facial expression recognition (FER), and expression retargeting have advanced significantly, enabling the creation of nuanced and lifelike facial dynamics. Notable models such as DiffSHEG \cite{chen2024diffshegdiffusionbasedapproachrealtime} revolutionized the field by integrating multimodal data and leveraging diffusion-based frameworks to achieve synchronized, expressive animations. These innovations power applications ranging from virtual avatars in immersive environments to adaptive tools in healthcare, driving progress in human-computer interaction while addressing challenges like real-time performance and identity preservation.

In the Image section, we discuss the role of generative AI in image synthesis and editing, highlighting its transformative impact on creative and practical applications. We categorize image datasets into generation and editing tasks, leveraging resources like LAION \cite{schuhmann2022laion}, COCO \cite{lin2014microsoft}, and ADE20K \cite{zhou2019semantic}, which provide extensive data for training. Cutting-edge models like diffusion-based architectures and hybrid frameworks enable high-quality image generation and manipulation. Techniques like ControlNet \cite{zhang2023adding} and disentangled latent space methods offer enhanced precision in aligning outputs with user inputs, whether for compositional synthesis or attribute adjustments.

In the Avatar section, we explore the advancements in generative AI for avatar creation, focusing on generating lifelike and stylized digital representations in both 2D and 3D. We examine datasets like WildAvatar \cite{huang2024wildavatarwebscaleinthewildvideo} and RenderMe-360 \cite{2023renderme360}, which provide multimodal data, including video, 3D body meshes, and audio, enabling the development of realistic avatars with detailed facial expressions and dynamic motions. Key models, such as SMPL \cite{SMPL} and FLAME \cite{flame}, serve as foundational frameworks for parametric body modeling. At the same time, methods like NeRF-based and diffusion-based approaches advance the synthesis of animate avatars with fine-grained control over appearance and motion.

In the Gesture section, we delve into generative AI's role in gesture generation, emphasizing its importance in creating human-like movements for interactive and immersive experiences. We review datasets such as BEAT \cite{liu2022beat} and Trinity Speech-Gesture \cite{articleexperess}, which integrate modalities like audio, text, and motion capture data to model realistic and context-aware gestures. Advanced models, including GAN-based and transformer-based architectures, enable co-speech gesture generation and stylistic customization, capturing temporal and semantic alignment with input signals. 

In the Motion section, we examine the advancements in generative AI for human motion synthesis, focusing on text-constrained motion generation. This task involves creating realistic and temporally coherent sequences of human body poses based on natural language descriptions. Cutting-edge models, such as MotionGPT \cite{zhang2024motiongptfinetunedllmsgeneralpurpose} and Motiondiffuse \cite{motiondiffuse}, have revolutionized this domain by leveraging variational autoencoders, diffusion-based frameworks, and large language models (LLMs) to capture diverse motion patterns and align them with textual prompts. These systems utilize representations like skeletal keypoints, hierarchical body joint rotations, or marker-based methods to ensure precision and adaptability. 

In the Object section, we delve into the transformative potential of generative AI in text-to-3D object generation, a critical technology for animation, gaming, and immersive environments. This task synthesizes 3D models with realistic geometry, textures, and material properties from textual descriptions, enabling seamless integration into virtual worlds. Key innovations include using neural radiance fields (NeRFs) for volumetric rendering and diffusion-based frameworks for high-fidelity geometry and texture synthesis. Techniques such as multi-view optimization and amortized training for real-time inference have significantly improved efficiency and quality. These advancements empower animators and designers to create detailed 3D assets rapidly, streamlining workflows for storytelling, world-building, and interactive media.

Finally, in the Texture section, we explore the advancements in generative AI for creating detailed surface properties, such as patterns, colors, and material characteristics, which are essential for enhancing the realism of 3D models. Texture generation is pivotal in animation and gaming by enriching geometric models with intricate details, enabling lifelike or stylized appearances. Recent innovations include text-guided texture synthesis, where natural language descriptions drive the creation of specific textures, and techniques like neural rendering and multi-view consistency ensure seamless integration across perspectives. 

In addition to these technical discussions, we include an Open Problems \& Research Directions section that identifies current challenges in generative AI for character animation, such as dataset limitations, real-time performance constraints, and ethical considerations. This is followed by a Conclusion, summarizing key findings and reflecting on future possibilities in this rapidly evolving field.

\paragraph{Contributions}
In this work, \textbf{(i)} we present an in-depth survey of the core components of animated character design, systematically analyzing the role of generative AI in each aspect of the animation pipeline. We review recent advancements in datasets, evaluation metrics, leading models, and applications for each character animation component. An overview of the different components and subfields is illustrated in \textbf{Figure \ref{fig:character_generation}}.
\textbf{(ii)} To support newcomers, we provide a comprehensive background section introducing foundational models and evaluation metrics, equipping readers with the necessary knowledge to enter the field (\textbf{Appendix~\S\ref{sec:app_background}}).
\textbf{(iii)} We propose a well-defined and systematic taxonomy (\textbf{Figure \ref{fig:taxonomy_full}}) that categorizes state-of-the-art models based on their primary contributions to character animation, highlighting key methodologies and emerging directions in AI-driven animation.
\textbf{(iv)} To facilitate future research, we compile and publicly share essential resources, including datasets, benchmarks, models, and evaluation tools, fostering accessibility in the field.
\textbf{(v)} We identify current research trends and knowledge gaps, offering insights and recommendations to guide future advancements in generative AI for character animation. 

\begin{center}

\definecolor{hidden-black}{RGB}{0,0,0} 

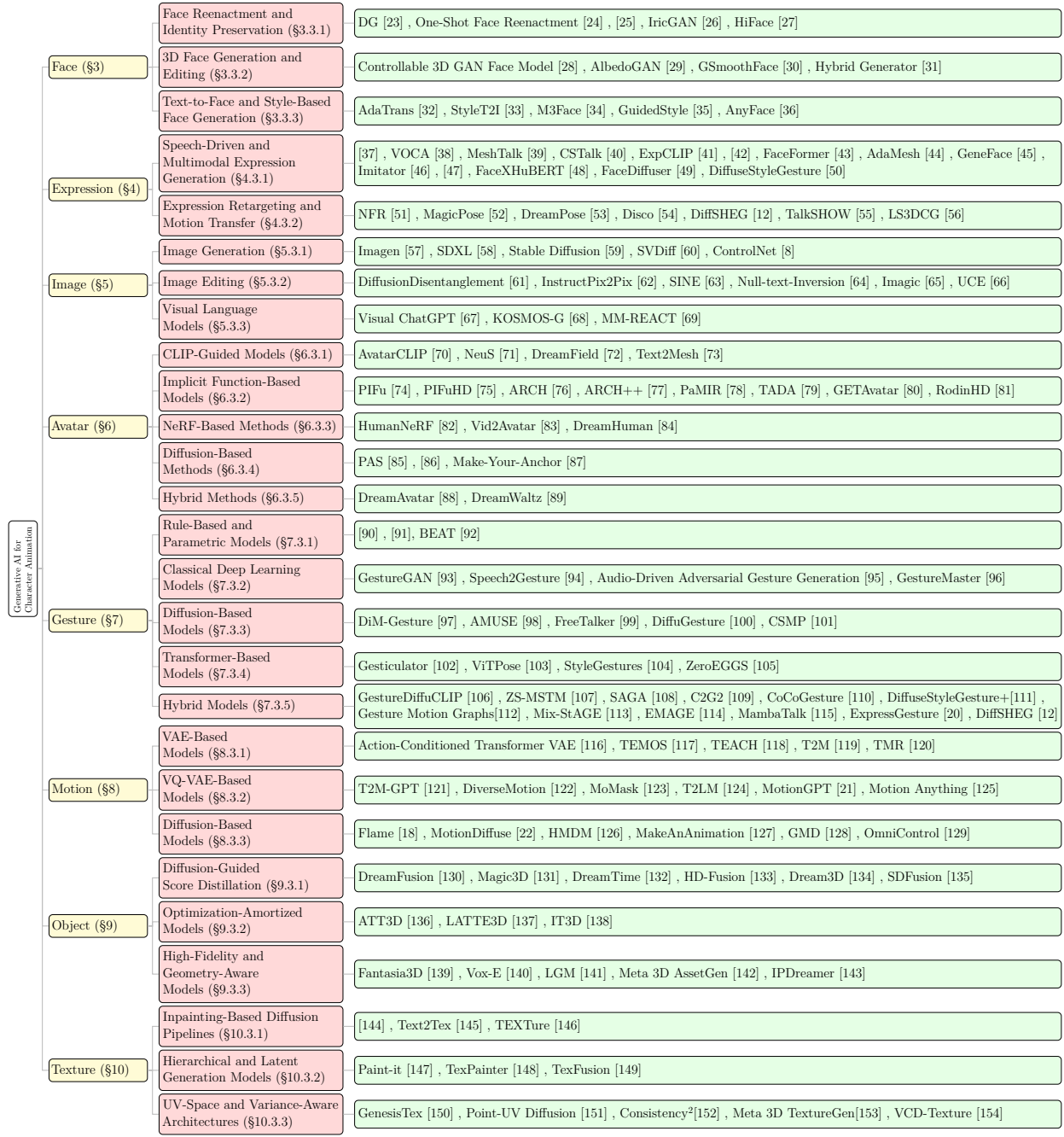
\begin{figure*}[t!]
    \centering
    \resizebox{\textwidth}{!}{
        \begin{forest}
            forked edges,
            for tree={
                child anchor=west,
                parent anchor=east,
                grow'=east,
                anchor=west,
                base=left,
                font=\normalsize,
                rectangle,
                draw=hidden-black,
                rounded corners,
                minimum height=2em,
                minimum width=4em,
                edge+={gray!50, line width=1pt},
                s sep=3pt,
                inner xsep=0.4em,
                inner ysep=0.6em,
                line width=0.8pt,
                text width=8.5em,
                where level=1{
                    text width=10em,
                    inner xsep=0.3em,
                    inner ysep=0.5em,
                    font=\Large,
                }{},
                where level=2{
                    text width=19em,
                    font=\Large,
                }{},
                where level=3{
                    text width=7em,
                }{},
                ver/.style={
                    fill=white!50,
                    rotate=90,
                    child anchor=north,
                    parent anchor=south,
                    anchor=center,
                    text width=9.5em
                },
                leaf/.style={
                    font=\Large,     
                    align=left,
                    text opacity=1,
                    inner sep=2pt,
                    fill opacity=.5,
                    fill=green!20, 
                    text=black,
                    text width=75.3em,
                    inner xsep=0.4em,
                    inner ysep=0.6em,
                    draw,
                }, 
            },
            [Generative AI for Character Animation, ver
                [Face~(\S\ref{sec_face}), fill=yellow!20
                    [{Face Reenactment and \\ Identity Preservation~(\S\ref{face_face reenactment_and_identity_ preservation})}, fill=red!15 [DG~\cite{Hsu_2022_CVPR} {,}
One-Shot Face Reenactment~\cite{Xu2022DesigningOU} {,}
\cite{DBLP:journals/corr/abs-2202-00046} {,}
IricGAN~\cite{IricGAN} {,}
HiFace~\cite{Chai_2023_ICCV}, leaf]]
                    [{3D Face Generation and \\Editing~(\S\ref{face_3D_face generation_and_editing})}, fill=red!15 [Controllable 3D GAN Face Model~\cite{Taherkhani_2023_WACV} {,}
AlbedoGAN~\cite{rai2023towards} {,}
GSmoothFace~\cite{Zhang2023GSmoothFaceGS} {,}
Hybrid Generator~\cite{HybridGenerator}, leaf]]
                    [{Text-to-Face and Style-Based \\ Face Generation~(\S\ref{face_text-to-face_and_style-based_face_generation})}, fill=red!15 [AdaTrans~\cite{Huang_2023_ICCV} {,}
StyleT2I~\cite{Li_2022_CVPR} {,}
M3Face~\cite{M3Face} {,}
GuidedStyle~\cite{HOU2022209} {,}
AnyFace~\cite{Sun_2022_CVPR}, leaf]]
                ]
                [Expression~(\S\ref{sec_expression}), fill=yellow!20
                    [{Speech-Driven and \\ Multimodal Expression \\ Generation~(\S\ref{expression_speech-driven_and_multimodal_expression_generation})}, fill=red!15 [\cite{Fan2021JointAM} {,}
VOCA~\cite{Cudeiro2019CaptureLA} {,}
MeshTalk~\cite{richard2021meshtalk} {,}
CSTalk~\cite{liang2024cstalkcorrelationsupervisedspeechdriven} {,}
ExpCLIP~\cite{zhong2023expclipbridgingtextfacial} {,}
\cite{bozkurt2023personalizedspeechdrivenexpressive3d} {,}
FaceFormer~\cite{faceformer2022} {,}
AdaMesh~\cite{chen2024adameshpersonalizedfacialexpressions} {,} 
GeneFace~\cite{ye2023geneface} {,} \\
Imitator~\cite{Thambiraja_2023_ICCV} {,}
\cite{10.1145/3623053.3623369} {,} 
FaceXHuBERT~\cite{FaceXHuBERT_Haque_ICMI23} {,} 
FaceDiffuser~\cite{10.1145/3623264.3624447} {,}
DiffuseStyleGesture~\cite{ijcai2023p650}, leaf]]
                    [{Expression Retargeting and \\ Motion Transfer~(\S\ref{expression_expression_retargeting_and_motion_transfer})}, fill=red!15 [NFR~\cite{10.1145/3588432.3591556} {,}
MagicPose~\cite{chang2024magicposerealistichumanposes} {,}
DreamPose~\cite{karras2023dreamposefashionimagetovideosynthesis} {,}
Disco~\cite{wang2023disco} {,}
DiffSHEG~\cite{chen2024diffshegdiffusionbasedapproachrealtime} {,}
TalkSHOW~\cite{Yi2022GeneratingH3} {,}
LS3DCG~\cite{10.1145/3472306.3478335}, leaf]]
                ]
                [Image~(\S\ref{sec_image}), fill=yellow!20
                    [{Image Generation~(\S\ref{image_image_generation})}, fill=red!15 [Imagen~\cite{saharia2022photorealistic} {,}
SDXL~\cite{podell2023sdxl} {,}
Stable Diffusion~\cite{rombach-high-res} {,}
SVDiff~\cite{han2023svdiff} {,}
ControlNet~\cite{zhang2023adding}, leaf]]
                    [{Image Editing~(\S\ref{image_image_editing})}, fill=red!15 [DiffusionDisentanglement~\cite{wu2023uncovering} {,}
InstructPix2Pix~\cite{brooks2023instructpix2pix} {,}
SINE~\cite{zhang2023sine} {,}
Null-text-Inversion~\cite{mokady2023null} {,}
Imagic~\cite{kawar2023imagic} {,}
UCE~\cite{gandikota2024unified}, leaf]]
                    [{Visual Language \\ Models~(\S\ref{image_visual_language_models})}, fill=red!15 [Visual ChatGPT~\cite{wu2023visual} {,}
KOSMOS-G~\cite{pan2023kosmos} {,}
MM-REACT~\cite{yang2023mm}, leaf]]
                ]
                [Avatar~(\S\ref{sec_avatar}), fill=yellow!20
                    [{CLIP-Guided Models~(\S\ref{avatar_CLIP-guided_models})}, fill=red!15 [AvatarCLIP~\cite{10.1145/3528223.3530094} {,}
NeuS~\cite{10.5555/3540261.3542342} {,}
DreamField~\cite{jain2021dreamfields} {,}
Text2Mesh~\cite{text2mesh}, leaf]]
                    [{Implicit Function-Based \\Models~(\S\ref{avatar_implicit_function-based_models})}, fill=red!15 [PIFu~\cite{saito2019pifu} {,}
PIFuHD~\cite{saito2020pifuhd} {,}
ARCH~\cite{arch} {,}
ARCH++~\cite{archpp} {,}
PaMIR~\cite{author2020paper1} {,}
TADA~\cite{liao2023tadatextanimatabledigital} {,}
GETAvatar~\cite{zhang2023getavatar} {,}
RodinHD~\cite{zhang2024rodinhd}, leaf]]
                    [{NeRF-Based Methods~(\S\ref{avatar_NeRF-based_methods})}, fill=red!15 [HumanNeRF~\cite{weng2022humannerf} {,}
Vid2Avatar~\cite{guo2023vid2avatar} {,}
DreamHuman~\cite{kolotouros2023dreamhuman}, leaf]]
                    [{Diffusion-Based \\ Methods~(\S\ref{avatar_diffusion-based_methods})}, fill=red!15 [PAS~\cite{azadi2023textconditionalcontextualizedavatarszeroshot} {,}
\cite{Bergman2023Articulated3H} {,}
Make-Your-Anchor~\cite{huang2024makeyouranchor}, leaf]]
                    [{Hybrid Methods~(\S\ref{avatar_hybrid_methods})}, fill=red!15 [DreamAvatar~\cite{cao2024dreamavatar} {,}
DreamWaltz~\cite{huang2023dreamwaltz}, leaf]]
                ]
                [Gesture~(\S\ref{sec_gesture}), fill=yellow!20
                    [{Rule-Based and \\ Parametric Models~(\S\ref{gesture_rule-based_and_parametric_models})}, fill=red!15 [\cite{inproceedingsevaluaaaa} {,}
\cite{:10.2312/egst.20141042}{,} BEAT \cite{cassell2001beat}, leaf]]
                    [{Classical Deep Learning \\ Models~(\S\ref{gesture_deep_learning-based_models})}, fill=red!15 [GestureGAN~\cite{tang2019gestureganhandgesturetogesturetranslation} {,}
Speech2Gesture~\cite{ginosar2019gestures} {,}
Audio-Driven Adversarial Gesture Generation~\cite{zhu2023taming} {,}
GestureMaster~\cite{10.1145/3536221.3558063}, leaf]]
                    [{Diffusion-Based \\ Models~(\S\ref{gesture_diffu_models})}, fill=red!15 [DiM-Gesture~\cite{zhang2024dimgesturecospeechgesturegeneration} {,}
AMUSE~\cite{Chhatre_2024_CVPR} {,}
FreeTalker~\cite{yang2024Freetalker} {,}
DiffuGesture~\cite{10.1145/3610661.3616552} {,}
CSMP~\cite{Deichler_2023}, leaf]]
                    [{Transformer-Based \\ Models~(\S\ref{gesture_transformer-based_models})}, fill=red!15 [Gesticulator~\cite{Kucherenko_2020} {,}
ViTPose~\cite{xu2022vitposesimplevisiontransformer} {,}
StyleGestures~\cite{alexanderson2020style} {,}
ZeroEGGS~\cite{ghorbani2022zeroeggszeroshotexamplebasedgesture}, leaf]]
                    [{Hybrid Models~(\S\ref{gesture_hybrid_models})}, fill=red!15 [GestureDiffuCLIP~\cite{ao2023gesturediffuclipgesturediffusionmodel} {,}
ZS-MSTM~\cite{fares2023zsmstmzeroshotstyletransfer} {,}
SAGA~\cite{sagaaaaaaa} {,}
C2G2~\cite{ji2023c2g2controllablecospeechgesture} {,}
CoCoGesture~\cite{qi2024cocogesturecoherentcospeech3d} {,} 
DiffuseStyleGesture+\cite{Yang_2023} {,} \\
Gesture Motion Graphs\cite{mewwwwwww} {,} 
Mix-StAGE~\cite{ahuja2020styletransfercospeechgesture} {,}
EMAGE~\cite{liu2024emageunifiedholisticcospeech} {,} 
MambaTalk~\cite{xu2025mambatalkefficientholisticgesture} {,}
ExpressGesture~\cite{articleexperess} {,}
DiffSHEG~\cite{chen2024diffshegdiffusionbasedapproachrealtime}, leaf]]
                ]
                [Motion~(\S\ref{sec_motion}), fill=yellow!20
    [{VAE-Based \\ Models~(\S\ref{motion_vae})}, fill=red!15 [Action-Conditioned Transformer VAE \cite{DBLP:journals/corr/abs-2104-05670} {,}
TEMOS \cite{petrovich22temos} {,}
TEACH \cite{athanasiou2022teachtemporalactioncomposition} {,}
T2M \cite{Guo_2022_CVPR} {,}
TMR \cite{petrovich2023tmrtexttomotionretrievalusing}, leaf]]
    [{VQ-VAE-Based \\ Models~(\S\ref{motion_vq_vae})}, fill=red!15 [T2M-GPT \cite{t2m-gpt} {,}
DiverseMotion \cite{diversemotion} {,}
MoMask \cite{guo2023momaskgenerativemaskedmodeling} {,}
T2LM \cite{lee2024t2lmlongterm3dhuman} {,}
MotionGPT \cite{zhang2024motiongptfinetunedllmsgeneralpurpose} {,}
Motion Anything \cite{zhang2025motion}, leaf]]
    [{Diffusion-Based \\ Models~(\S\ref{motion_diffusion})}, fill=red!15 [Flame \cite{flame} {,}
MotionDiffuse \cite{motiondiffuse} {,}
HMDM \cite{hmdm} {,}
MakeAnAnimation \cite{azadi2023makeananimationlargescaletextconditional3d} {,}
GMD \cite{gmd} {,}
OmniControl \cite{omnicontrol}, leaf]]
]
                [Object~(\S\ref{sec_object}), fill=yellow!20
                    [{Diffusion-Guided \\ Score Distillation~(\S\ref{object_diffusion_guided_score_distillation})}, fill=red!15 [DreamFusion~\cite{poole2022dreamfusion} {,}
Magic3D~\cite{lin2023magic3d} {,}
DreamTime~\cite{huang2024dreamtimeimprovedoptimizationstrategy} {,}
HD-Fusion~\cite{wu2023hdfusiondetailedtextto3dgeneration} {,}
Dream3D~\cite{xu2023dream3d} {,}
SDFusion~\cite{cheng2023sdfusion}, leaf]]
                    [{Optimization-Amortized \\ Models~(\S\ref{object_optimization_amortized_models})}, fill=red!15 [ATT3D~\cite{lorraine2023att3d} {,}
LATTE3D~\cite{xie2024latte3d} {,}
IT3D~\cite{chen2024it3d}, leaf]]
                    [{High-Fidelity and \\ Geometry-Aware \\ Models~(\S\ref{object_high_fidelity_and_geometry_aware_models})}, fill=red!15 [Fantasia3D~\cite{chen2023fantasia3d} {,}
Vox-E~\cite{sella2023vox} {,}
LGM~\cite{10.1007/978-3-031-73235-5_1} {,}
Meta 3D AssetGen~\cite{siddiqui2024meta3dassetgentexttomesh} {,}
IPDreamer~\cite{zeng2023ipdreamer}, leaf]]
                ]
                [Texture~(\S\ref{sec_texture}), fill=yellow!20
                    [{Inpainting-Based Diffusion Pipelines~(\S\ref{texture_inpainting_based_diffusion_pipelines})}, fill=red!15 [\cite{10203911} {,}
Text2Tex~\cite{chen2023text2textextdriventexturesynthesis} {,}
TEXTure~\cite{richardson2023texturetextguidedtexturing3d}, leaf]]
                    [{Hierarchical and Latent \\ Generation Models~(\S\ref{texture_hierarchical_and_latent_generation_models})}, fill=red!15 [
Paint-it~\cite{youwang2024paintittexttotexturesynthesisdeep} {,}
TexPainter~\cite{zhang2024texpaintergenerativemeshtexturing} {,}
TexFusion~\cite{cao2023texfusionsynthesizing3dtextures}, leaf]]
                    [{UV-Space and Variance-Aware \\ Architectures~(\S\ref{texture_UV_space_and_variance_aware_architectures})}, fill=red!15 [GenesisTex~\cite{gao2024genesistexadaptingimagedenoising} {,}
                    Point-UV Diffusion~\cite{yu2023texturegeneration3dmeshes} {,}
Consistency\textsuperscript{2}\cite{wang2024consistency2consistentfast3d} {,}
Meta 3D TextureGen\cite{bensadoun2024meta3dtexturegenfast} {,}
VCD-Texture~\cite{liu2024vcdtexturevariancealignmentbased}, leaf]]
                ]
            ]
        \end{forest}
    }
    \caption{Taxonomy of recent advances in generative AI for character animation, organized by key components within the animation environment.}
    \label{fig:taxonomy_full}
\end{figure*}

\end{center}

\section{Related Work}

Several surveys have recently reviewed generative AI techniques applied to specific animation domains. For instance, \cite{Kammoun2022FaceGANsSurvey} examines GAN-based face image synthesis and editing, detailing network architectures and evaluation protocols to produce realistic static face images. In the domain of temporal facial animation, \cite{Toshpulatov2023TalkingFaceSurvey} reviews deep learning methods for audio-driven talking head generation, including text-to-video conversion and speech-to-lip synchronization, while \cite{Dhanyalakshmi2024DeepfakeReenactSurvey} investigates face and full-body reenactment techniques utilized in deepfake generation. These surveys emphasize advances in lip synchronization, expression transfer, and identity preservation in generated faces.

A different set of surveys addresses the generation of body motions and gestures. \cite{Nyatsanga2023GestureSurvey} offers a comprehensive review of co-speech gesture generation and compares data-driven models that convert audio or text inputs into realistic character gestures. In a broader perspective, \cite{Zhu2024MotionSurvey} covers human motion synthesis methods conditioned on various modalities such as text, audio, and scene context. Their work categorizes motion generation approaches such as sequence-to-sequence models and diffusion-based motion synthesizers. They also discuss standard datasets and metrics for evaluating kinematic realism and contextual coherence.

Recent studies on virtual avatars and digital humans have focused on three-dimensional character modeling. \cite{Wang2024AvatarSurvey} reviews techniques for creating 3D human avatars, covering methods from geometry reconstruction that employ implicit neural representations and NeRF \cite{NERF} to fully generative avatar synthesis driven by high-level control signals. Their taxonomy spans body shape modeling, pose animation, and neural rendering of human appearance. Similarly, \cite{Xu2023ShapeGenSurvey} focuses on generating 3D object shapes based on voxels, point clouds, and meshes. This survey discusses how deep generative models such as GANs, VAEs, and diffusion models learn to produce diverse, high-quality 3D geometries.

More broadly, surveys on image synthesis provide valuable context for generative content creation applied to animation. \cite{Li2024ImageGenSurvey} chronicles the evolution of deep image generation models from early GANs and VAEs to state-of-the-art diffusion and transformer-based generators, highlighting improvements in output fidelity and diversity. Although texture generation has not been the sole focus of a dedicated survey, it is often treated as an essential component within comprehensive reviews on image or 3D content generation, particularly in discussions on style transfer and material synthesis.

In contrast to these focused investigations, our survey provides a unified perspective on generative AI for character animation, encompassing face, expression, gesture, motion generation, avatar creation, and visual elements such as objects and textures. This comprehensive framework organizes diverse methodologies into a coherent taxonomy and highlights interconnections across traditionally distinct subfields. 
For instance, the relationship between techniques used in facial expression synthesis and those applied in body gesture animation can be discovered through our survey. Furthermore, we provide an in-depth background on recent advancements to support newcomers in the field. To facilitate further research, we introduce key datasets used across these subfields and make relevant resources, including benchmarks and innovations, publicly available. Additionally, our survey integrates recent advancements in diffusion-based models and multimodal transformers while addressing standardized evaluation criteria for generative AI in animation. By bridging these gaps, we offer a comprehensive roadmap that is valuable to researchers and practitioners developing AI-driven character animation systems.

\section{Face}\label{sec_face}
Face generation, considering both 2D and 3D representations, is an expanding area of research aimed at synthesizing photorealistic and contextually consistent facial imagery \cite{rai2023towards}. Whether creating static portraits or dynamic, fully articulated 3D head models, the ability to generate convincing faces underpins various applications, from virtual assistants and entertainment to identity-protected data augmentation \cite{Chai_2023_ICCV}. This task often demands capturing subtle characteristics of human appearance, such as facial landmarks, skin textures, and expressions, while maintaining coherence across variations in lighting, pose, and emotional states. Generating realistic faces requires a deep understanding of human physiology and perception, coupled with advances in generative modeling, computer graphics, and machine learning. Recent progress, fueled by deep neural networks and data-driven techniques, has empowered models to produce faces with striking fidelity and personality. Furthermore, these approaches are integral to character animation workflows, allowing models to seamlessly integrate facial expressions, lip movements, and emotive nuances with full-body gestures \cite{Toshpulatov2023TalkingFaceSurvey}. Additionally, parametric face models such as FLAME \cite{flame}, when combined with advanced generative architectures, facilitate high-fidelity animation sequences that preserve identity and expression details across various motion contexts \cite{Dhanyalakshmi2024DeepfakeReenactSurvey}. As a result, face generation has evolved from simplistic compositing techniques to sophisticated frameworks capable of delivering nuanced, lifelike results that further bridge the gap between virtual and real-world experiences.

\subsection{Dataset}
\begin{table*}[t]
\centering
\renewcommand{\arraystretch}{1.4}
\resizebox{\textwidth}{!}{%
    \begin{tabular}{|>{\centering\arraybackslash}p{7cm}|p{7cm}|P{4cm}|c|}
\hline
\multicolumn{1}{|c|}{\textbf{Name}}  & \multicolumn{1}{c|}{\textbf{Statistics}} & \multicolumn{1}{c|}{\textbf{Modalities}} & \multicolumn{1}{c|}{\textbf{Link}} \\ \hline
RaFD \cite{Langner01122010}  & More than 8,000 images. Images of 67 models displaying eight facial expressions, photographed from five different angles. & Images & \href{https://rafd.socsci.ru.nl/}{RaFD} \\ \hline
MPIE \cite{MPIE} & Over 750,000 images with a broad range of variations in facial expressions, head poses, and lighting conditions. & Images & \href{https://www.cs.cmu.edu/afs/cs/project/PIE/MultiPie/Multi-Pie/Home.html}{MPIE} \\ \hline
VoxCeleb1 \cite{VoxCeleb1} & More than 100,000 utterances from 1,251 celebrities. & Audio, Video & \href{https://www.robots.ox.ac.uk/~vgg/data/voxceleb/}{VoxCeleb1} \\ \hline
VoxCeleb2 \cite{VoxCeleb1} & Over 1 million utterances from 6,112 celebrities. & Audio, Video & \href{https://www.robots.ox.ac.uk/~vgg/data/voxceleb/}{VoxCeleb2} \\ \hline
CelebA-HQ \cite{CelebA_HQ} & 30,000 images at a resolution of 1024x1024, providing detailed facial images of celebrities. & Images & \href{https://opendatalab.com/OpenDataLab/CelebA-HQ}{CelebA-HQ} \\ \hline
FaceForensics \cite{FaceForensics} & Over 1,000 video sequences with various face manipulations. & Video & \href{https://justusthies.github.io/posts/faceforensics/}{FaceForensics} \\ \hline
300-VW \cite{300-VW}, \cite{7298989}, \cite{7406475} & About 300 videos of faces in various scenarios and lighting conditions. & Video & \href{https://ibug.doc.ic.ac.uk/resources/300-VW/}{300-VW} \\ \hline
FFHQ \cite{FFHQ} & 70,000 images with extensive diversity, capturing various facial features, accessories, and environments. & Images & \href{https://www.computer.org/csdl/journal/tp/2021/12/08977347/1h2AHNHb9bW}{FFHQ} \\ \hline
AffectNet \cite{AffectNet} & Over 1 million images collected from the internet, with annotations for 11 different facial expressions and emotions. & Images & \href{http://mohammadmahoor.com/affectnet/}{AffectNet} \\ \hline
M\textsuperscript{3} CelebA \cite{M3Face} & Over 150K facial images annotated with semantic segmentation, facial landmarks, and captions in multiple languages. & Images, Text & \href{https://huggingface.co/datasets/m3face/M3CelebA/viewer}{M\textsuperscript{3} CelebA} \\ \hline
CUB \cite{WahCUB_200_2011} & Over 11,000 images of 200 bird species, each annotated with various attributes like species, part locations, and bounding boxes. & Images & \href{https://www.vision.caltech.edu/datasets/cub_200_2011/}{CUB} \\ \hline
CelebA-Dialog \cite{jiang2021talkedit} & 202,599 face images from 10,177 identities, annotated with 5 fine-grained attributes: Bangs, Eyeglasses, Beard, Smiling, Age, along with captions and user editing requests. & Images, Text & \href{https://mmlab.ie.cuhk.edu.hk/projects/CelebA/CelebA_Dialog.html}{CelebA-Dialog} \\ \hline
LS3D-W \cite{bulat2017far} & A dataset of 230,000 3D facial landmarks. & Images & \href{https://www.adrianbulat.com/face-alignment}{LS3D-W} \\ \hline
MERL-RAV \cite{kumar2020luvli} & Over 19,000 face images with diverse head pose, all annotated by 68 point landmarks and visibility status. & Audio, Video & \href{https://github.com/abhi1kumar/MERL-RAV_dataset}{MERL-RAV} \\ \hline
AFLW2000-3D \cite{DBLP:journals/corr/ZhuLLSL15} & Contains 2000 images with 68-point 3D facial landmarks, used to evaluate 3D facial landmark detection models with diverse head poses. & Images, 3D/Point Cloud Data & \href{https://github.com/tensorflow/datasets/blob/master/docs/catalog/aflw2k3d.md}{AFLW2000-3D} \\ \hline
FaceScape \cite{zhu2023facescape} & Over 18K textured 3D faces, captured from 938 subjects, each with 20 specific expressions. & 3D/Point Cloud Data & \href{https://facescape.nju.edu.cn/}{FaceScape} \\ \hline
    \end{tabular}%
}
\caption{Summary of different facial datasets, highlighting their key statistical characteristics and providing their corresponding links.}
\label{tab:face_datasets}
\end{table*}

Face datasets are crucial for various applications in character animation, including face generation, manipulation, recognition, and editing.
Face datasets can broadly be categorized into three main types: i) face generation, ii) face editing, and iii) face recognition manipulation datasets. Datasets for face generation include images of people with changing expressions, different poses, and changing lighting conditions. Such datasets will enable models to learn varied representations of human faces under various conditions. RaFD \cite{Langner01122010} and MPIE \cite{MPIE} have been two of the most common face datasets used in these contexts. RaFD offers systematically controlled facial expressions, gaze directions, and camera angles, validated through extensive user studies to ensure both emotional clarity and naturalness. Meanwhile, the Multi-PIE (MPIE) dataset significantly expands on pose, illumination, and expression variations across multiple recording sessions, making it a cornerstone for studying robust face recognition and generation under real-world conditions.

Face editing datasets, on the other hand, are designed to modify facial attributes such as age, emotion, or hairstyle in a way that preserves identity. These datasets typically contain high-resolution face images with rich annotations, including facial landmarks, emotions, and other semantic features. Examples include the CelebA-HQ \cite{CelebA_HQ} and the AffectNet \cite{AffectNet} datasets.

Finally, face recognition and manipulation datasets focus on tasks like face landmark detection, pose estimation, and identity manipulation. These are further annotated in terms of more attributes, like facial landmarks and bounding boxes, to further train the model and/or its evaluation processes. For instance, VoxCeleb1 \cite{VoxCeleb1} and VoxCeleb2 \cite{VoxCeleb1} are used for voice-and-face recognition tasks, while the 300-VW \cite{300-VW} is used for video data in facial landmark tracking in dynamic environments.
Table \ref{tab:face_datasets} presents a summary of widely used face datasets, their key statistics, modalities, and corresponding links.

\subsection{Evaluation}
Evaluating face generation models involves a multi-faceted approach that captures various aspects such as visual realism, perceptual quality, identity consistency, and task-specific alignment. These metrics ensure that synthesized faces not only replicate the appearance of real-world faces but also meet application-specific requirements, ranging from realistic avatars to context-aware virtual assistants.

A widely used metric for assessing the overall quality of face generation is the \textit{Fréchet Inception Distance (FID)}. This metric compares the distribution of features between real and generated images, providing a measure of how closely the generated faces resemble real-world examples. A lower FID score indicates a closer match and higher fidelity. Complementing FID, the \textit{CLIP Score} evaluates semantic alignment between generated faces and text or image prompts. This metric is particularly valuable in conditional generation tasks, ensuring the output matches the intended input.

To evaluate emotional expressiveness in generated faces, the \textit{Expression Score} measures the accuracy and intensity of facial expressions, such as happiness, anger, or sadness. This is crucial for applications where emotional communication is key. Additionally, the \textit{Mean Squared Error (MSE)} offers a straightforward comparison by quantifying the pixel-wise or feature-wise difference between generated faces and ground truth data. While MSE is an objective metric, it is often combined with perceptual metrics to provide a more nuanced assessment.

Perceptual quality is crucial in face evaluation, with metrics like \textit{LPIPS} capturing subtle visual differences such as texture and lighting, and \textit{Identity Consistency} ensuring recognizable and stable facial features across variations. For anatomical accuracy, \textit{Landmark Accuracy} assesses the precision of facial key points, while \textit{Lip Sync Quality} evaluates synchronization between lip movements and audio for a natural speech-driven generation.
Error-based metrics like \textit{Median Error} measure average deviations robustly, and \textit{Multi-View Identity Consistency (MVIC)} ensure identity consistency across angles, vital for applications like 3D modeling and virtual reality.

Advanced metrics are often introduced for specific scenarios. For example, the \textit{Fréchet Video Distance (FVD)} adapts the concept of FID to sequential data, measuring the temporal coherence and realism of generated videos. This metric is particularly important when evaluating face-generation models that output dynamic or time-sequenced content. The \textit{Perceptual Face Similarity (PFS)} metric assesses high-level perceptual features, focusing on human-aligned judgments of realism. In cases involving 3D face generation, the 3D Face Alignment Score evaluates the geometric accuracy of generated faces relative to ground truth 3D models or vertex data, ensuring precise depth and spatial alignment.

A range of recent methods has improved face generation with respect to visual quality, realism, and controllability. More specifically, StyleGAN-based models (e.g., StyleGAN2 \cite{Karras2019stylegan2} and StyleGAN3 \cite{Karras2021}) have demonstrated superior latent-space manipulation with fine-grained facial attribute control. Diffusion-based models have also gained popularity, utilizing iterative denoising methods for generating high-fidelity and photorealistic faces. Besides that, multiview-consistent models such as EG3D \cite{Chan2021} generalize beauty to 3D representations with well-coherent geometry across views. These state-of-the-art methods also usually involve advanced metrics such as FID, LPIPS, and Identity Consistency to evaluate aesthetic quality and identity consistency of synthesized faces, closely following the multi-faceted evaluation methodologies outlined above.

\subsection{Models} Face generation models generate photorealistic 2D or 3D faces by leveraging generative models such as GANs. The recent trends in the area include multi-component architectures where static and dynamic features are separated for the exact control of facial expressions and textures. Basic models like StyleGAN \cite{FFHQ} and ResNet \cite{7780459} form the backbone, while innovations involving emotion conditioning and speech-driven synthesis make them more realistic and interactive.

\subsubsection{Face Reenactment and Identity Preservation}\label{face_face reenactment_and_identity_ preservation} 
\begin{figure}[t]
\begin{center}
  \includegraphics[width=\textwidth]{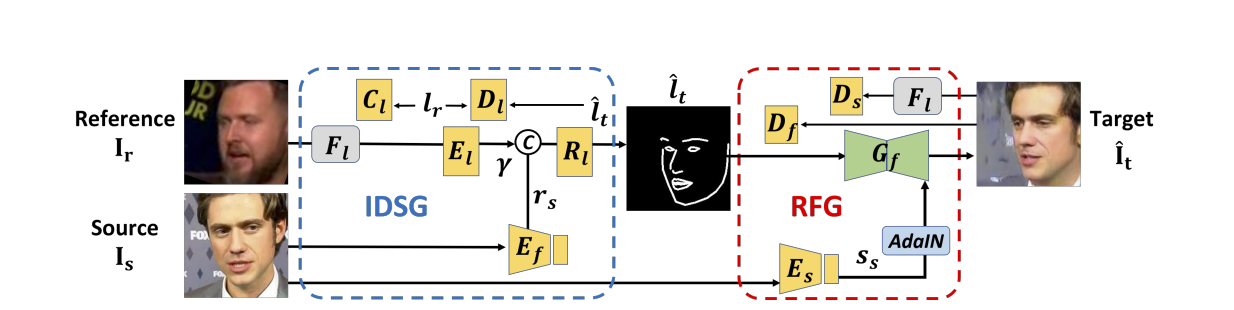}
\end{center}
\caption{\label{fig:DG}
Overview of the Dual-Generator (DG)  \cite{Hsu_2022_CVPR} network, which consists of two generators: the ID-preserving Shape Generator (IDSG) and the Reenacted Face Generator (RFG). 
Given a source face $I_s$ and a reference face $I_r$, the IDSG transforms the reference’s actions into landmarks $\hat{l}_t$. 
Using these landmarks and $I_s$, the RFG produces a reenacted face $\hat{I}_t$ that matches the pose and expression of $I_r$ while preserving the identity of $I_s$. Reprinted from \cite{Hsu_2022_CVPR}.
}
\end{figure}

These models focus on transferring expressions and poses from one identity to another while preserving the source or target identity. Early methods, such as Dual-Generator (DG) \cite{Hsu_2022_CVPR}, address large-pose reenactment using two modules: the ID-Preserving Shape Generator and the Reenacted Face Generator, the latter based on StarGAN2 \cite{choi2020starganv2diverseimage}. Figure \ref{fig:DG} shows more details of the architecture. DG employs 3D landmarks to guide expression transfer and introduces a visible local shape loss ($L_{vls}$) to mitigate occlusions in rotated faces. While effective in preserving shape integrity, its reliance on structural priors such as 3D landmarks limits robustness.

Building on these limitations, One-Shot Face Reenactment \cite{Xu2022DesigningOU} removes dependencies on explicit landmark detection and 3D coefficients. Instead, it introduces a Feature Disentanglement module to separate identity and attributes, along with Feature Displacement Fields to spatially align identity features with target expressions or poses. Additionally, its Identity Transfer module leverages self-attention to enforce consistency. While more robust than prior approaches, it remains inadequate for extreme poses and struggles with complex transformations.

For more efficient reenactment, \cite{DBLP:journals/corr/abs-2202-00046} directly maps pose and expression variations from a 3D face model into the StyleGAN2 \cite{Karras2019stylegan2} latent space. This does not require explicit identity/pose embeddings and gives realistic, identity-preserving results, but has a limited expressive linear mapping. IricGAN \cite{IricGAN} further improves controllability using two modules: a Hierarchical Feature Combination module to preserve identity and semantics, and an Attribute Regression Module for intensity-aware smooth edits. It replaces pre-trained recognition networks with a custom training approach, requiring meticulous tuning to ensure stability.

Finally, HiFace \cite{Chai_2023_ICCV} advances 3D face modeling by disentangling static and dynamic details using its SD-DeTail (Static and Dynamic Decoupling for DeTail Reconstruction) Module. It uses ResNet-50 \cite{7780459} for feature extraction and AdaIN (Adaptive Instance Normalization) \cite{huang2017arbitrarystyletransferrealtime} for detail synthesis to enable smooth, high-fidelity animation transitions past the entanglement limitation of earlier methods. It allows the model to reconstruct a high-fidelity 3D face from a single image with realistic and animatable details.

\subsubsection{3D Face Generation and Editing}\label{face_3D_face generation_and_editing} These models generate or edit 3D faces, often with control over expressions, lighting, or geometry. To improve identity preservation and expression control, a Controllable 3D GAN Face Model \cite{Taherkhani_2023_WACV} utilizes a Supervised Auto-Encoder for disentangling identity and expression into disjoint latent spaces, maintaining their correlation through a shared decoder. Expression levels are controlled precisely through a Conditional GAN (cGAN) \cite{mirza2014conditionalgenerativeadversarialnets}, which is essentially characterized by a loss function incorporating both a reconstruction and a classification term. Although this approach is effective in separation and management, it remains data-intensive and computationally elaborate because of its complicated structure. Pushing beyond the limitations of illumination control and texture realism, AlbedoGAN \cite{rai2023towards} proposes a self-supervised model that utilizes StyleGAN2 \cite{Karras2019stylegan2} latent space to generate high-resolution albedo and very detailed 3D facial geometry. It improves texture coherence with mesh refinement combined with a per-vertex displacement map integrated into the FLAME model \cite{flame}, while also ensuring identity consistency with a symmetric reconstruction loss. Notably, it incorporates CLIP to enable text-based 3D face editing, a feature absent in earlier models. Nevertheless, it remains heavily dependent on StyleGAN2 \cite{Karras2019stylegan2} and struggles with capturing fine details, such as hair.

Focusing on animation and motion realism, GSmoothFace \cite{Zhang2023GSmoothFaceGS} introduces a speech-driven talking face generation model through fine-grained 3D modeling. Its two-stage pipeline is comprised of a Target Adaptive Face Translation module and Morphology Augmented Face Blending (MAFB), enabling identity-coherent and stable video synthesis. Innovations like bias-based cross-attention improve lip synchronization, while MAFB alleviates distortions in face blending. Concentrating on deep learning with conventional graphics, the Hybrid Generator \cite{HybridGenerator} unites StyleGAN2-based neural generation with interpretable elements such as the FLAME 3D head model and a differentiable renderer. This hybrid framework enables photorealistic but controllable adjustment of facial identity, expression, and pose, providing accuracy and interpretability that cannot be attained with solely neural or graphics-based models.

\subsubsection{Text-to-Face and Style-Based Face Generation}\label{face_text-to-face_and_style-based_face_generation} These models generate faces from text prompts or manipulate attributes in the latent space. Traditional approaches rely on direct linear manipulation of latent codes, but AdaTrans \cite{Huang_2023_ICCV} introduces a more adaptive mechanism using nonlinear latent space transformations. This improves the model’s ability to handle complex conditional edits while preserving image realism, surpassing previous linear editing methods based on StyleGAN \cite{FFHQ}. Advancing textual control, StyleT2I \cite{Li_2022_CVPR} addresses attribute compositionality and faithfulness issues by incorporating a CLIP-guided contrastive loss and a Text-to-Direction module. The latter learns latent directions corresponding to textual descriptions, while Compositional Attribute Adjustment ensures that multiple attributes are correctly and coherently expressed.

The multimodal extension with M3Face \cite{M3Face} facilitates the generation and editing process by utilizing multilingual text, segmentation masks, or landmarks. It uses models such as Muse \cite{chang2023musetexttoimagegenerationmasked} and VQ-GAN \cite{esser2021tamingtransformershighresolutionimage} to extract conditioning inputs, which are subsequently fed into ControlNet \cite{zhang2023adding} for synthesis and optimized with Imagic for high-fidelity fine-tuning. The architecture is able to seamlessly combine the generation and editing process in an end-to-end pipeline, thereby greatly enhancing accessibility and versatility for various languages and input modalities.

Focusing on semantics and control, GuidedStyle \cite{HOU2022209} introduces a knowledge network built from a pre-trained attribute classifier to guide semantic face editing in StyleGAN \cite{FFHQ}. Sparse attention enables controlled, layer-wise editing, preserving disentanglement and preventing unintended attribute changes, ensuring precise and interpretable edits. AnyFace \cite{Sun_2022_CVPR} marks a step toward open-world, free-form text-to-face generation. It employs a two-stream framework that separately handles text-to-face synthesis and face reconstruction, improving visual-text alignment. Leveraging CLIP encoders and a Cross-Modal Distillation module, it integrates text and image features within StyleGAN’s latent space. Additionally, a Diverse Triplet Loss promotes variation and semantic consistency, addressing challenges such as mode collapse and the rigid vocabularies of earlier models.

\begin{figure}[t]
\begin{center}
  \includegraphics[width=\textwidth]{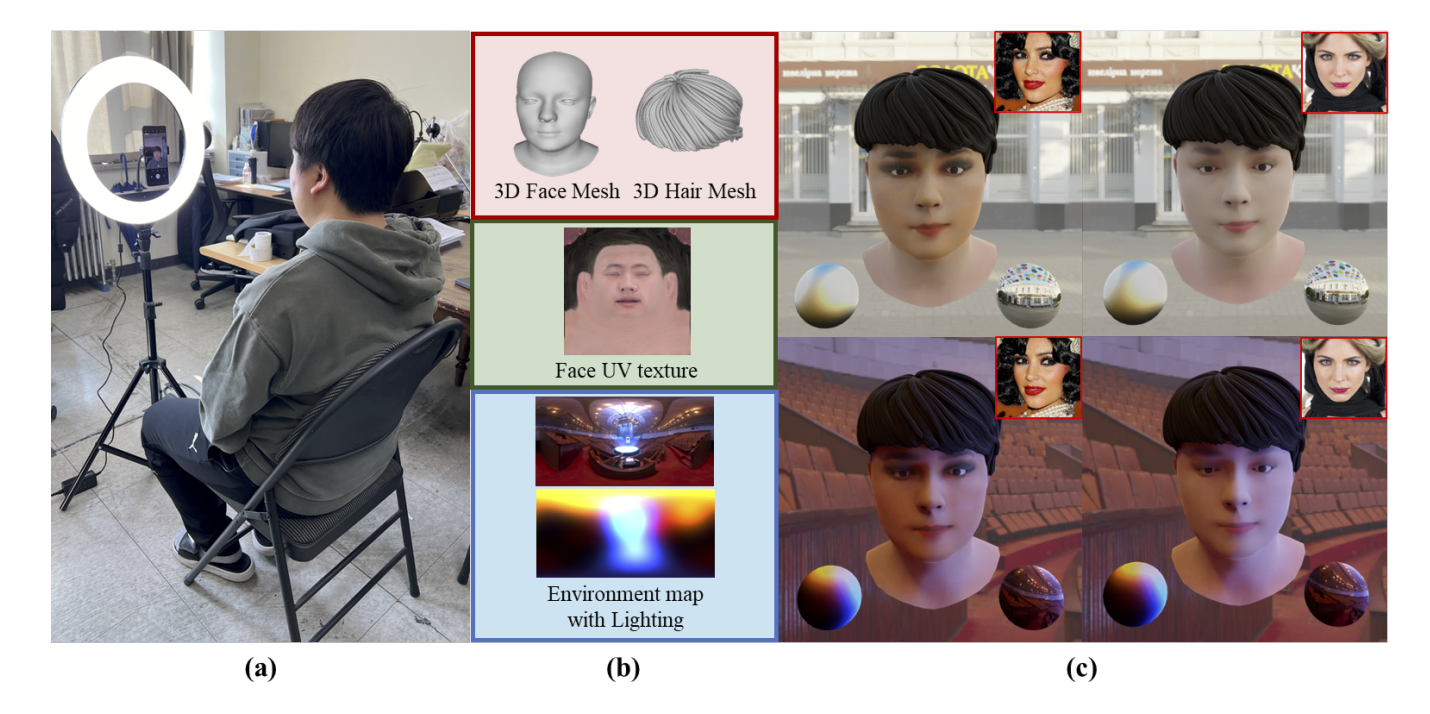}
\end{center}
\caption{\label{fig:In-ViTe}
Overview of InViTe \cite{10.24963/ijcai.2024/1008} in three stages: 
(a) capturing a user’s face to produce a personalized 3D model, 
(b) generating intermediate outputs for rendering, 
and (c) performing 3D face manipulation (such as makeup style changes) on a mobile device. 
Reprinted from \cite{10.24963/ijcai.2024/1008}.
}
\end{figure}

\subsection{Application}
Face generation and editing, both in 2D and 3D, are integral to many applications in generative AI, including virtual avatars, face recognition, animation, and expression transfer. In 3D, generative models such as GANs \cite{goodfellow2014generativeadversarialnetworks} and cGANs \cite{mirza2014conditionalgenerativeadversarialnets} are central to creating realistic virtual avatars for use in video games, virtual reality (VR), and augmented reality (AR). With these models, users are able to create highly detailed customizations, adjusting face features, expressions, and textures to make avatars or characters that look so real for personalized experiences. The Individual Virtual Transfer (In-ViTe) \cite{10.24963/ijcai.2024/1008} system, as Figure \ref{fig:In-ViTe} represents, allows users to create and customize 3D faces according to their preferences, contributing to the personalization of virtual avatars. AvatarGen \cite{zhang2022avatargen3dgenerativemodel} is a 3D generative model that facilitates the unsupervised creation of clothed virtual humans with various appearances and animatable poses, important for applications in AR/VR. They also improve facial recognition systems, enhance animation for movies, and generate synthetic training data for AI models. Expression transfer in 3D faces is critical for dynamic content like movie characters and virtual assistants. These models are also applied in biometric security systems to recognize faces under varying conditions.

In 2D, face generation plays a critical role in creative fields such as digital art, personalized avatars, and portrait synthesis. Models like StyleGAN \cite{FFHQ} have been widely used for generating realistic portraits, enabling applications in gaming, advertising, and content creation. Text-to-image and text-to-3D models are becoming increasingly popular for generating faces based on user prompts, allowing for high customization in virtual environments and applications like virtual assistants or digital twins in health and wellness sectors \cite{wang2025singleimagefacegeneralisable}. The capability to modify not only appearance but also expressions and poses enables new avenues for interactive media, advertising, and film, where characters could respond dynamically to user interactions or narrative changes.

\section{Expression}\label{sec_expression}
Facial expressions result from small muscle movements in the face and serve as key indicators of a person’s emotional state \cite{HARLEY201689}. The Facial Action Coding System (FACS) was the first widely adopted and empirically validated framework for classifying emotions based on facial expressions \cite{Ekman1978}.

Facial expressions are central to various tasks in computer vision and animation. They play a crucial role in nonverbal communication and are essential for generating believable and engaging animated characters \cite{zhong2023expclipbridgingtextfacial}. For instance, accurate lip sync and facial expressions enhance the expressiveness, personality, and storytelling of animated characters. Similarly, facial expression recognition (FER) is a key task in computer vision that focuses on identifying and categorizing emotional expressions. FER has applications in healthcare, security, and driver safety, contributing to its adoption in human-computer interaction for intelligent and adaptive systems \cite{SAJJAD2023817}. Another related task is Facial Expression Retargeting, which involves generating new images of a person with controlled poses and facial expressions while preserving their identity \cite{10.1145/3588432.3591556, chang2024magicposerealistichumanposes}.

\subsection{Dataset}
Facial expression-related tasks often rely on an auxiliary modality that supports the primary objective, whether in animation, recognition, or other applications. To accurately capture and represent facial expressions, datasets typically include detailed annotations of facial movements using blend shapes and Action Units (AUs).

\paragraph{Blendshapes} Blendshapes are predefined facial deformations that can be combined to generate a wide range of expressions. They are particularly important in animation and facial expression synthesis, allowing for precise control over facial features and enabling smooth, realistic transitions between expressions.

\paragraph{Action Units (AUs)} AUs are a core component of FACS, which categorizes facial movements based on the muscles involved. Each AU corresponds to a specific muscle action, such as raising the eyebrows or tightening the lips. By annotating facial expressions with AUs, datasets provide a granular and interpretable way to analyze and generate facial expressions, making them essential for tasks like Facial Expression Recognition (FER) and animation.

Datasets such as TEAD (Text-Expression Aligned Dataset) \cite{zhong2023expclipbridgingtextfacial} leverage both blend shapes and AUs to align natural language semantics with facial expressions. TEAD links text descriptions with emotion tags, AUs, and blend shape weights, facilitating tasks like emotion-driven text-to-face generation. Similarly, the BEAT dataset \cite{liu2022beat}, widely used in speech-driven facial animation research, integrates 52-dimensional blend shape weights with audio, emotion, and gesture modalities, making it a valuable resource for multimodal emotion expression studies.

Beyond facial expression data, most datasets incorporate auxiliary modalities such as speech, images, or video to support various applications. For instance, VOCASET \cite{Cudeiro2019CaptureLA} pairs high-fidelity 4D facial scans with synchronized audio, enabling the generation of 3D facial animations from speech input. The SHOW dataset \cite{Yi2022GeneratingH3} combines video, audio, facial expressions, and pose data to generate holistic 3D human motion from speech, highlighting the role of auxiliary modalities in enhancing expressiveness and realism.

By integrating blend shapes, AUs, and auxiliary modalities, facial expression datasets provide essential tools for developing sophisticated models capable of capturing the subtle nuances of human expression. More details on these datasets can be found in Table \ref{table:expression}.

\begin{table*}[t]
\centering
\renewcommand{\arraystretch}{1.4}
\resizebox{\textwidth}{!}{%
    \begin{tabular}{|>{\centering\arraybackslash}p{7cm}|p{7cm}|P{4cm}|c|}
\hline
\multicolumn{1}{|c|}{\textbf{Name}}  & \multicolumn{1}{c|}{\textbf{Statistics}} & \multicolumn{1}{c|}{\textbf{Modalities}} & \multicolumn{1}{c|}{\textbf{Link}} \\ \hline

BEAT  \cite{liu2022beat}  & 76 hours of speech data, paired with 52D facial blend shape weights; 30 speakers performing in 8 distinct emotional styles across 4 languages. & Audio, Images, Video, Text & \href{https://pantomatrix.github.io/BEAT/}{BEAT} \\ \hline

MEAD  \cite{10.1007/978-3-030-58589-1_42} & A talking-face video corpus featuring 60 actors and actresses talking with eight different emotions at three intensity levels; approximately 40 hours of audio-visual clips per person and view. & Video, Audio, Text, Images & \href{https://wywu.github.io/projects/MEAD/MEAD.html}{MEAD} \\ \hline

TEAD \cite{zhong2023expclipbridgingtextfacial}  & 50,000 quadruples, each including text, emotion tags, Action Units, blend shape weights, and situation sentences. & Text, Images & - \\ \hline

JAFFE  \cite{Lyons1998TheJF} & 213 images of 10 Japanese female models posing 7 facial expressions, annotated with average semantic ratings from 60 annotators. & Images & \href{https://zenodo.org/records/3451524}{JAFFE} \\ \hline

MMI Facial Expression \cite{1521424} & Over 2900 videos and high-resolution still images of 75 subjects. & Video, Images, Text & \href{https://mmifacedb.eu/}{MMI} \\ \hline

Multiface \cite{wuu2023multifacedatasetneuralface} & High-quality recordings of the faces of 13 identities. An average of 23,000 frames per subject; each frame includes roughly 160 different camera views. & Images, Audio, Tabular Data & \href{https://github.com/facebookresearch/multiface}{Multiface} \\ \hline

ICT FaceKit \cite{Li2020LearningFO} & 4,000 high-resolution facial scans of 79 subjects (34 female, 45 male) aged 18--67, plus 99 full-head scans and 26 expressions per subject. & 3D/Point Cloud Data, Images & \href{https://github.com/ICT-VGL/ICT-FaceKit}{ICT FaceKit} \\ \hline

TikTok Dataset \cite{Jafarian_2021_CVPR_TikTok} & Over 300 single-person dance videos (10--15 seconds each), extracted at 30fps, yielding 100K+ frames. Includes segmented images and computed UV coordinates. & Video, Images, Tabular Data & \href{https://www.yasamin.page/hdnet_tiktok\#h.jr9ifesshn7v}{TikTok Dataset} \\ \hline

Everybody Dance Now \cite{chan2019dance} & Long single-dancer videos for training and evaluation; includes both self-filmed videos and short YouTube videos. & Video, Tabular Data & \href{https://carolineec.github.io/everybody_dance_now/}{Everybody Dance Now} \\ \hline

Obama Weekly Footage \cite{10.1145/3072959.3073640} & 17 hours of video footage, nearly two million frames, spanning eight years. & Video, Audio & \href{https://grail.cs.washington.edu/projects/AudioToObama/}{Obama Weekly Footage} \\ \hline

VoxCeleb2 \cite{chung18b_interspeech} & Over 1 million utterances from over 6,000 speakers, collected from YouTube videos with 61\% male speakers. & Audio, Video & \href{https://www.robots.ox.ac.uk/~vgg/data/voxceleb/vox2.html}{VoxCeleb2} \\ \hline

BIWI \cite{10.1007/978-3-642-23123-0_11} & Over 15K images of 20 people recorded with a Kinect while turning their heads around freely. & 3D/Point Cloud Data, Images, Tabular Data, Text & \href{https://paperswithcode.com/dataset/biwi-kinect-head-pose}{BIWI} \\ \hline

VOCASET \cite{Cudeiro2019CaptureLA} & About 29 minutes of high-fidelity 4D scans captured at 60fps, synchronized with audio; features 12 speakers with 40 sequences per subject (each sequence consists of English sentences lasting 3--5 seconds). & 3D/Point Cloud Data, Audio & \href{https://voca.is.tue.mpg.de/}{VOCASET} \\ \hline

SHOW  \cite{Yi2022GeneratingH3} & Contains SMPLX \cite{Pavlakos2019ExpressiveBC} parameters of 4 persons reconstructed from videos; includes 88-frame motion clips for training and validation. & Video, Audio, Images, Tabular Data & \href{https://github.com/yhw-yhw/SHOW}{SHOW} \\ \hline
    \end{tabular}%
}
\caption{Overview of various datasets related to facial expressions, including their main statistical features, modalities, and links.}
\label{table:expression}
\end{table*}

\subsection{Evaluation}
Evaluating facial expression models requires a diverse set of metrics to assess their performance across key dimensions such as realism, identity preservation, synchronization, and diversity. These metrics offer critical insights into different methods' effectiveness and applicability in real-world scenarios. The evaluation process broadly consists of objective metrics for quantitative analysis and user studies for perceptual validation.

Quantitative metrics serve as the foundation for performance assessment, targeting specific aspects of model output. For instance, \textit{Lip Vertex Error (LVE)} and \textit{Emotional Vertex Error (EVE)} measure spatial accuracy by computing the Euclidean distance between predicted and ground-truth vertices for lip and emotional regions, respectively. Similarly, \textit{Mean Absolute Error (MAE)} evaluates the alignment of generated expressions with ground-truth Action Units, providing a fine-grained analysis of expression dynamics. These metrics help ensure precise motion synthesis while identifying areas for improvement, such as synchronization with speech or text inputs.

Generative metrics such as \textit{Fréchet Inception Distance (FID)} and \textit{Learned Perceptual Image Patch Similarity (LPIPS)} are widely used to evaluate perceptual quality. FID measures distributional similarity between generated and real images, capturing visual fidelity, while LPIPS leverages deep learning features to assess structural coherence. Additionally, the \textit{Structural Similarity Index (SSIM) \cite{wangzhou2004image}} and \textit{Peak Signal-to-Noise Ratio (PSNR)} wang2004image quantify low-level image quality, detecting deviations from ground-truth facial features.

Identity preservation is another critical evaluation aspect, particularly for tasks like expression retargeting. Metrics such as \textit{Face-Cosine Similarity} \cite{chang2024magicposerealistichumanposes} measure identity consistency by comparing facial feature embeddings of generated and source images, ensuring that expressions remain faithful to the original identity. For dynamic expressions, temporal and motion-based metrics play a key role. \textit{Fréchet Motion Distance (FMD)}, \textit{Fréchet Expression Distance (FED)}, and \textit{Fréchet Gesture Distance (FGD)} assess the realism and synchronization of motion sequences. Additionally, the \textit{Percent of Correct Motions (PCM)} and \textit{Semantic Relevance Gesture Recognition (SRGR)} evaluate how well facial motions align with speech and semantic content, enhancing the contextual appropriateness of generated animations.

\paragraph{User studies} These complement objective evaluations by capturing subjective perceptions of realism and expressiveness. Participants rate the naturalness and emotional resonance of generated expressions, providing insights that quantitative metrics may overlook. A/B tests are commonly used in these studies to compare models against baselines, highlighting their relative strengths and weaknesses.

Benchmark evaluations in facial expression modeling have evolved to assess models across multiple dimensions, including realism, identity preservation, synchronization, and computational efficiency. Recent advancements, particularly diffusion-based frameworks, have set new standards for generating expressive and temporally coherent animations. Benchmarks now emphasize models that handle diverse datasets and multimodal inputs, reflecting the growing demand for real-time and context-aware applications. State-of-the-art approaches, such as AdaMesh \cite{chen2024adameshpersonalizedfacialexpressions} and DiffSHEG \cite{chen2024diffshegdiffusionbasedapproachrealtime}, showcase architectures that balance expressiveness with computational efficiency. These benchmarks serve as a guiding framework for future advancements in generative AI for facial expression modeling.

\begin{figure}[t]
\begin{center}
  \includegraphics[width=\textwidth]{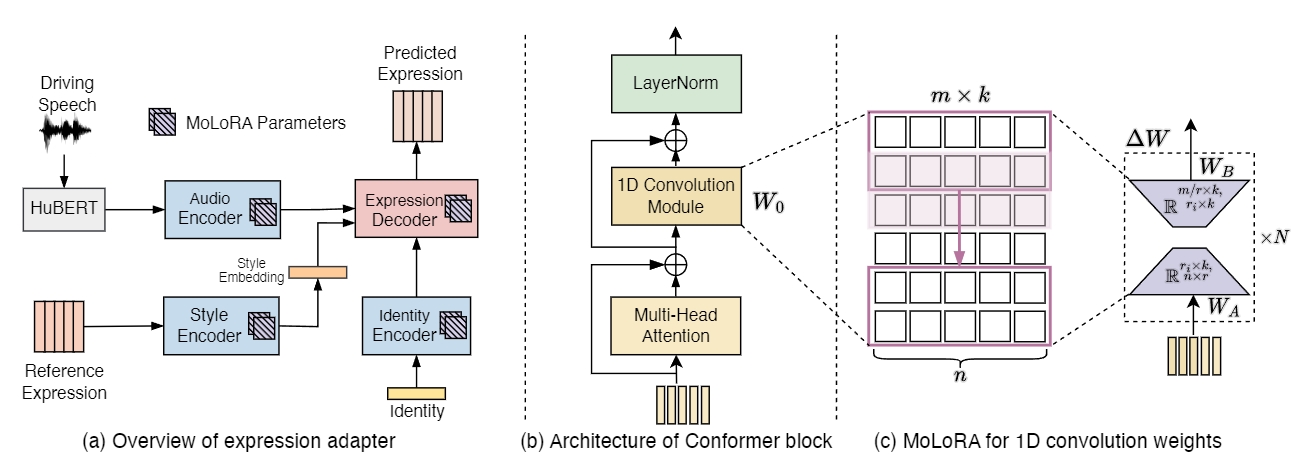}
\end{center}
\caption{\label{fig:adamesh}
Overview of AdaMesh \cite{chen2024adameshpersonalizedfacialexpressions} model: (a) The expression adapter integrates MoLoRA \cite{zadouri2023pushing} parameters (striped patches) into pre-trained encoders and the decoder to enable efficient adaptation for facial expressions. (b) Architecture of the Conformer block \cite{9747691}, showcasing its 1D convolution module and multi-head attention. (c) Illustration of MoLoRA applied to the convolution operator, with low-rank decomposition enhancing adaptation efficiency. MoLoRA is also applied to linear layers in the expression adapter. 
Reprinted from \cite{chen2024adameshpersonalizedfacialexpressions}.
}
\end{figure}

\subsection{Models}
Recent advancements in facial expression modeling have significantly improved expression retargeting and speech-driven animation. Expression retargeting enables the transfer of facial expressions while preserving the target's identity, a crucial capability for animation and virtual reality applications. Meanwhile, speech-driven models have evolved to synthesize expressive facial motions in response to audio or text inputs, enhancing the realism of animated characters.

\subsubsection{Speech-Driven and Multimodal Expression Generation}\label{expression_speech-driven_and_multimodal_expression_generation}

\cite{Fan2021JointAM} introduces a joint audio-text model for 3D facial animation, integrating a GPT-2-based text encoder with a dilated convolution audio encoder. This bimodal approach improves upper-face expressiveness and lip synchronization compared to VOCA \cite{Cudeiro2019CaptureLA} and MeshTalk \cite{richard2021meshtalk}, though it lacks head and gaze control. Similarly, CSTalk \cite{liang2024cstalkcorrelationsupervisedspeechdriven} employs a transformer-based encoder to capture correlations across facial regions, enhancing realism in emotional speech-driven animations. However, its expressivity is limited to five emotions.

ExpCLIP \cite{zhong2023expclipbridgingtextfacial} aligns text, image, and expression embeddings via CLIP encoders, enabling expressive speech-driven facial animation from text/image prompts. It uses the TEAD dataset and Expression Prompt Augmentation to adapt to diverse emotional styles. \cite{bozkurt2023personalizedspeechdrivenexpressive3d} enhances personalization by disentangling style and content representations, improving identity retention and transition smoothness. Compared to FaceFormer \cite{faceformer2022}, it achieves superior audio-visual synchronization, though computational efficiency remains a challenge.

AdaMesh \cite{chen2024adameshpersonalizedfacialexpressions} introduces an Expression Adapter (MoLoRA-enhanced) and Pose Adapter (retrieval-based) for personalized speech-driven facial animation. Compared to GeneFace \cite{ye2023geneface} and Imitator \cite{Thambiraja_2023_ICCV}, it demonstrates improved expressiveness, diversity, and synchronization. Figure \ref{fig:adamesh} shows more architectural details.
Similarly, \cite{10.1145/3623053.3623369} explores disentangling emotional expressiveness with FaceXHuBERT \cite{FaceXHuBERT_Haque_ICMI23} and FaceDiffuser \cite{10.1145/3623264.3624447}, highlighting stochastic approaches for motion variability.

\subsubsection{Expression Retargeting and Motion Transfer}\label{expression_expression_retargeting_and_motion_transfer}
\begin{figure}[t!]
\begin{center}
  \includegraphics[width=\textwidth]{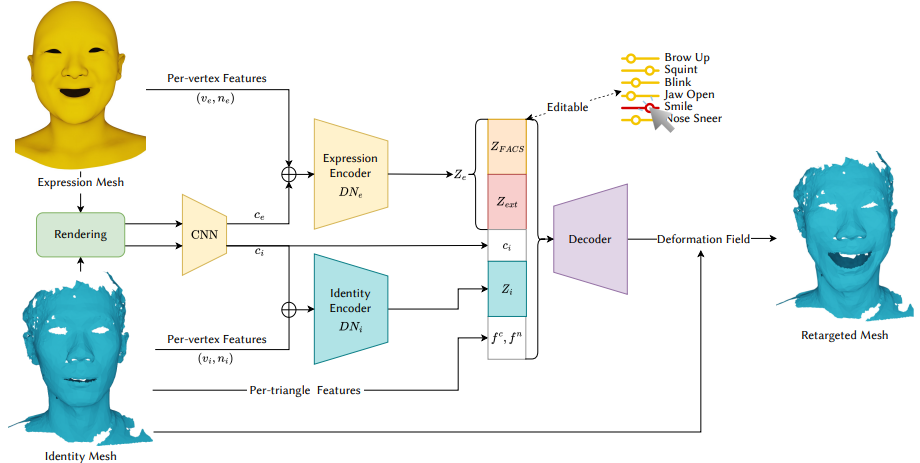}
\end{center}
\caption{\label{fig:nfr}
Overview of Neural Face Rigging (NFR) \cite{10.1145/3588432.3591556}. The model extracts an expression code \( z_e \) from an unrigged mesh with an unknown expression (yellow) and an identity code \( z_i \) from a target neutral mesh (cyan). The extracted codes are combined to generate a retargeted mesh (blue) with transferred expressions and identity-preserving deformations. A part of \( z_e \), denoted \( z_{\text{FACS}} \), enables interpretable rigging parameters, making NFR an artist-friendly tool for automatic rigging and expression retargeting.
Reprinted from \cite{10.1145/3588432.3591556}.
}
\end{figure}

Neural Face Rigging (NFR) \cite{10.1145/3588432.3591556} automates 3D mesh rigging, encoding interpretable deformation parameters aligned with ICT \cite{Li2020LearningFO} and Multiface \cite{wuu2023multifacedatasetneuralface}, enabling fine-grained facial expression transfer. More details of the NFR architecture can be found in Figure \ref{fig:nfr}.
MagicPose \cite{chang2024magicposerealistichumanposes} leverages diffusion models for 2D facial expression retargeting, balancing identity preservation and motion control through Multi-Source Attention and Pose ControlNet. Compared to DreamPose \cite{karras2023dreamposefashionimagetovideosynthesis} and Disco \cite{wang2023disco}, it excels in identity retention and generalization but struggles with extreme expressions.

DiffSHEG \cite{chen2024diffshegdiffusionbasedapproachrealtime} pioneers joint 3D facial expression and gesture synthesis, enforcing speech-driven alignment via uni-directional conditioning. It introduces Fast Out-Painting-based Partial Autoregressive Sampling (FOPPAS) for seamless, real-time motion generation. Compared to TalkSHOW \cite{Yi2022GeneratingH3}, LS3DCG \cite{10.1145/3472306.3478335}, and DiffuseStyleGesture \cite{ijcai2023p650}, it achieves higher realism and synchronization, though boundary inconsistencies persist in long sequences.

\subsection{Application}
Advancements in facial expression modeling have enabled a wide range of applications across industries. Realistic facial animations enhance immersion in gaming by making non-playable characters (NPCs) more expressive and emotionally engaging. Similarly, virtual reality (VR) benefits from these technologies by enabling personalized, synchronized 3D avatars. Models like AdaMesh \cite{chen2024adameshpersonalizedfacialexpressions} and DiffSHEG \cite{chen2024diffshegdiffusionbasedapproachrealtime} facilitate expressive avatar animation, though real-time performance remains a challenge due to computational constraints.

In interactive media and animation, these technologies streamline digital human creation, reducing the need for manual animation while improving expressiveness. CSTalk \cite{liang2024cstalkcorrelationsupervisedspeechdriven} and FaceDiffuser \cite{10.1145/3623264.3624447} enable nuanced emotional representation for storytelling, metaverse content, and digital assistants. Furthermore, real-time generation methods, such as DiffSHEG’s FOPPAS \cite{chen2024diffshegdiffusionbasedapproachrealtime}, demonstrate the potential for live-streamed virtual performances by ensuring smooth and adaptive facial animation. However, achieving consistent quality across different platforms and balancing computational efficiency with expressiveness remain key challenges.

Beyond entertainment, facial expression models are increasingly applied in healthcare and education. Expression recognition systems support therapeutic applications and adaptive learning environments \cite{9674818, 10.1093/bioinformatics/btae239}. However, privacy concerns, demographic biases, and the need for robust data protection necessitate further research. Future efforts should focus on expanding emotion representations and improving model scalability to meet the growing demands across these domains while ensuring ethical deployment.
    
\section{Image}\label{sec_image}
Image generation using deep learning models has garnered significant attention in recent years. Advances in this field have led to the development of powerful models such as GANs \cite{NIPS2014_f033ed80} and diffusion models \cite{NEURIPS2020_4c5bcfec}, which can produce highly realistic and creative images. These models have numerous applications in digital art \cite{li2024realtimegen}, design \cite{paananen2024using}, and augmented reality \cite{zhao2018compensation}. More importantly, in animation, image generation models play a crucial role in creating character visuals, synthesizing facial expressions, and rendering textures \cite{jing2022image}. For instance, diffusion models can generate high-quality facial images that serve as references for character design \cite{Kim_2023_CVPR}. In contrast, GAN-based models can create diverse character poses and aid in motion synthesis \cite{hinz2022charactergan}.
These techniques significantly reduce the time and cost required for creating visual assets in the animation production pipeline.

\subsection{Dataset}

\begin{table*}[t]
\centering
\renewcommand{\arraystretch}{1.4}
\resizebox{\textwidth}{!}{%
    \begin{tabular}{|>{\centering\arraybackslash}p{7cm}|p{7cm}|P{4cm}|c|}
\hline
\multicolumn{1}{|c|}{\textbf{Name}} & \multicolumn{1}{c|}{\textbf{Statistics}} & \multicolumn{1}{c|}{\textbf{Modalities}} & \multicolumn{1}{c|}{\textbf{Link}} \\ \hline
LAION-5B \cite{schuhmann2022laion} & 5,85 billion CLIP-filtered image-text pairs & Images, Text & \href{https://laion.ai/blog/laion-5b/}{LAION-5B} \\ \hline
LAION-400M \cite{schuhmann2021laion} & 400M English (image, text) pairs & Images, Text & \href{https://laion.ai/blog/laion-400-open-dataset/}{LAION-400M} \\ \hline
LAION-Aesthetics v2 \cite{schuhmann2022laion} & 1,2B aesthetics scores of $\geq$4.5\newline939M aesthetics scores of $\geq$4.75\newline600M aesthetics scores of $\geq$5\newline12M aesthetics scores of $\geq$6\newline3M aesthetics scores of $\geq$6.25\newline625K aesthetics scores of $\geq$6.5 & Images, Text & \href{https://laion.ai/blog/laion-aesthetics/}{LAION-Aesthetics v2} \\ \hline
Open Images V7 \cite{OpenImages} & 9M images annotated with image-level labels, object bounding boxes, object segmentation masks, visual relationships, and localized narratives & Images & \href{https://storage.googleapis.com/openimages/web/index.html}{Open Images V7} \\ \hline
COYO \cite{kakaobrain2022coyo-700m} & 747M image-text pairs & Images, Text & \href{https://github.com/kakaobrain/coyo-dataset}{COYO} \\ \hline
Conceptual Captions \cite{sharma-etal-2018-conceptual} & 3.3M images annotated with captions & Images, Text & \href{https://ai.google.com/research/ConceptualCaptions/}{Conceptual Captions} \\ \hline
COCO \cite{lin2014microsoft} & 330K images (>200K labeled)\newline1.5 million object instances\newline80 object categories\newline91 stuff categories\newline5 captions per image\newline250,000 people with key points & Images, Text & \href{https://cocodataset.org}{COCO} \\ \hline
ShareGPT \cite{chen2025sharegpt4v} & 100k highly descriptive image-caption & Images, Text & \href{https://huggingface.co/datasets/Lin-Chen/ShareGPT4V}{ShareGPT} \\ \hline
ADE20K \cite{zhou2019semantic} & 20,210 images in the training set\newline2,000 images in the validation set\newline3,000 images in the testing set & Images & \href{https://groups.csail.mit.edu/vision/datasets/ADE20K/}{ADE20K} \\ \hline
    \end{tabular}%
}
\caption{Overview of various image datasets, including their main statistical features, modalities, and link.}
\label{image-dataset-table}
\end{table*}

Image datasets can be categorized into different types based on their applications, primarily falling into two main groups: (i) image generation datasets and (ii) image editing datasets.

Image generation datasets typically include images, descriptions, and, in some cases, control conditions \cite{zhang2023adding}. Most research relies on widely used public datasets such as LAION \cite{schuhmann2022laion} and COCO \cite{lin2014microsoft}, though other sources like Unsplash \cite{unsplash} and Pixabay \cite{pixabay} are sometimes used for data collection \cite{kawar2023imagic}.

Image editing datasets contain the original and edited images and a description of the intended modification. Since maintaining similarity between pre- and post-edit images is crucial while ensuring alignment with the textual description, creative techniques are often employed to construct such datasets. A common approach involves leveraging a large language model (LLM) and a text-to-image model \cite{brooks2023instructpix2pix}. First, the LLM generates captions for the input image, edit instructions, and output captions reflecting the modifications. Then, a pre-trained text-to-image model converts the caption pairs (describing pre-edit and post-edit images) into corresponding image pairs \cite{brooks2023instructpix2pix}. Table \ref{image-dataset-table} presents an overview of widely used image generation and editing datasets.

\subsection{Evaluation}
In evaluating image generation models, a comprehensive set of quantitative and qualitative metrics is used to assess their quality and performance. Quantitative metrics such as \textit{Fréchet Inception Distance (FID)}, \textit{Learned Perceptual Image Patch Similarity (LPIPS)}, and \textit{Kernel Inception Distance (KID)} are widely employed to evaluate the distributional similarity of generated images to real images. These metrics compare images' high- and low-level features to measure their statistical alignment. Additionally, metrics like \textit{CLIP Score} are used to assess the semantic alignment between the image and the input text \cite{gandikota2024unified, brooks2023instructpix2pix}. \textit{PSNR (Peak Signal-to-Noise Ratio)} \cite{wang2004image} and \textit{SSIM (Structural Similarity Index)} \cite{wangzhou2004image} are also used to evaluate image reconstruction quality based on the level of detail preservation and noise in the output \cite{mokady2023null}. The results of these evaluations are compared with the best available models to analyze the model's performance more accurately.

Alongside these quantitative metrics, qualitative evaluations also play a significant role in assessing the perceptual quality of images. In many studies, human evaluation examines the generated content's visual realism and semantic consistency with the text or initial conditions. For example, in studies utilizing Amazon Mechanical Turk or similar platforms for human surveys, participants have assessed the realism, aesthetics, and semantic alignment of the images, such as SVDiff \cite{han2023svdiff} and Imagic \cite{kawar2023imagic}. These methods are highly effective for evaluating metrics such as visual similarity, content coherence, and overall output quality, as the ultimate goal of many image generation models is to create images that appear natural and meaningful to humans \cite{zhang2023adding, kawar2023imagic}.
In addition to qualitative and quantitative metrics, some studies have examined the impact of model parameter variations on final performance. For instance, changing hyperparameters such as learning rate, batch size, or the number of sampling steps can significantly affect the output quality \cite{li2024sketch, wu2023visual, kawar2023imagic}.

\begin{figure}[t]
    \centering
    \includegraphics[width=\textwidth]{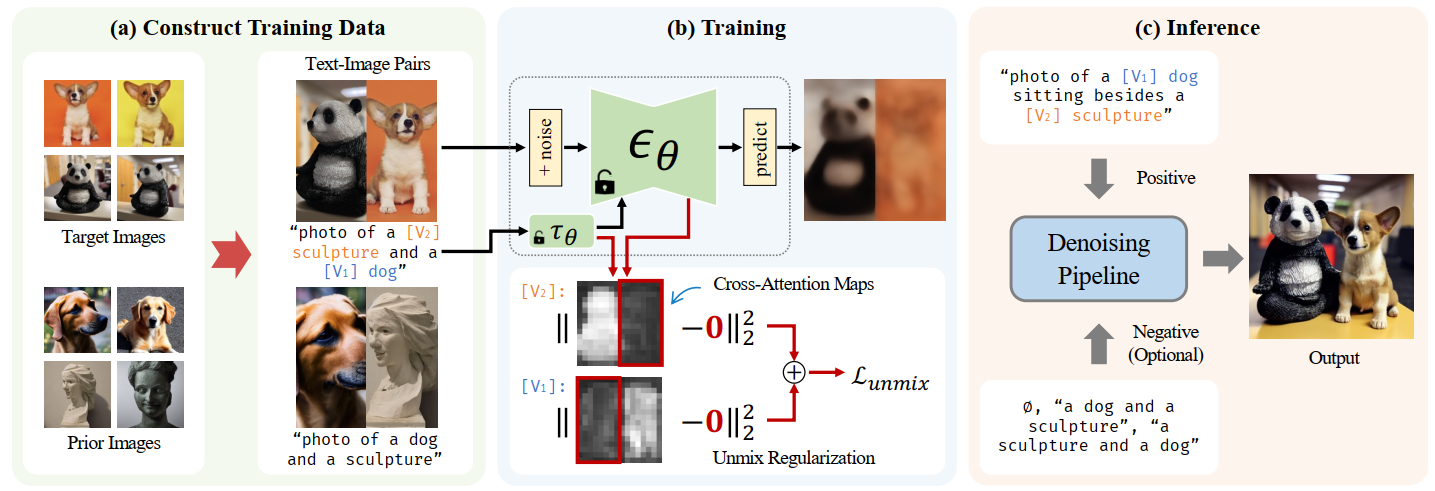}
    \caption{The Cut-Mix-Unmix data augmentation technique for multi-subject generation, implemented in SVDiff \cite{han2023svdiff}. The Cut-Mix-Unmix method is a data augmentation technique designed to train image generation models to handle multiple subjects. By creating composite images and corresponding textual prompts, the method teaches the model to distinguish between different concepts. Through Unmix regularization and applying MSE to the attention maps, the model is encouraged to separate better the subjects (e.g., a dog and a panda).
    Reprinted from \cite{han2023svdiff}.
    }
    \label{fig:cut_mix_unmix}
\end{figure}

Another approach to evaluating the outputs of generative models involves comparing different models based on factors such as image quality, resource consumption, and efficiency \cite{karadaug2023transforming, he2024generating, wu2023uncovering, gao2023llama, yang2023mm, zhang2023sine, kawar2023imagic}. In this method, models are compared against each other to identify their strengths and weaknesses. Furthermore, in some studies, models are evaluated under different conditions and on diverse datasets to measure their generalization capabilities. Well-known benchmarks such as MS-COCO \cite{lin2014microsoft}, DreamBench \cite{ruiz2023dreambooth}, TEdBench \cite{kawar2023imagic} and LAION-5B \cite{schuhmann2022laion} are widely used for evaluating image generation models \cite{pan2023kosmos, saharia2022photorealistic, han2023svdiff}.

The evaluation of image generation models is a multidimensional process combining quantitative metrics and qualitative assessments to understand a model’s capabilities comprehensively. Metrics such as \textit{FID}, \textit{LPIPS}, \textit{CLIP Score}, \textit{PSNR}, and \textit{SSIM} offer valuable insights into the statistical, perceptual, and semantic aspects of image quality. At the same time, human evaluations provide a more accurate reflection of user satisfaction in real-world scenarios. Recent studies, such as KOSMOS-G \cite{pan2023kosmos}, Imagic \cite{kawar2023imagic} and ControlNet \cite{zhang2023adding}, through comparisons with other models, analysis of the metrics above, and human assessments, have shown that their proposed methods are better at preserving the semantic content of images and delivering higher quality outputs.

\subsection{Models}

\subsubsection{Image Generation}\label{image_image_generation}
Diffusion models have achieved remarkable success in text-to-image generation, enabling the creation of high-quality images from text prompts or other modalities. Using the most advanced techniques in deep learning, these models have generated high-quality images with precise alignment to text descriptions. In this regard, models like Imagen  \cite{saharia2022photorealistic} and SDXL \cite{podell2023sdxl} have significantly improved the accuracy and quality of text-to-image generation. Imagen \cite{saharia2022photorealistic} builds on the power of large transformer language models in understanding text and hinges on the strength of diffusion models in high-fidelity image generation \cite{saharia2022photorealistic}. SDXL \cite{podell2023sdxl} is also a latent diffusion model for text-to-image synthesis. Compared to previous versions of Stable Diffusion \cite{rombach-high-res}, SDXL leverages a three times larger U-Net backbone \cite{DBLP:journals/corr/RonnebergerFB15}. 

In parallel with the advancements in high-quality text-to-image generation through diffusion models, there is an increasing demand for more efficient customization techniques. Existing methods for customizing diffusion models face several challenges, especially when dealing with multiple personalized subjects and minimizing the risk of overfitting. Additionally, the large number of parameters in these models leads to inefficiencies in storage. SVDiff \cite{han2023svdiff} has been proposed to address these limitations by introducing a compact parameter space called spectral shift for diffusion model fine-tuning. The Cut Mix-Unmix data augmentation technique further enhances the quality of multi-subject generation, allowing for handling similar categories. The way this mechanism works is illustrated in Figure \ref{fig:cut_mix_unmix}.

Despite these advances, text-to-image models remain limited in controlling spatial composition. ControlNet \cite{zhang2023adding} addresses this by locking large pre-trained diffusion models and reusing their deep, robust encoding layers (trained on billions of images) as a strong backbone for learning a diverse set of conditional controls. Extensive results show that ControlNet \cite{zhang2023adding} may facilitate broader applications in controlling image diffusion models.

\subsubsection{Image Editing}\label{image_image_editing}

\begin{figure}[t]
    \centering
    \includegraphics[width=0.8\textwidth]{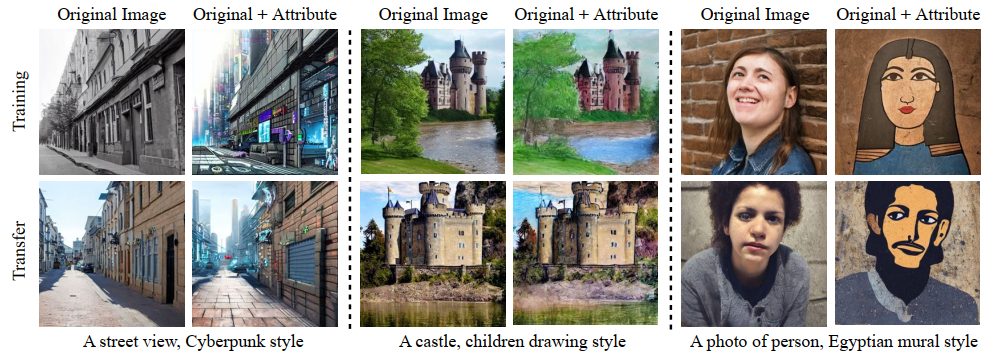}
    \caption{
    Examples of image editing using the DiffusionDisentanglement model \cite{wu2023uncovering}. Each row displays a text description combining style-neutral content and target attribute descriptions (separated by commas). For each attribute section, the first row shows results on optimization images, while the second row demonstrates transfer to unseen images. The left column contains source images, and the right column shows the corresponding modified images.
    Reprinted from \cite{wu2023uncovering}.
    }
    \label{fig:disentanglement_example}
\end{figure}

In addition to image generation, developing models for image editing through text input is significant. Various models and approaches have been proposed in this area, each aiming to improve the accuracy and quality of editing with a different strategy.
DiffusionDisentanglement \cite{wu2023uncovering} demonstrates that Stable Diffusion models inherently can disentangle style and content. This property can be activated by partially replacing text embeddings, and with only 50 parameters optimized, it outperforms more complex methods. An example of this model’s capability in editing images is shown in Figure \ref{fig:disentanglement_example}. Similarly, InstructPix2Pix \cite{brooks2023instructpix2pix} tackles image editing by creating a paired dataset of “instruction-image” samples and training a supervised model based on Stable Diffusion, achieving success even in challenging edits.

Building on these efforts, several models have been designed to address specific limitations in different stages of the editing process. SINE \cite{zhang2023sine} innovatively employs classifier-free guidance by both modifying the guidance mechanism and injecting content seeds in the early denoising steps. In the same direction, Null-text-Inversion \cite{mokady2023null} enables high-fidelity reconstructions without re-training the model or conditional embeddings, simply by optimizing the unconditional (null) embedding.

To achieve a better balance between fidelity to the original image and alignment with the target text, Imagic \cite{kawar2023imagic} combines text embedding optimization, model fine-tuning, and linear interpolation between embeddings. On a broader level, Unified Concept Editing (UCE) \cite{gandikota2024unified} offers an integrated approach for targeted editing of diffusion models, allowing the removal of bias, offensive content, or copyrighted material using only textual descriptions, without compromising the core concepts encoded in the model.

\subsubsection{Visual Language Models}\label{image_visual_language_models}

In addition to text-based image generation and editing advancements, significant efforts have also been made in multimodal interaction with images and text. In this direction, models are not only capable of understanding and generating visual content but can also engage in multimodal dialogue, reason over inputs, and provide coherent responses aligned with both textual and visual information.

\begin{figure}[t]
    \centering
    \includegraphics[width=0.6\textwidth]{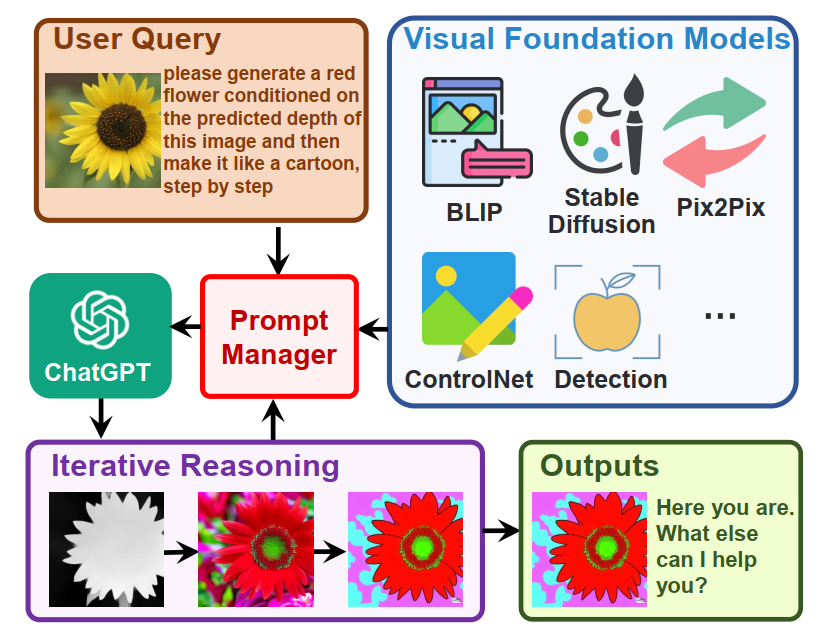}
    \caption{Visual ChatGPT architecture \cite{wu2023visual} showing the workflow from user query to output. The system processes complex instructions on a flower image through a prompt manager that coordinates visual foundation models (BLIP, Stable Diffusion, ControlNet, etc.). The example demonstrates iterative reasoning through depth estimation and style transfer to transform a yellow flower into a red cartoon version. Reprinted from \cite{wu2023visual}.
    }
    \label{fig:visualgpt}
\end{figure}

In this context, models such as Visual ChatGPT \cite{wu2023visual} combine ChatGPT with various vision models to enable responses to complex visual queries. This model utilizes a prompt manager and chain-of-thought reasoning to break down user requests into manageable tasks for vision modules. The architecture of this model is shown in Figure \ref{fig:visualgpt}. Similarly, KOSMOS-G \cite{pan2023kosmos} establishes a connection between interleaved inputs and the image decoder using its multimodal large language model and its proposed AlignerNet component, enabling seamless concept-level guidance. This novel AlignerNet is a crucial bridge, specifically trained to address the misalignment between the MLLM's output space and the input space required by a frozen image decoder (like Stable Diffusion's U-Net). This alignment, achieved using supervision from CLIP's text encoder, allows the MLLM to effectively condition the image generation process without modifying the decoder. Building on this line of work, MM-REACT \cite{yang2023mm} introduces a multimodal reasoning-and-action framework that has demonstrated its effectiveness in complex tasks such as multi-image reasoning, multi-hop document understanding, and open-world concept comprehension.

\subsection{Application}
Image generation has been one of the most extensively studied research problems in computer vision, with many competitive generative models introduced over the past decade. Recently, new image generation methods, capable of producing high-quality and high-resolution images in various domains, have garnered significant attention in research \cite{yang2023mm}. These models can generate nearly any image by simply feeding in a relevant text, thereby dramatically transforming the landscape of artistic applications \cite{mokady2023null}.

Image generation technologies have found extensive applications across various fields in recent years. In the entertainment and media industry \cite{hasan2024applications}, these technologies have played a pivotal role in creating realistic visual content, including designing digital characters \cite{avrahami2024chosen} and simulated environments. Similarly, in design and architecture, image generation models have enhanced the process of visualizing designs and presenting architectural concepts by simulating photorealistic images \cite{karadaug2023transforming}.

Specifically, in the animation industry, image generation models have accelerated the concept art process and the creation of complex backgrounds \cite{brisco2023exploring, chung2002progressive}, boosting productivity and reducing production time. Generative AI models are also used for enhancing visual effects, refining character designs, and streamlining texture generation \cite{long2024sketchar, yin2024innovative}.
Furthermore, in the fields of e-commerce and marketing, synthetic image generation has significantly contributed to product personalization and improved user experiences \cite{chen2024virtualmodel}. In the domains of medicine \cite{nguyen2023new} and education, these models have enabled synthetic data generation for training and scientific research. Additionally, in social networks and metaverse spaces, these technologies have provided tools for creating digital identities and facilitating interactions in virtual environments.

With the rise of powerful large language models (LLMs) and growing attention to this area, image generation models have been integrated with LLMs to effectively align visual and textual modalities, enabling better comprehension of human instructions \cite{li2024enhanced}.
In this context, models equipped with image generation and understanding capabilities have proven highly effective in areas such as mathematical and textual reasoning, understanding visual jokes and memes, spatial and coordinate comprehension, visual planning and prediction, multi-image reasoning, multi-step document understanding from charts, floor plans, flowcharts, tables, open-world concept comprehension and video analysis and summarization \cite{yang2023mm}.

Text-based image editing has also recently gained substantial attention. New image generation models have been utilized in diverse applications such as image reconstruction, adversarial refinement, image compression, image classification, and many others \cite{kawar2023imagic}.
Remarkable advancements in image generation have expanded the boundaries of technical capabilities and redefined how humans interact with visual and textual data. These technologies can potentially bring about profound transformations across various industries by providing innovative tools for creativity, analysis, and interaction.

\section{Avatar}\label{sec_avatar}
Avatar generation in generative AI focuses on synthesizing digital representations of individuals or characters using advanced machine learning techniques. Recent progress in deep learning architectures and large-scale multimodal datasets has driven significant advancements in this domain. Modern methods achieve visually realistic and behaviorally expressive avatars by seamlessly integrating a variety of modalities, including text, images, video, audio, and motion data. These developments support virtual reality, gaming, film production, and digital communication applications.

Key tasks in avatar generation include detailed 3D body reconstruction, expressive facial animation, and motion synthesis for dynamic interactions. Parametric models like SMPL (Skinned Multi-Person Linear Model) \cite{SMPL} and FLAME (Faces Learned with an Articulated Model) \cite{flame} serve as foundational frameworks, offering compact and interpretable representations of human geometry and motion. These models facilitate precise pose estimation, motion retargeting, and animation synthesis. Furthermore, advancements in multimodal fusion have allowed avatars to go beyond just replicating physical appearance; they can now convey expressive behaviors like synchronized speech and gestures, greatly enhancing their realism and interactivity.

\subsection{Dataset}
Developing large-scale and diverse datasets has been instrumental in advancing avatar generation, providing the necessary training data for realistic and generalizable models. These datasets capture various human appearances, motions, and interactions through multimodal inputs. While video and image data form the visual foundation, 3D body meshes, SMPL parameters, and textured meshes enable detailed geometric and motion representations. Additionally, audio and textual annotations enrich these datasets by incorporating behavioral contexts, such as speech-driven expressions or descriptive cues for stylized synthesis.

A notable example is WildAvatar \cite{huang2024wildavatarwebscaleinthewildvideo}, a large-scale dataset featuring over 10,000 individuals sourced from in-the-wild YouTube videos. Integrating video, 3D body motion, and audio modalities bridges the gap between controlled laboratory datasets and the diverse real-world scenarios essential for generalizable avatar generation. Similarly, RenderMe-360 \cite{2023renderme360} marks a milestone in 3D avatar creation, capturing over 243 million head frames from 500 identities using a 60-camera system. With rich annotations, including FLAME parameters, UV maps, and action units, it supports high-fidelity reconstruction and animation.

Datasets such as HuMMan \cite{10.1007/978-3-031-20071-7_33} and AMASS \cite{AMASS} further highlight the importance of multimodal integration. HuMMan provides video, point clouds, and SMPL parameters for 1,000 subjects, supporting applications ranging from action recognition to parametric recovery. Conversely, AMASS unifies 15 motion capture datasets, offering over 42 hours of motion data annotated with SMPL parameters, making it a foundational resource for motion dynamics and retargeting research.

The availability of these datasets has driven the development of sophisticated models capable of handling key challenges in avatar generation, such as dynamic motion synthesis and expressive behavior modeling. Including body meshes and SMPL parameters enables precise motion analysis, while point clouds and textured meshes capture fine-grained details crucial for photorealistic reconstruction. As the demand for high-fidelity and adaptable avatars grows, dataset expansion remains critical in advancing the field. A comprehensive overview of these and other influential datasets is provided in Table \ref{table:avatar}.

\begin{table*}[t!]
\centering
\renewcommand{\arraystretch}{1.4}
\resizebox{\textwidth}{!}{%
    \begin{tabular}{|>{\centering\arraybackslash}p{7cm}|p{7cm}|P{4cm}|c|}
\hline
\multicolumn{1}{|c|}{\textbf{Name}}  & \multicolumn{1}{c|}{\textbf{Statistics}} & \multicolumn{1}{c|}{\textbf{Modalities}} & \multicolumn{1}{c|}{\textbf{Link}} \\ \hline

WildAvatar \cite{huang2024wildavatarwebscaleinthewildvideo} & 
Over 10,000 human subjects; extracted from YouTube; significantly richer than previous datasets for 3D human avatar creation & 
Video, 3D/Point Cloud Data, Audio & 
\href{https://wildavatar.github.io/}{WildAvatar} \\ \hline

ZJU-MoCap \cite{peng2021neural} & 
Multi-camera system with 20+ synchronized cameras; includes SMPL-X parameters for detailed motion capture of body, hand, and face; complex actions such as twirling, Taichi, and punching & 
Video, 3D/Point Cloud Data & 
\href{https://chingswy.github.io/Dataset-Demo/}{ZJU-MoCap} \\ \hline

TalkSHOW \cite{Yi2022GeneratingH3} & 
26.9 hours of in-the-wild talking videos from 4 speakers; expressive 3D whole-body meshes reconstructed at 30 fps, synchronized with audio at 22 kHz & 
Audio, 3D/Point Cloud Data & 
\href{https://talkshow.is.tue.mpg.de/}{TalkSHOW} \\ \hline

HuMMan \cite{10.1007/978-3-031-20071-7_33} & 
1,000 human subjects, 400k sequences, 60M frames; include point clouds, SMPL parameters, and textured meshes for multimodal sensing & 
Video, 3D/Point Cloud Data & 
\href{https://caizhongang.com/projects/HuMMan/}{HuMMan} \\ \hline

BUFF \cite{Zhang_2017_CVPR} & 
6 subjects performing motions in two clothing styles; 13,632 3D scans with high-resolution ground-truth minimally-clothed shapes & 
3D/Point Cloud Data & 
\href{https://buff.is.tue.mpg.de/}{BUFF} \\ \hline

AMASS \cite{AMASS} & 
Combines 15 motion capture datasets into a unified framework with over 42 hours of motion data; 346 subjects and 11,451 motions with SMPL pose parameters, 3D shape parameters, and soft-tissue coefficients & 
3D/Point Cloud Data & 
\href{https://amass.is.tue.mpg.de/}{AMASS} \\ \hline

3DPW \cite{3DPW} & 
60 video sequences with accurate 3D poses using video and IMU data; 18 re-poseable 3D body models with different clothing variations & 
Video, Time-Series Data, 3D/Point Cloud Data & 
\href{https://virtualhumans.mpi-inf.mpg.de/3DPW/}{3DPW} \\ \hline

AIST++ \cite{aist++} & 
10,108,015 frames of 3D key points with corresponding images; 1,408 dance motion sequences spanning 10 dance genres with synchronized music & 
Video, Audio, 3D/Point Cloud Data & 
\href{https://google.github.io/aistplusplus_dataset/}{AIST++} \\ \hline

RenderMe-360 \cite{2023renderme360} & 
Over 243 million head frames from 500 identities; includes FLAME parameters, UV maps, action units, textured meshes, and diverse annotations & 
Video, 3D/Point Cloud Data & 
\href{https://renderme-360.github.io/}{RenderMe-360} \\ \hline

PuzzleIOI \cite{xiu2024puzzleavatar} & 
41 subjects with nearly 1,000 Outfit-of-the-Day (OOTD) configurations; includes paired ground-truth 3D body scans for challenging partial photos & 
Images, 3D/Point Cloud Data, Text & 
\href{https://puzzleavatar.is.tue.mpg.de/}{PuzzleIOI} \\ \hline

    \end{tabular}
}
\caption{Overview of various datasets related to avatar generation, including their main statistical features, modalities, and links.}
\label{table:avatar}
\end{table*}

\subsection{Evaluation}
Evaluating avatar generation models involves assessing multiple dimensions, including geometry accuracy, texture fidelity, identity preservation, temporal consistency, and animation robustness. These evaluations rely on quantitative metrics, qualitative assessments, and ablation studies to comprehensively benchmark performance.

\begin{figure}[t!]
\begin{center}
  \includegraphics[width=\textwidth]{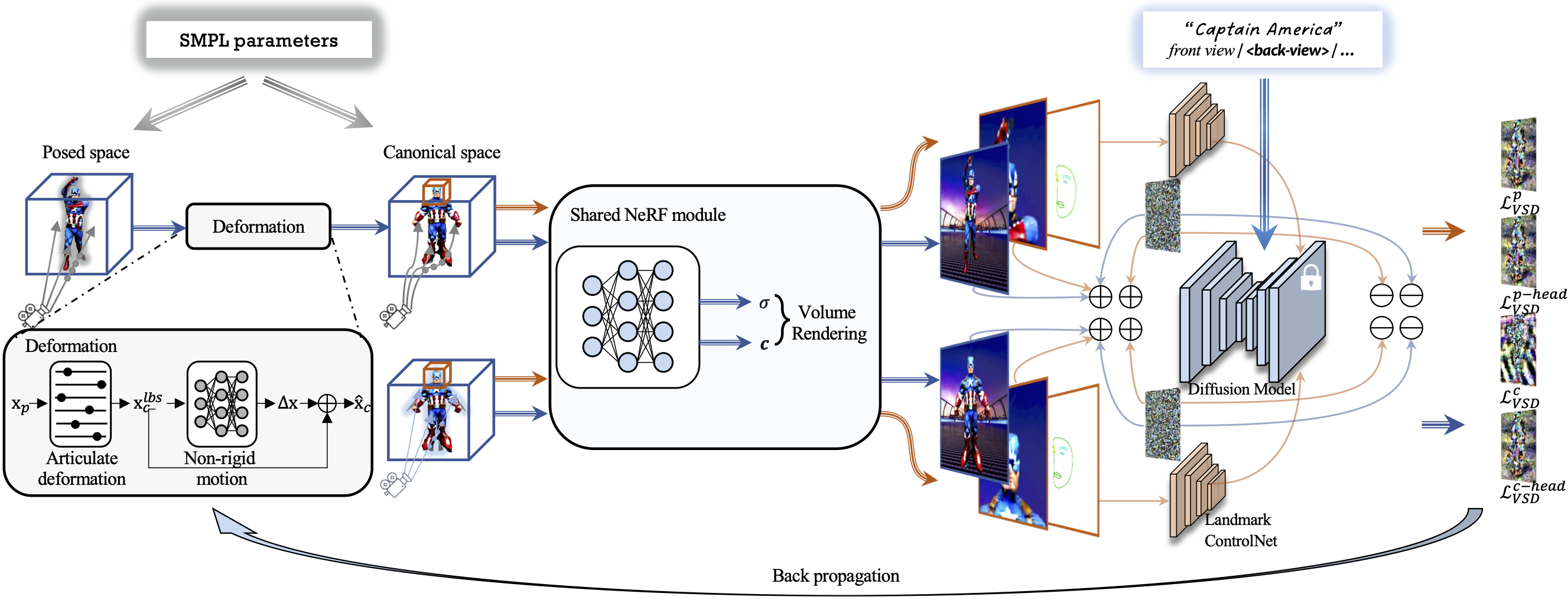}
\end{center}
\caption{\label{fig:dreamavatar}
DreamAvatar's dual-observation-space architecture shows how text prompts and SMPL parameters drive avatar generation through canonical and posed spaces. The system uses an SMPL-based deformation field (left), shared NeRF module (center), and diffusion model guidance with landmark ControlNet for head refinement (right) to produce high-quality results with consistent facial features. 
Reprinted from \cite{cao2024dreamavatar}.
}
\end{figure}

High-quality geometry and texture are fundamental for realistic avatars. Common evaluation metrics include \textit{Fréchet Inception Distance (FID)} and \textit{Learned Perceptual Image Patch Similarity (LPIPS)} for measuring texture realism, while \textit{Landmark Mean Distance (LMD)} evaluates geometric accuracy. TADA \cite{liao2023tadatextanimatabledigital} achieves strong FID and LPIPS scores by ensuring geometric-texture consistency, particularly in facial regions. Make-Your-Anchor \cite{huang2024makeyouranchor} similarly demonstrates high-quality geometry and texture, validated through FID and LMD.
Temporal consistency ensures stable and artifact-free animations. Metrics such as \textit{Fréchet Video Distance (FVD)} and LMD measure motion stability across frames. Make-Your-Anchor \cite{huang2024makeyouranchor} improves temporal coherence through batch-overlapped temporal denoising and a two-stage training strategy, reducing motion artifacts.

Maintaining identity across transformations is critical for personalized avatars. \textit{CLIP Text-Image Direction Similarity (CLIP-S)} and \textit{CLIP Direction Consistency (CLIP-C)} assess identity retention. GaussianAvatar-Editor \cite{liu2025gaussianavatar} enhances identity preservation through occlusion-aware rendering and adversarial learning. Animation robustness focuses on the ability of avatars to perform realistic pose-dependent deformations and maintain semantic consistency during dynamic sequences. Metrics such as \textit{Animation Compatibility}, as introduced by TADA \cite{liao2023tadatextanimatabledigital}, evaluate the alignment of geometry and texture during motion. Ablation studies from DreamWaltz \cite{huang2023dreamwaltz} underscore the importance of density weighting and two-stage training in generating realistic and detailed motion dynamics.

Qualitative assessments and user studies are pivotal in evaluating the perceptual aspects of realism that cannot be captured by quantitative metrics alone. DreamWaltz \cite{huang2023dreamwaltz} outperforms DreamAvatar \cite{cao2024dreamavatar} and AvatarCraft \cite{jiang2023avatarcraft} in user preference studies based on geometry and texture quality. AvatarVerse \cite{zhang2023avatarverse} achieves an 85\% preference rate over DreamFusion \cite{poole2022dreamfusion}, DreamAvatar, DreamWaltz, and DreamHuman \cite{kolotouros2023dreamhuman}, demonstrating superior texture fidelity.
Ablation studies provide insights into model design. Make-Your-Anchor \cite{huang2024makeyouranchor} shows that temporal denoising significantly enhances facial details and motion stability. GaussianAvatar-Editor \cite{liu2025gaussianavatar} demonstrates how Weighted Alpha Blending reduces occlusion artifacts, improving CLIP-S scores.

Benchmark comparisons reveal various avatar generation models' unique strengths and limitations, providing valuable insights into state-of-the-art advancements. These studies demonstrate that no single model excels across all evaluation metrics; instead, each exhibits specific strengths and trade-offs. 
Earlier models such as DreamFusion \cite{poole2022dreamfusion}, AvatarCLIP \cite{10.1145/3528223.3530094}, Latent-NeRF \cite{metzer2022latent}, and Text2Mesh \cite{text2mesh} laid the foundation for avatar synthesis but lack the robustness and flexibility of recent approaches.
More advanced approaches reflect progress in addressing geometry, texture realism, motion consistency, and identity preservation challenges, collectively shaping the future of avatar generation.
For instance, DreamWaltz \cite{huang2023dreamwaltz} showcases exceptional geometry and texture fidelity, particularly for complex avatars with intricate clothing details, while AvatarVerse \cite{zhang2023avatarverse} excels in generating high-quality, fine-grained textures and maintaining stability across arbitrary poses. DreamHuman \cite{kolotouros2023dreamhuman} achieves diverse and realistic avatar animations, addressing pose-dependent deformations effectively. TADA \cite{liao2023tadatextanimatabledigital} stands out for its animation compatibility and semantic alignment, offering superior control during motion sequences.

\subsection{Models}
Generating realistic and controllable 3D avatars has become a prominent research area. Various generative models have been proposed to address the challenges of creating high-fidelity avatars with diverse appearances, motions, and styles. These models can be broadly categorized based on their underlying techniques.

\subsubsection{CLIP-Guided Models} \label{avatar_CLIP-guided_models}
Models leveraging CLIP \cite{DBLP:conf/icml/RadfordKHRGASAM21} have revolutionized avatar generation by enabling text-driven approaches. AvatarCLIP \cite{10.1145/3528223.3530094} is a prime example, offering a zero-shot framework for generating and animating 3D avatars from natural language descriptions. This pipeline translates text into static and animatable 3D avatars, employing a shape VAE for initial geometry generation guided by CLIP to ensure textual alignment. NeuS (Neural Implicit Surface) \cite{10.5555/3540261.3542342} is also integrated for high-quality geometry and photorealistic rendering. The motion generation phase involves selecting candidate poses from a precomputed codebook using CLIP and synthesizing smooth motions via a motion VAE.  AvatarCLIP demonstrates superior geometry, vivid textures, and animation consistency compared to models like DreamField \cite{jain2021dreamfields} and Text2Mesh \cite{text2mesh}. DreamField \cite{jain2021dreamfields} adapts NeRF for text-driven 3D object generation but struggles with detailed geometry, while Text2Mesh \cite{text2mesh} stylizes existing meshes using CLIP guidance but faces stability and flexibility challenges with diverse text descriptions.

\subsubsection{Implicit Function-Based Models}\label{avatar_implicit_function-based_models}
Implicit function-based models have played a significant role in achieving high-resolution, detailed 3D avatar reconstructions by leveraging continuous surface representations. These methods have evolved from single-view reconstruction to more sophisticated, animatable avatars with fine-grained control. Early work, such as PIFu (Pixel-Aligned Implicit Function) \cite{saito2019pifu}, introduced pixel-aligned feature extraction for reconstructing 3D surfaces from single-view 2D images. By projecting 3D points into the image space and retrieving features via a CNN, PIFu enabled high-resolution surface reconstruction. Its successor, PIFuHD \cite{saito2020pifuhd}, extended this approach with multi-scale feature extraction, enhancing both global shape understanding and fine surface details.

ARCH (Animatable Reconstruction of Clothed Humans) \cite{arch} introduced a canonical space transformation via a parametric body model to address pose variations and occlusions in human modeling. ARCH captured intricate surface details, including clothing folds, by learning an implicit function conditioned on pose-normalized coordinates. ARCH++ \cite{archpp} further improved geometry encoding and texture generation, refining the realism of reconstructed avatars. Expanding on parametric models, PaMIR (Parametric Model-Conditioned Implicit Representation) \cite{author2020paper1} integrated an SMPL body model into an implicit function framework. By refining depth predictions and SMPL parameters, PaMIR achieved closer alignment between inferred 3D surfaces and input images and reduced artifacts caused by depth ambiguities.

Recent advancements have focused on text-driven generation and improved articulation. TADA (Text to Animatable Dynamic Avatar) \cite{liao2023tadatextanimatabledigital} introduced SMPL-X models with learnable displacements, optimizing geometry and texture through Score Distillation Sampling losses to generate animatable avatars from text descriptions. GETAvatar (Generative Textured Meshes for Animatable Human Avatars) \cite{zhang2023getavatar} further advanced explicit mesh-based modeling by employing signed distance fields (SDFs) in a canonical space, deforming them via SMPL-based transformations to match body shapes and poses while leveraging a normal field trained on 3D scans for enhanced geometric detail. More recently, RodinHD \cite{zhang2024rodinhd} has emphasized high-resolution avatar creation from single portrait images, constructing detailed 3D blueprints through triplane representations and refining features with cascaded diffusion models to achieve superior texture and geometric fidelity. These developments reflect a broader evolution of implicit function-based models, transitioning from single-view reconstructions to high-fidelity, animatable avatars by integrating text-driven methods and neural rendering techniques for greater realism and control.

\subsubsection{NeRF-Based Methods}\label{avatar_NeRF-based_methods}
NeRF-based methods have emerged as a powerful approach for 3D human avatar modeling, addressing scenarios from static camera setups to dynamic environments. These methods generally employ deformation fields based on SMPL(-X) models to map points from observation to canonical space, incorporating articulated deformation and non-rigid motion correction.

\begin{figure}[t!]
\begin{center}
  \includegraphics[width=\textwidth]{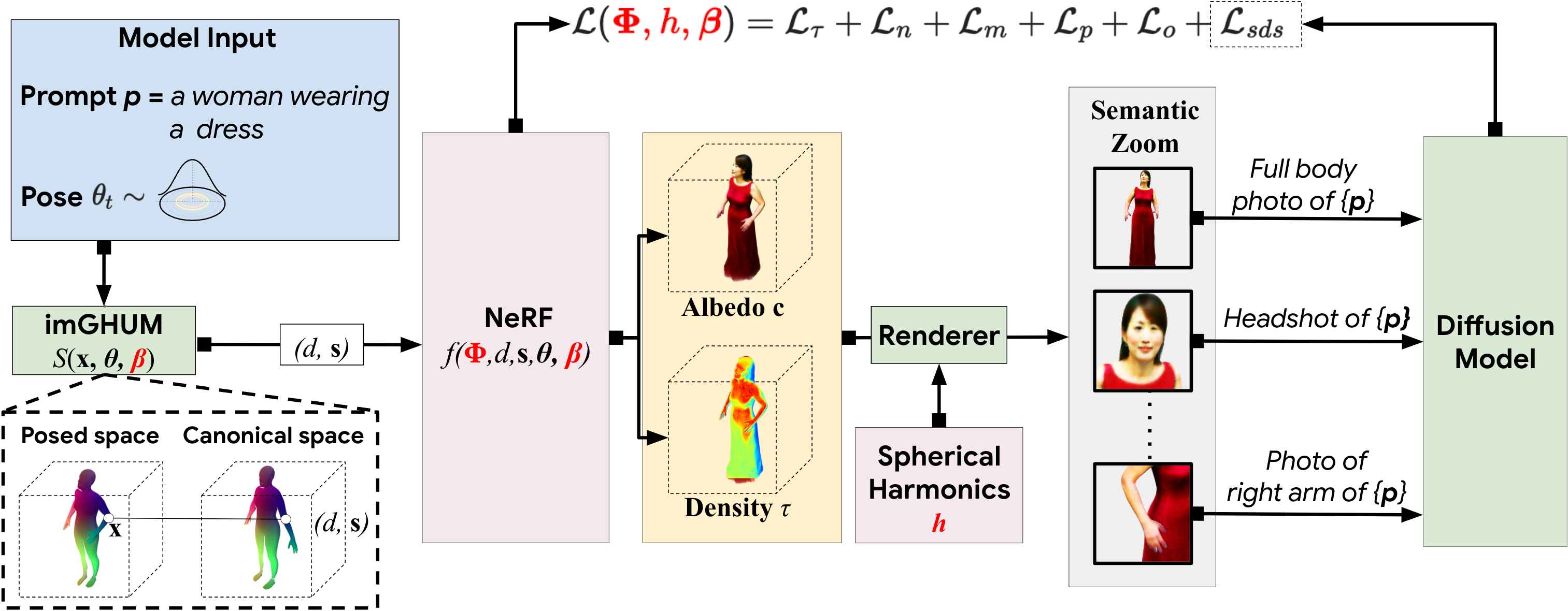}
\end{center}
\caption{\label{fig:dreamhuman}
Pipeline of DreamHuman \cite{kolotouros2023dreamhuman}: the model takes a text prompt \( p \) (e.g., \textit{a woman wearing a dress}) and generates a 3D animatable avatar using a deformable, pose-conditioned NeRF constrained by the imGHUM \cite{Alldieck2021imGHUMIG} body model. The pipeline incorporates semantic zooming for critical regions (face, hands) to improve detail and employs Score Distillation Sampling (SDS) \cite{poole2022dreamfusion} for optimization, producing high-fidelity avatars with pose control. 
Reprinted from \cite{kolotouros2023dreamhuman}.
}
\end{figure}

HumanNeRF \cite{weng2022humannerf} pioneered the application of deformation fields for dynamic human models from monocular images. Neural Body introduced structured latent codes anchored to SMPL model vertices, processed using SparseConvNet, to provide regularization. Neural Human Performer captured information directly in observation, utilizing a skeletal feature bank and transformers. Vid2Avatar \cite{guo2023vid2avatar} proposed jointly modeling the human and scene background using two separate neural radiance fields. DreamHuman \cite{kolotouros2023dreamhuman} as illustrated in Figure \ref{fig:dreamhuman}, generates animatable 3D human avatars from textual descriptions by combining NeRF with imGHUM \cite{Alldieck2021imGHUMIG}, using human body shape statistics to create anatomically correct avatars with realistic body proportions that deform naturally when posed.

\subsubsection{Diffusion-Based Methods}\label{avatar_diffusion-based_methods}
Diffusion-based methods have gained prominence in avatar generation, leveraging pre-trained 2D text-to-image diffusion models to guide the creation of high-quality 3D assets. Personalized Avatar Scene (PAS) \cite{azadi2023textconditionalcontextualizedavatarszeroshot} generates customized 3D avatars in various poses and scenes based on text descriptions, using a diffusion-based transformer model to generate 3D body poses from text. The model proposed in \cite{Bergman2023Articulated3H} employs a parametric 3D Morphable Model (3DMM) of the head (FLAME \cite{10.1145/3130800.3130813}) and combines it with diffusion models to optimize both geometry and texture for generating 3D head avatars directly from textual prompts. The Make-Your-Anchor system \cite{huang2024makeyouranchor} introduces a novel approach for generating 2D anchor-style avatars capable of realistic full-body motion and expression, utilizing a Structure-Guided Diffusion Model (SGDM).

\subsubsection{Hybrid Methods}\label{avatar_hybrid_methods}
Recent approaches combine multiple methodologies to overcome the limitations of single-technique models. DreamAvatar \cite{cao2024dreamavatar} exemplifies this trend by integrating shape priors, diffusion models, and NeRF architecture in a dual-observation-space (DOS) framework, as illustrated in Figure \ref{fig:dreamavatar}. By leveraging SMPL \cite{SMPL} for anatomical guidance through density fields, it ensures structurally consistent avatars while addressing geometry issues common in pure diffusion methods. The system's dual-space optimization works simultaneously in canonical and posed spaces, enabling complete texture generation with pose-faithful geometry. To solve the "Janus" problem (inconsistent facial features across views), it employs joint optimization with specialized head-focused VSD loss using ControlNet \cite{zhang2023adding}. While offering superior geometric accuracy and controllable shape modifications compared to methods like DreamWaltz \cite{huang2023dreamwaltz}, limitations include a lack of animation capabilities and inherited biases from pretrained diffusion models. As avatar generation advances, such hybrid approaches that strategically combine strengths from different paradigms represent a promising direction for future research.

\subsection{Application}
Advancements in avatar generation technologies have transformed industries such as gaming, virtual reality (VR), and entertainment. Realistic avatars with lifelike facial expressions, synchronized body movements, and real-time motion capture enhance immersion and interactivity, supported by large-scale datasets like WildAvatar \cite{huang2024wildavatarwebscaleinthewildvideo} and RenderMe-360 \cite{2023renderme360}. These advancements enrich storytelling, multiplayer experiences, and player engagement \cite{liu2024reportmethodsapplicationscrafting}.
In social media and the metaverse, avatars facilitate personalized digital identities through multimodal inputs such as text and speech. Generative models create photorealistic textures and natural motion, fostering richer interactions in virtual spaces. Similarly, e-commerce and education benefit from AI-driven avatars, which serve as virtual assistants, instructors, or customer support agents \cite{huang2024makeyouranchor}.

The entertainment industry leverages high-fidelity 3D models for digital humans in films and animations, reducing production costs while enhancing creative flexibility \cite{Zeng2023AvatarBoothHA}. Beyond media, healthcare employs avatars for therapy and rehabilitation, while education integrates them as adaptive virtual tutors \cite{liu2024reportmethodsapplicationscrafting}. However, real-time performance, scalability, and cross-platform compatibility must be addressed to meet growing demands. Furthermore, ethical concerns related to privacy, inclusivity, and deepfake misuse remain critical considerations.

\section{Gesture}\label{sec_gesture}
Gesture generation is a vital area of research that focuses on synthesizing human-like movements to enhance communication and interaction in virtual environments. The task involves producing realistic and contextually appropriate gestures, often synchronized with speech or other non-verbal communication. It finds applications in domains such as virtual avatars, social robots, gaming, and animation \cite{Nyatsanga_2023}. The challenge of gesture generation lies in its multimodal nature, requiring the integration of audio, text, and visual inputs to produce coherent and expressive movements \cite{cheng2025hopheterogeneoustopologybasedmultimodal}. Achieving naturalistic gestures demands a deep understanding of human motion, cultural norms, and contextual cues. Consequently, it is an interdisciplinary field combining computer vision, machine learning, linguistics, and psychology. Recent advancements in deep learning and data-driven approaches have significantly improved the ability to generate complex gestures \cite{habibie2021learning}. These advancements have propelled the field forward, enabling the creation of models capable of mimicking human behavior with greater fidelity.

\subsection{Dataset}
Gesture models heavily depend on the quality and diversity of their training data. The efficacy of these models is directly correlated with the richness and representativeness of the datasets used to train them \cite{Kucherenko_2019}. As gesture generation advances, there has been a steady increase in the number and sophistication of datasets suitable for machine learning applications in human gesture analysis and synthesis.
Table \ref{table:GES-dataset} provides a comprehensive overview of major datasets for gesture generation, highlighting their key characteristics and contributions to the field. In recent years, there has been a growing focus on gesture generation, driven by the quest to achieve more natural and human-like animations \cite{Kucherenko_2020}. This increased attention has led to the development of more specialized and diverse datasets, each contributing to our understanding of different aspects of human gestures.

Gesture datasets capture multiple modalities to facilitate comprehensive analysis and synthesis of human motion. The primary modality is \textit{gesture data}, obtained through motion capture or video recordings, which provides the core movement information. \textit{Audio recordings} often accompany gestures, enabling researchers to explore the relationship between speech and motion. Additionally, \textit{text annotations}, including transcripts or descriptions of spoken content, support tasks such as text-to-gesture generation. To enhance contextual understanding, some datasets further incorporate \textit{gesture properties}, such as segmentation, categorical labels, or semantic descriptions. Finally, \textit{emotion annotations} play a crucial role in modeling expressive gestures by associating movements with underlying affective states.

Integrating these modalities in datasets allows researchers to develop more sophisticated and context-aware gesture-generation models. For instance, incorporating emotional annotations has led to significant advancements in generating emotionally appropriate gestures \cite{inproceedings99}.
As generative AI continues to evolve, we anticipate the emergence of even more comprehensive and nuanced gesture datasets. These future datasets may incorporate additional modalities such as physiological data or environmental context, further enhancing our ability to generate truly natural and contextually appropriate gestures \cite{article10}.

\begin{table*}[t]
\centering
\renewcommand{\arraystretch}{1.4}
\resizebox{\textwidth}{!}{%
    \begin{tabular}{|>{\centering\arraybackslash}p{7cm}|p{7cm}|P{4cm}|c|}
\hline
\multicolumn{1}{|c|}{\textbf{Name}} & \multicolumn{1}{c|}{\textbf{Statistics}} & \multicolumn{1}{c|}{\textbf{Modalities}} & \multicolumn{1}{c|}{\textbf{Link}} \\ \hline

IEMOCAP \cite{article}  & 
151 recorded dialogue videos, with 2 speakers per session, totaling 302 videos. Annotated for 9 emotions and valence, arousal, and dominance. Contains approximately 12 hours of audiovisual data.  
& Video, Audio, Text, Tabular Data  
& \href{https://sail.usc.edu/iemocap/}{IEMOCAP} \\ \hline

SaGA \cite{inproceedings}  & 
25 dialogues between interlocutors (50 total). Language: German. Published speakers: 6, unpublished speakers: 19. Annotated gestures: 1,764 (total corpus). Total video duration: 1 hour.  
& Video, Audio, Tabular Data  
& \href{https://www.phonetik.uni-muenchen.de/Bas/BasSaGAeng.html}{SaGA} \\ \hline

Creative-IT \cite{inproceedings2}  & 
Data from 16 actors (male and female). Affective dyadic interactions range from 2 to 10 minutes each. Approximately 8 sessions of audiovisual data were released.  
& Video, Audio, Text, Tabular Data  
& \href{https://sail.usc.edu/CreativeIT/ImprovRelease.html}{CreativeIT} \\ \hline

CMU Panoptic \cite{Joo_2017_TPAMI}  & 
3D facial landmarks from 65 sequences (5.5 hours). Contains 1.5 million 3D skeletons.  
& Video, Audio, Text  
& \href{http://domedb.perception.cs.cmu.edu/}{CMU Panoptic} \\ \hline

Speech-Gesture \cite{ginosar2019learning}  & 
A 144-hour dataset featuring 10 speakers. Includes frame-by-frame, automatically detected pose annotations.  
& Video, Audio  
& \href{https://people.eecs.berkeley.edu/~shiry/projects/speech2gesture/}{Speech-Gesture} \\ \hline

Talking With Hands 16.2M \cite{talk}  & 
16.2 million frames (50 hours) of two-person, face-to-face, spontaneous conversations. Strong covariance in arm and hand features.  
& Video, Audio  
& \href{https://github.com/facebookresearch/TalkingWithHands32M}{Talking With Hands} \\ \hline

PATS \cite{ahuja2020no}  & 
25 speakers, 251 hours of data, approximately 84,000 intervals. Mean interval length: 10.7 seconds.  
& Video, Audio, Text  
& \href{http://chahuja.com/pats/}{PATS} \\ \hline

Trinity Speech-Gesture II \cite{articleexperess}  & 
244 minutes of motion capture and audio (23 takes). Includes one male native English speaker. The skeleton consists of 69 joints.  
& Video, Audio, Tabular Data  
& \href{https://trinityspeechgesture.scss.tcd.ie/}{Trinity} \\ \hline

SaGA++ \cite{kucherenko2021speech2properties2gestures}  & 
25 recordings, totaling 4 hours of data.  
& Video, Audio, Text, Tabular Data  
& \href{https://svito-zar.github.io/speech2properties2gestures/}{SaGA++} \\ \hline

ZEGGS \cite{ghorbani2022zeroeggszeroshotexamplebasedgesture}  & 
67 monologue sequences with 19 different motion styles. Performed by a female actor speaking English. Total duration: 134.65 minutes.  
& Video, Audio  
& \href{https://github.com/ubisoft/ubisoft-laforge-ZeroEGGS}{ZEGGS} \\ \hline

BEAT \cite{liu2022beat}  & 
76 hours of 3D motion capture data from 30 speakers. Covers 8 emotions and 4 languages. Includes 32 million frame-level emotion and semantic relevance annotations.  
& Video, Audio, Text, Tabular Data  
& \href{https://pantomatrix.github.io/BEAT/}{BEAT} \\ \hline

BEAT2 \cite{liu2024emageunifiedholisticcospeech}  & 
60 hours of mesh-level, motion-captured co-speech gesture data. Integrates SMPL-X body and FLAME head parameters. Enhances modeling of head, neck, and finger movements.  
& Video, Audio, Tabular Data  
& \href{https://pantomatrix.github.io/EMAGE/}{BEAT2} \\ \hline

GAMT \cite{inbook}  & 
176 video clips of volunteers using math terms and gestures. Covers 8 classes of mathematical terms and gestures.  
& Video, Audio, Text  
& \href{https://openaccess.thecvf.com/content/CVPR2024W/MAR/html/Maidment_Using_Language-Aligned_Gesture_Embeddings_for_Understanding_Gestures_Accompanying_Math_Terms_CVPRW_2024_paper.html}{GAMT} \\ \hline

SeG \cite{zhang2024semanticgesticulatorsemanticsawarecospeech}  & 
208 types of global semantic gestures. 544 motion files recorded from a male performer. Each gesture is represented in 2.6 variations on average.  
& Video, Audio, Tabular Data  
& \href{https://pku-mocca.github.io/Semantic-Gesticulator-Page/}{SeG} \\ \hline

DND Group Gesture \cite{mughal2024convofusionmultimodalconversationaldiffusion}  & 
6 hours of gesture data from 5 individuals playing Dungeons \& Dragons. Recorded over 4 sessions (total duration: 6 hours). Includes beat, iconic, deictic, and metaphoric gestures.  
& Video, Audio, Tabular Data  
& \href{https://github.com/m-hamza-mughal/convofusion}{DND} \\ \hline

    \end{tabular}
}
\caption{Overview of various gesture datasets, including their main statistical features, modalities, and links.}
\label{table:GES-dataset}
\end{table*}

\subsection{Evaluation}

Evaluating gesture synthesis models requires a comprehensive set of metrics to capture the multifaceted nature of gestures, encompassing aspects such as style, appropriateness, semantic alignment, and physical realism. The choice of metrics plays a pivotal role in assessing the quality and effectiveness of gesture generation, as each metric targets specific aspects of performance. A key objective in gesture evaluation is measuring \textit{human-likeness}, which assesses how natural and lifelike the generated gestures appear. This is often evaluated through user studies, where participants judge the realism of the gestures. Alongside human likeness, \textit{appropriateness} is critical for determining whether the gestures align contextually with the accompanying input, such as speech or textual prompts. Both metrics are essential for capturing the subjective quality of generated gestures.

\textit{Style} is another important consideration in gesture synthesis, particularly when the generated motions must adhere to specific stylistic constraints. Metrics such as \textit{style correctness}, evaluated through user studies with dataset labels or random prompts, quantify how well the generated gestures conform to intended stylistic features. Complementing this, \textit{Style Recognition Accuracy (SRA)} provides an objective measure of the model’s ability to produce gestures matching predefined styles \cite{guo2025motionlabunifiedhumanmotion}. To ensure that gestures align semantically with input content, metrics such as \textit{Semantics-Relevant Gesture Recall (SRGR)} \cite{liu2022beat} and the \textit{Semantic Score (SC)} evaluate how well the generated gestures reflect the semantic meaning of the input. These metrics are particularly valuable in scenarios where gestures must convey meaningful and contextually appropriate information.

Physical realism is assessed using metrics that evaluate the accuracy and smoothness of motion. \textit{Mean Absolute Joint Error (MAJE)} quantifies the average error in joint positions by comparing the generated gestures to ground truth data. Similarly, \textit{Mean Acceleration Difference (MAD)} captures the smoothness of the generated motion by measuring acceleration discrepancies. Metrics like \textit{average jerk and acceleration} further ensure that the gestures are dynamically realistic. Additionally, statistical methods such as \textit{Canonical Correlation Analysis (CCA)} and \textit{Hellinger Distance} provide insights into the similarity of joint distributions between real and synthesized data. Temporal synchronization between gestures and input signals is another critical dimension of evaluation. \textit{Beat Consistency (BC)} measures the alignment of gestures with rhythmic features in speech or music, ensuring temporal coherence \cite{li2021aichoreographermusicconditioned}.  The \textit{Probability of Correct Keypoints (PCK)} metric, often employed in pose estimation, evaluates the spatial accuracy of key points relative to ground truth within a predefined threshold \cite{6380498}.

Finally, several metrics focus on assessing the perceptual quality of the generated gestures. The \textit{Fréchet Gesture Distance (FGD)} \cite{Yoon_2020} evaluates the distributional similarity between generated and real gestures, drawing on statistical comparisons. Similarly, metrics such as \textit{Learned Perceptual Image Patch Similarity (LPIPS)} \cite{8578166}, \textit{Fréchet Inception Distance (FID)} \cite{heusel2018ganstrainedtimescaleupdate}, and \textit{Fréchet Video Distance (FVD)} \cite{unterthiner2019accurategenerativemodelsvideo}, which were initially developed for images and videos, are adapted here to measure perceptual and distributional qualities in gesture evaluation. In cases where facial expressions are integral to gesture synthesis, metrics like \textit{Mean Squared Error (MSE)} and \textit{L1 Vertex Difference (LVD)} \cite{yi2023generatingholistic3dhuman} provide additional accuracy measurements for facial key points or vertices. By leveraging this diverse set of metrics, gesture evaluation frameworks aim to comprehensively assess the quality of generated gestures across visual, physical, and semantic dimensions. These metrics ensure that gesture synthesis models achieve fidelity to human motion but also expressiveness and contextual relevance, addressing the complex challenges of modern gesture generation.

Beyond individual evaluation metrics, standardized benchmarks are crucial in assessing and comparing gesture synthesis models. The GENEA Challenge \cite{10.1145/3577190.3616120} is one of the most comprehensive benchmarking platforms for co-speech gesture generation, providing a structured framework for evaluating state-of-the-art models under controlled conditions. By employing both objective metrics and large-scale user studies, the challenge enables rigorous assessment of gesture naturalness, appropriateness, and perceptual quality \cite{nagy2024genealeaderboardextended}. State-of-the-art systems such as EMAGE \cite{liu2024emageunifiedholisticcospeech} and MambaTalk \cite{xu2025mambatalkefficientholisticgesture} exemplify recent advancements, leveraging Transformer-based architectures, state space models, and multimodal learning techniques to enhance gesture diversity, synchronization, and efficiency. However, challenges such as maintaining long-term gesture consistency, reducing motion artifacts, and improving the interpretability of generated motions remain active research areas. Initiatives like GENEA drive progress toward more robust and human-like gesture generation models by continuously refining evaluation methodologies and fostering community-driven benchmarking efforts.

\subsection{Models}
\subsubsection{Rule-Based and Parametric Models}\label{gesture_rule-based_and_parametric_models}
Traditional approaches to gesture generation primarily rely on rule-based systems and parametric techniques to synthesize realistic gestures by leveraging predefined knowledge and high-level control parameters. These models often use handcrafted rules and heuristics based on observations of human behavior, allowing for expressive and controllable gesture generation. One common approach within these systems is parameter-based procedural animation, where high-level parameters, such as emotion, speech intensity, or rhythm, guide the selection and interpolation of predefined keyframes \cite{inproceedingsevaluaaaa}. By smoothly blending between keyframes, these methods can create convincing and coherent motion sequences in response to dynamic inputs. Another effective technique in rule-based models is using blendshape models for generating hand and finger gestures. Blendshape models define a set of base shapes (or blend shapes) and blend between them using weights, enabling fine-grained control over specific regions of the hand \cite{:10.2312/egst.20141042}.

In addition to traditional methods, BEAT (Behavior Expression Animation Toolkit) \cite{cassell2001beat} provides a rule-based framework that automatically analyzes text input to suggest appropriate gestures based on linguistic and contextual rules. It generates synchronized speech, facial expressions, and gestures by applying a comprehensive set of behavior generation rules derived from extensive studies of human communicative patterns. This makes it particularly effective for creating natural nonverbal behaviors for virtual agents.
These models are particularly advantageous for applications requiring detailed hand movements, as they allow for both low-level control and smooth transitions between gestures. B blend shape models achieve realistic and nuanced animations by incorporating morphing techniques, making them a popular choice in human-computer interaction systems. While rule-based and parametric models offer flexibility and explicit control over gesture generation, they often require extensive manual design and fine-tuning to cover diverse scenarios and variations in human motion. 

\subsubsection{Classical Deep Learning Models}\label{gesture_deep_learning-based_models}

Advancements in deep learning have led to the development of gesture prediction and generation models that leverage neural networks to synthesize human gestures end-to-end. These models typically incorporate architectures like Generative Adversarial Networks (GANs) \cite{goodfellow2014generative} and Recurrent Neural Networks (RNNs) \cite{rumelhart1985learning} to model the relationship between input modalities such as audio, text, and gestures. One notable model, GestureGAN \cite{tang2019gestureganhandgesturetogesturetranslation}, uses GANs to generate gesture animations conditioned on audio inputs. The model employs a generator-discriminator framework where the generator learns to synthesize realistic gesture sequences, and the discriminator evaluates the realism and coherence of these sequences. The network effectively captures the dynamics of hand gestures, enabling robust gesture-to-gesture translation even in challenging scenarios. Another significant model, Speech2Gesture \cite{ginosar2019gestures}, generates co-speech gestures directly from speech features using Long Short-Term Memory (LSTM) networks \cite{hochreiter1997long} or Recurrent Neural Networks (RNNs) \cite{rumelhart1985learning}. This architecture effectively captures temporal dependencies between speech and gestures, utilizing specialized audio processing layers to establish a strong correlation between speech dynamics and gestural movements. 

Building upon the need for personalized gestures, Audio-Driven Adversarial Gesture Generation \cite{zhu2023taming} combines GANs with Conditional Variational Autoencoders (CVAE). This model captures the multimodal nature of gestures by aligning audio and motion features within a shared latent space, resulting in a more nuanced generation of audio-driven gestures. Another notable approach, GestureMaster \cite{10.1145/3536221.3558063}, employs graph neural networks (GNNs) \cite{scarselli2008graph} to represent gesture relationships. By adopting a graph-based framework, GestureMaster effectively captures the spatial and temporal dependencies within gesture sequences, enhancing the model’s ability to produce naturalistic hand and body gestures. 

\begin{figure}
    \centering
    \includegraphics[width=\textwidth]{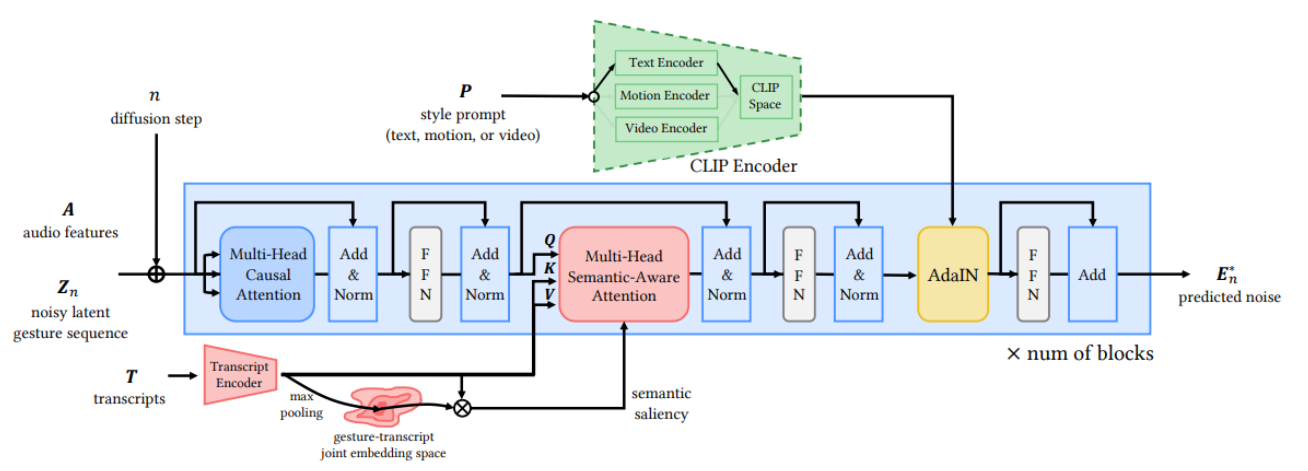}
    \caption{GestureDiffuCLIP \cite {ao2023gesturediffuclipgesturediffusionmodel} integrates a CLIP encoder for semantic alignment and a diffusion process for refining gesture sequences. The model uses multi-head causal and semantic-aware attention mechanisms and adaptive instance normalization (AdaIN) to generate expressive gestures. 
    Reprinted from \cite{ao2023gesturediffuclipgesturediffusionmodel}.
    }
    \label{fig:gesture_fig}
\end{figure}

\subsubsection{Diffusion-Based Models}\label{gesture_diffu_models} 
In recent years, diffusion-based models have emerged as a powerful alternative for generating high-quality, coherent human gestures. These models leverage a generative process inspired by diffusion \cite{ho2020denoisingdiffusionprobabilisticmodels}, gradually transforming random noise into structured outputs through a series of iterative steps. By modeling the underlying distribution of gestures, diffusion models excel at capturing complex, high-dimensional data and producing realistic, continuous motion sequences. 

DiM-Gesture \cite{zhang2024dimgesturecospeechgesturegeneration} utilizes an adaptive layer normalization mechanism called Mamba-2 \cite{dao2024transformersssmsgeneralizedmodels}, which adapts well to different speakers by leveraging multi-source data during training \cite{dao2024transformersssmsgeneralizedmodels}. This model focuses on generating realistic co-speech gestures using speech information as the driving input. Similarly, AMUSE \cite{Chhatre_2024_CVPR} employs a disentangled latent diffusion technique to generate emotionally expressive 3D body animations. AMUSE effectively controls emotions through a multi-stage training pipeline by separating emotional expressions and gestures. 

Addressing broader speaker movements, the FreeTalker model \cite{yang2024Freetalker} introduces a diffusion-based framework with classifier-free guidance for style control. FreeTalker generates natural transitions between gesture clips using a generative prior, called DoubleTake, resulting in coherent and spontaneous movements beyond just co-speech gestures. DiffuGesture \cite{10.1145/3610661.3616552} focuses on generating gestures in two-person dialogues using specialized diffusion techniques. By modeling interpersonal gesture interactions, DiffuGesture ensures natural and synchronized gestures between conversational partners. Expanding on diffusion-based multimodal modeling, CSMP \cite{Deichler_2023} employs a co-speech gesture generation approach that utilizes joint text and audio representations. By capturing intricate relationships between these modalities, CSMP produces coherent gestures that align well with both speech content and rhythm.

\subsubsection{Transformer-Based Models}\label{gesture_transformer-based_models}  
Transformer architectures have significantly influenced gesture generation, enhancing contextual understanding, multimodal integration, and stylistic adaptation. These models leverage self-attention mechanisms and cross-modal learning to generate expressive and contextually relevant gestures. One of the pioneering Transformer-based models, Gesticulator \cite{Kucherenko_2020}, employs a multimodal Transformer architecture to generate gestures conditioned on both text and audio inputs. Gesticulator produces natural and synchronized gestures that accurately reflect co-speech dynamics by effectively combining semantic information from text with prosodic speech cues.
Expanding on vision-based Transformers, ViTPose \cite{xu2022vitposesimplevisiontransformer} applies Vision Transformers (ViTs) to human pose estimation, serving as a foundational approach for gesture synthesis. By accurately capturing human pose dynamics, ViTPose enhances gesture generation models by providing precise motion representations. 

Addressing the need for style adaptation, StyleGestures \cite{alexanderson2020style} integrates Transformer-based sequence modeling with style tokens to synthesize gestures that align with individual speaker characteristics. This encoder-decoder framework learns stylistic variations in gesture production, enabling the generation of gestures that match the personal speaking style of different individuals. Lastly, ZeroEGGS \cite{ghorbani2022zeroeggszeroshotexamplebasedgesture} presents a zero-shot learning paradigm for gesture generation. Using example-based learning, ZeroEGGS generalizes gestures to unseen speaker styles, offering high flexibility in synthesizing co-speech gestures that match diverse speaker preferences. These Transformer-based models showcase the power of self-attention mechanisms and cross-modal integration, significantly improving gesture generation in terms of contextual accuracy, personalization, and multimodal synchronization.

\subsubsection{Hybrid Models}\label{gesture_hybrid_models} 
Hybrid models combine different deep learning architectures, such as Transformers, diffusion-based techniques, and adversarial learning, to enhance the accuracy and expressiveness of gesture generation. These models leverage multimodal learning, cross-modal attention mechanisms, and style adaptation to synthesize natural and context-aware gestures. For example, GestureDiffuCLIP \cite{ao2023gesturediffuclipgesturediffusionmodel} integrates contrastive language-image pre-training (CLIP) with a diffusion-based process to iteratively refine gesture sequences. The model aligns textual descriptions with motion outputs using multi-head causal and semantic-aware attention mechanisms, while adaptive instance normalization (AdaIN) generates expressive gestures that maintain semantic coherence \cite{huang2017arbitrarystyletransferrealtime}. This process is illustrated in Figure \ref{fig:gesture_fig}. ZS-MSTM \cite{fares2023zsmstmzeroshotstyletransfer} introduces a zero-shot style transfer method for gesture animation, driven by both text and speech inputs, utilizing adversarial disentanglement to separate style and content features, enabling gesture style transfer across different speakers without requiring speaker-specific training data.

\begin{figure}
    \centering
    \includegraphics[width=0.5\textwidth]{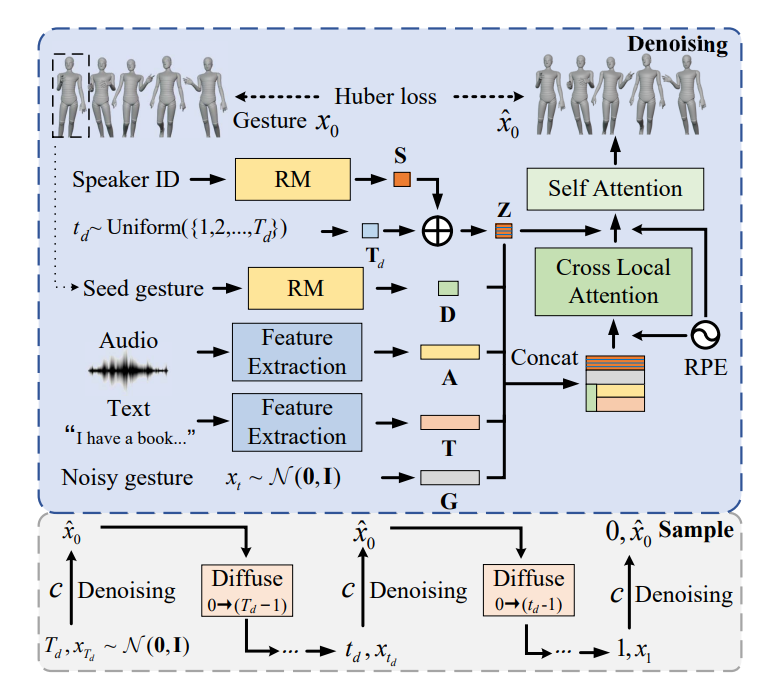}
    \caption{The architecture of \textit{DiffuseStyleGesture+} \cite{Yang_2023}, showcasing its multimodal integration for co-speech gesture generation. The model incorporates speaker IDs, seed gestures, audio, and text through feature extraction and representation modules. A diffusion process is employed to refine noisy gestures into expressive outputs, leveraging self-attention and cross-local attention mechanisms. The system also uses Huber loss \cite{huber1992robust} for enhanced denoising performance and produces stylistically diverse gesture outputs. Reprinted from \cite{Yang_2023}.
    }
    \label{fig:gesture_fig2}
\end{figure}

Models like SAGA (Style and Grammar-Aware Gesture Generation) \cite{sagaaaaaaa} combine recurrent and Transformer-based approaches to integrate grammatical and stylistic features for gesture synthesis. It combines an LSTM-based encoder-decoder for sequence modeling with a Transformer-based grammar encoder, improving alignment with linguistic structures. C2G2 \cite{ji2023c2g2controllablecospeechgesture} focuses on modular control over gesture synthesis, offering a framework for controllable co-speech gesture generation. This framework allows users to specify gesture attributes such as style, speed, and intensity, providing flexibility in gesture animation. Similarly, CoCoGesture \cite{qi2024cocogesturecoherentcospeech3d} employs a Transformer-based diffusion model that leverages a large-scale dataset, GES-X, and integrates a mixture-of-experts (MoE) framework to align gestures with human speech effectively, without relying on auxiliary text inputs \cite{10.5555/3586589.3586709}. DiffuseStyleGesture+ \cite{Yang_2023} extends diffusion-based gesture generation by incorporating multimodal inputs such as speaker IDs, seed gestures, audio, and text. Leveraging self-attention and cross-local attention mechanisms, the model refines gesture outputs to generate personalized and stylistically diverse gestures. Figure \ref{fig:gesture_fig2} shows this model's architecture. ExpressGesture \cite{articleexperess} synthesizes expressive gestures by integrating emotion recognition, allowing for the creation of movements that match the content of speech and reflect the underlying sentiment, resulting in emotionally expressive gestures. DiffSHEG \cite{chen2024diffshegdiffusionbasedapproachrealtime} utilizes a diffusion-based approach for real-time speech-driven 3D expression and gesture generation, refining motion outputs through iterative denoising steps to enhance the realism of gesture animations.

Gesture Motion Graphs \cite{mewwwwwww} aims for few-shot gesture reenactment using graph-based modeling of motion sequences, capturing temporal relationships within gesture transitions to adapt effectively to novel speakers with limited training data. Mix-StAGE \cite{ahuja2020styletransfercospeechgesture} incorporates an attention-based encoder-decoder architecture with a style encoder, capturing individual speaker styles. Combining spatial and temporal attention mechanisms enables expressive and personalized gesture synthesis while maintaining temporal coherence. These hybrid models showcase the power of combining various deep learning paradigms to produce expressive, controllable, and semantically aligned gesture synthesis.

EMAGE \cite{liu2024emageunifiedholisticcospeech} introduces a framework for generating full-body human gestures from audio and masked gestures, offering comprehensive control over facial, body, hand, and global movements. Using a Masked Audio Gesture Transformer, EMAGE effectively encodes and synthesizes audio-to-gesture generation while incorporating masked gesture priors to boost inference performance. The model also adaptively merges speech features with compositional VQ-VAEs \cite{oord2018neuraldiscreterepresentationlearning}, generating diverse and high-fidelity gestures synchronized with audio input. EMAGE's flexible design accepts predefined spatial-temporal gesture inputs, producing holistic, audio-synchronized gesture outputs. State-of-the-art gesture synthesis models often rely on diffusion and attention mechanisms, but their high computational complexity limits efficiency in generating long and diverse sequences. MambaTalk addresses this challenge by leveraging Selective State Space Models (SSMs) \cite{gu2022efficientlymodelinglongsequences}, offering a more efficient and scalable approach to holistic gesture generation \cite{xu2025mambatalkefficientholisticgesture}. By implementing a two-stage modeling strategy with discrete motion priors, MambaTalk enhances gesture quality while maintaining low latency. Built on the Mamba block \cite{gu2024mambalineartimesequencemodeling}, it enables multimodal integration, improving gesture diversity and rhythm without the heavy computational cost of Transformers or diffusion models.

\subsection{Application}
Gestures are foundational to enhancing the realism and emotional depth of generative AI and animations, making them essential for creating engaging and immersive experiences across various domains. In animation, gestures bring virtual characters to life by imbuing them with human-like expressiveness and natural movements, critical for storytelling in movies, video games, and cinematic experiences \cite{torshizi2025largelanguagemodelsvirtual}. In gaming, gesture synthesis enables lifelike avatars and NPCs (non-playable characters) to generate contextually appropriate motions from inputs such as speech, text, or user actions, which heighten the player's sense of immersion and connection to the game world \cite{Gallotta_2024}. Techniques like co-speech gesture generation have been proposed to produce human gestures synchronized with audio, enhancing avatar realism in interactive gaming \cite{article234}. Similarly, in movies and videos, gesture-driven AI facilitates the creation of dynamic performances, allowing directors to craft complex scenes involving digital characters or augmented environments with precision and creativity \cite{article1111111}.

Beyond entertainment, gesture-based interactions are revolutionizing interfaces and workflows in professional fields. For instance, gesture animations play a critical role in VR and AR applications, enabling intuitive manipulation of 3D content for architects, designers, and engineers. In collaborative virtual spaces, real-time gesture-driven animations enhance communication by translating users' body language and expressions into digital form, fostering a sense of presence and shared engagement \cite{inproceedings77}. Additionally, gesture control systems are transforming user interfaces by replacing traditional input methods, offering seamless and intuitive ways to interact with digital environments, from gaming to human-machine collaboration \cite{ffff}. Studies have explored the evolution of gesture-based interactions and their impact on human-computer interaction, emphasizing their role in bridging the gap between physical and virtual worlds \cite{article1111}. Overall, gestures are a cornerstone for bridging the gap between physical and virtual worlds, making animations, gaming, movies, and videos more interactive, realistic, and impactful.

\section{Motion}\label{sec_motion}
Human motion describes the coordinated spatiotemporal dynamics of body movements, encompassing the interplay of limb articulation, facial expressions, hand gestures, and overall trajectories as a unified whole. It includes both low-level kinematic details (e.g., joint rotations, velocities) and high-level semantic actions (e.g., walking, dancing) \cite{Zhu2024MotionSurvey, HumanPoseEstimation}. Motion modeling is foundational to applications like animation, robotics, and virtual reality, where generating natural and expressive movements is critical. Traditional marker-based motion capture systems are limited to lab settings, but recent advances in markerless methods and deep learning enable scalable 3D motion estimation from RGB videos \cite{motionx++, HumanPoseEstimation}. Notably, whole-body motion synthesis requires capturing not just body poses but also fine-grained details like facial expressions and hand gestures, which are often omitted in older datasets \cite{motionx++}.
Furthermore, human motion synthesis involves generating 3D human motion sequences conditioned on textual descriptions (e.g., a person jumps while waving both hands). It bridges natural language understanding and motion dynamics, demanding alignment between linguistic semantics and kinematic plausibility \cite{azadi2023makeananimationlargescaletextconditional3d}.

\subsection{Dataset}
Early datasets focused on body-only motions with limited textual diversity, but recent efforts emphasize multimodal annotations (text, audio, video) and whole-body expressiveness \cite{motionx++, Zhu2024MotionSurvey, motionscript}. Challenges include temporal coherence, contextual grounding, and handling ambiguous textual inputs \cite{YE202235, Zhu2024MotionSurvey}.
From the technical perspective, the evolution of 3D human motion datasets has been closely tied to advancements in motion representation paradigms, each offering unique advantages and limitations. Early datasets, such as CMU MoCap \footnote{https://mocap.cs.cmu.edu/}, relied on marker trajectories and hierarchical joint-angle representations to capture lab-based motions like walking or jumping. These formats prioritized compatibility with animation tools like Blender but lacked standardization, leading to fragmented workflows. For instance, the KIT Motion-Language Dataset \cite{KIT} combined BVH files with FBX formats to support text-driven animation pipelines. However, its text annotations remained simplistic and limited to locomotive actions. BVH (Biovision Hierarchy) and FBX (Filmbox) are file formats used to store 3D human motion data, particularly from motion capture systems. BVH is a text-based format focusing on skeletal animation, defining the hierarchy of joints and their movements over time. FBX is a binary format that can include motion data and 3D models, scenes, and textures, making it suitable for integrating with animation software.

The introduction of parametric body models, notably SMPL \cite{SMPL} and its extension SMPL-X \cite{SMPL-X:2019}, revolutionized motion representation by unifying body, face, and hand movements into a single mesh topology. Datasets such as AMASS \cite{AMASS} consolidated 15 marker-based datasets into SMPL parameters, enabling large-scale training of generative models. However, these early efforts focused primarily on body poses, overlooking expressive details. Motion-X addressed this limitation \cite{motionx}, which introduced SMPL-X annotations to capture whole-body motions along with multimodal data such as text and audio. Before that, HumanML3D \cite{Guo_2022_CVPR} paired SMPL body parameters with crowdsourced text labels but lacked facial and hand dynamics, reflecting the common trade-off between scalability and expressiveness.
Textual representations have also evolved, with datasets employing various annotation strategies. BABEL \cite{BABEL} paired SMPL motions from AMASS with concise, verb-centric labels. In contrast, MotionScript \cite{motionscript} introduced rule-based natural language descriptions (e.g., left elbow bent at 45 degrees) to improve fine-grained text-to-motion alignment. Table \ref{table:motion-table} provides a more comprehensive list of key motion-related datasets.

\begin{table*}[t]
\centering
\renewcommand{\arraystretch}{1.4}
\resizebox{\textwidth}{!}{%
    \begin{tabular}{|>{\centering\arraybackslash}p{7cm}|p{7cm}|p{4cm}|c|}
\hline
\multicolumn{1}{|c|}{\textbf{Name}} & \multicolumn{1}{c|}{\textbf{Statistics}} & \multicolumn{1}{c|}{\textbf{Modalities}} & \multicolumn{1}{c|}{\textbf{Link}} \\ \hline

Motion-X++ \cite{motionx++} &  
19.5 million 3D poses across 120,500 sequences, synchronized with 80,800 RGB videos and 45,300 audio tracks, and annotated with free-form text descriptions.  
& 3D/Point Cloud Data, Text, Audio, Video  
& \href{https://github.com/IDEA-Research/Motion-X}{Motion-X++} \\ \hline

HumanMM (ms-Motion) \cite{humanmm} &  
120 long-sequence 3D motions reconstructed from 600 in-the-wild multi-shot videos, totaling 237 minutes of data. Includes rare interactions.  
& 3D/Point Cloud Data, Video  
& \href{https://github.com/zhangyuhong01/HumanMM-code}{HumanMM} \\ \hline

Multimodal Anatomical Motion \cite{multimodalhumanmotiondataset} &  
51,051 annotated poses with 53 anatomical landmarks, captured across 48 virtual camera views per pose. Includes 2,000+ pathological motion variations.  
& 3D/Point Cloud Data, Text  
& \href{https://data.mendeley.com/datasets/493s6f753v/2}{Multimodal Anatomical Motion} \\ \hline

AMASS \cite{AMASS} &  
11,265 motion clips aggregated from 15 mocap datasets (e.g., CMU, KIT), totaling 43 hours of motion data in SMPL format. Covers 100+ action categories.  
& 3D/Point Cloud Data  
& \href{https://amass.is.tue.mpg.de/}{AMASS} \\ \hline

HumanML3D \cite{Guo_2022_CVPR} &  
14,616 motion sequences (28.6 hours) paired with 44,970 free-form text descriptions spanning 200+ action categories.  
& 3D/Point Cloud Data, Text  
& \href{https://github.com/EricGuo5513/HumanML3D}{HumanML3D} \\ \hline

BABEL \cite{BABEL} &  
43 hours of motion data from AMASS, annotated with 250+ verb-centric action classes across 13,220 sequences. Includes temporal action boundaries.  
& 3D/Point Cloud Data, Text  
& \href{https://babel.is.tue.mpg.de/}{BABEL} \\ \hline

AIST++ \cite{aist++} &  
1,408 dance sequences (10.1 million frames) captured from 9 camera views, totaling 15 hours of multi-view RGB video data.  
& 3D/Point Cloud Data, Video  
& \href{https://google.github.io/aistplusplus_dataset/}{AIST++} \\ \hline

3DPW \cite{3DPW} &  
60 sequences (51,000 frames) captured in diverse indoor/outdoor environments, featuring challenging poses and natural object interactions.  
& 3D/Point Cloud Data, Video  
& \href{https://virtualhumans.mpi-inf.mpg.de/3DPW/}{3DPW} \\ \hline

PROX \cite{PROX} &  
20 subjects performing 12 interactive scenarios in 3D scenes, including 180 annotated RGB frames for scene-aware motion analysis.  
& 3D/Point Cloud Data, Images  
& \href{https://prox.is.tue.mpg.de/index.html}{PROX} \\ \hline

KIT-ML \cite{KIT} &  
3,911 motion clips (11.23 hours) with 6,278 natural language annotations containing 52,903 words, stored in BVH/FBX formats.  
& 3D/Point Cloud Data, Text  
& \href{https://git.h2t.iar.kit.edu/sw/motion-annotation}{KIT-ML} \\ \hline

CMU MoCap &  
2605 trials across 6 categories and 23 subcategories, performed by over 140 subjects.  
& 3D/Point Cloud Data, Audio  
& \href{https://mocap.cs.cmu.edu/}{CMU MoCap} \\ \hline
    \end{tabular}
}
\caption{Comprehensive collection of datasets for 3D human motion generation and synthesis, highlighting key statistics and modalities.} 
\label{table:motion-table}
\end{table*}

\subsection{Evaluation}
Text-to-motion models promise to animate characters with lifelike movements that faithfully reflect the nuances of written prompts, but their success hinges on rigorous evaluation. Ensuring that generated motions are realistic, diverse, and precisely aligned with the input text requires a multifaceted approach. Modern research has developed a sophisticated evaluation framework that balances quantitative metrics with human judgment, drawing inspiration from computer vision and natural language processing. By examining how contemporary models are assessed, we can uncover the factors defining their performance and the metrics illuminating their strengths and limitations.

At the heart of text-to-motion evaluation lies the pursuit of \textit{fidelity}, the degree to which generated motions match the realism of actual human movements. Researchers have adapted the \textit{Fréchet Inception Distance (FID)} to compare the feature distributions of synthetic and real motion sequences. Lower FID scores signal motions closely resembling their real-world counterparts, a critical factor for immersive applications like virtual reality.

Equally important is the concept of \textit{consistency}, which measures how well a generated motion captures the semantic intent of the input text. Without this alignment, even the most realistic motion risks becoming irrelevant. The \textit{R-Precision} metric \cite{xu2018attngan} has emerged as a cornerstone for evaluating this aspect, assessing whether a motion can retrieve its corresponding text prompt from a set of candidates. By embedding motions and texts into a shared space, often trained with contrastive loss, R-Precision calculates the proportion of times the correct text ranks among the top-k matches, typically at thresholds like top-1 or top-3. Complementing R-Precision, the \textit{MultiModal Distance} \cite{guo2022tm2t} quantifies the Euclidean distance between motion and text embeddings, with lower values indicating tighter semantic coupling.

\textit{Diversity} represents another critical dimension, ensuring that text-to-motion models can produce a rich variety of motions across different prompts and within the same one. A model that generates repetitive or stereotypical movements fails to capture the breadth of human expression, limiting its utility in creative applications. The \textit{Diversity} metric \cite{Textguided3H} calculates the average distance between pairs of randomly sampled motion sequences and evaluates a model’s ability to span the motion space. Meanwhile, \textit{Multimodality} \cite{Textguided3H} focuses on diversity within a single prompt, measuring the variance among multiple motions generated from the same text. These metrics collectively ensure that models remain versatile, avoiding the pitfall of overfitting to common motion patterns.

Beyond these quantitative measures, the subjective quality of generated motions plays an indispensable role in evaluation, as automated metrics alone cannot fully capture the nuances of human perception. \textit{User studies}, where human participants rate motions for naturalness and appropriateness, provide insights into aspects like emotional expressiveness or contextual relevance that quantitative metrics may not fully capture \cite{li2024motion}. Such studies highlight the importance of aligning technical performance with user experience, particularly in applications where audience engagement is paramount.

Over the past few years, the community has made significant headway in refining evaluation benchmarks to capture better how faithfully generated motions adhere to textual input. HumanML3D \cite{Guo_2022_CVPR}, for example, has emerged as a standard reference point, offering a broad range of test scenarios and consistently measured metrics like FID for visual realism, R-Precision for semantic alignment, and Diversity for sample variety. These benchmarks, along with others such as KIT Motion-Language \cite{KIT}, help uncover trade-offs across various modeling approaches, highlighting strengths and weaknesses in terms of text matching, temporal coherence, and multi-modal variations. They also reveal ongoing gaps in measurement, as no single metric fully captures subjective motion quality or subtle semantics. Models such as MoMask \cite{guo2023momaskgenerativemaskedmodeling}, DiverseMotion \cite{diversemotion}, and Motion Anything \cite{zhang2025motion} currently achieve state-of-the-art performance, excelling in text alignment, realism, or both, signifying steady progress towards more expressive and accurate text-driven motion generation.

\subsection{Models} 
Generative AI and Deep Learning have significantly advanced the field of generative modeling for 3D human motion, particularly for synthesis conditioned on inputs like natural language. Recent research leverages powerful architectures such as Variational Autoencoders (VAEs) and their variations and Diffusion Models.
Early foundational works established methods for bridging the gap between textual concepts and motion sequences. For instance, Language2Pose \cite{DBLP:journals/corr/abs-1907-01108} introduced a neural architecture to learn a joint embedding space where both language descriptions and 3D pose sequences could be encoded, enabling the generation of animations directly from text input. Subsequently, MotionCLIP \cite{motionclip} demonstrated the power of leveraging large pre-trained models by aligning a transformer-based motion autoencoder's latent space with the rich semantic space of CLIP \cite{DBLP:conf/icml/RadfordKHRGASAM21}. This alignment infused the motion generation process with strong semantic understanding, allowing for more nuanced and even abstract text-to-motion capabilities \cite{motionclip}.

\subsubsection{VAE-Based Models}\label{motion_vae} 
Variational Auto-Encoder (VAE) architectures are frequently employed in motion generation models because they capture complex motion patterns and learn meaningful latent representations. A prominent example is the Action-Conditioned Transformer VAE \cite{DBLP:journals/corr/abs-2104-05670}, a model designed for generating diverse and realistic 3D human motions conditioned on action labels. It combines transformers with a VAE structure, enabling the model to produce multiple motion variations from the same action condition. Typically, its encoder maps an input motion sequence into a latent Gaussian distribution. At the same time, the decoder uses samples from this distribution, along with the action label, to generate new motion sequences. This model is often adopted as a baseline in subsequent research due to its strong performance and flexibility. TEMOS (Text-To-Motions) \cite{petrovich22temos} adapts the architecture for text-conditioned generation of SMPL \cite{SMPL} body motions. It modifies the Action-Conditioned Transformer VAE (referred to as ACTOR \cite{DBLP:journals/corr/abs-2104-05670}) by replacing action conditioning with text prompts and introducing two symmetric encoders: one processing motion sequences and the other using frozen DistilBERT \cite{sanh2020distilbertdistilledversionbert} embeddings for text. Both encoders map their inputs to parameters of a Gaussian distribution in a shared latent space. The decoder then reconstructs motions while training objectives align the text and motion representations within this latent space, aiming for strong cross-modal consistency.

Addressing the challenge of generating coherent motion from sequential instructions, TEACH (Temporal Action Composition for 3D Humans) \cite{athanasiou2022teachtemporalactioncomposition} extends TEMOS \cite{petrovich22temos} to handle a \textit{sequence} of text descriptions. Its key innovation is a hierarchical generation strategy: it operates non-autoregressively within individual actions but autoregressively across the sequence of actions specified by the input texts. This allows for temporal composition and smoother transitions between consecutive motions described in the text sequence. Differing in architectural design, T2M (Text2Motion) \cite{Guo_2022_CVPR} employs a distinct two-stage framework for text-to-motion synthesis. It first pre-trains a convolutional auto-encoder to learn compact motion representations (motion codes/snippets). Subsequently, a separate temporal VAE module, incorporating RNNs and conditioned on text features, generates sequences of these motion codes and predicts the overall motion duration based on the input text. The pre-trained motion decoder then translates these generated codes into the final 3D motion.

\begin{figure}[t]
\begin{center}
  \includegraphics[width=\textwidth]{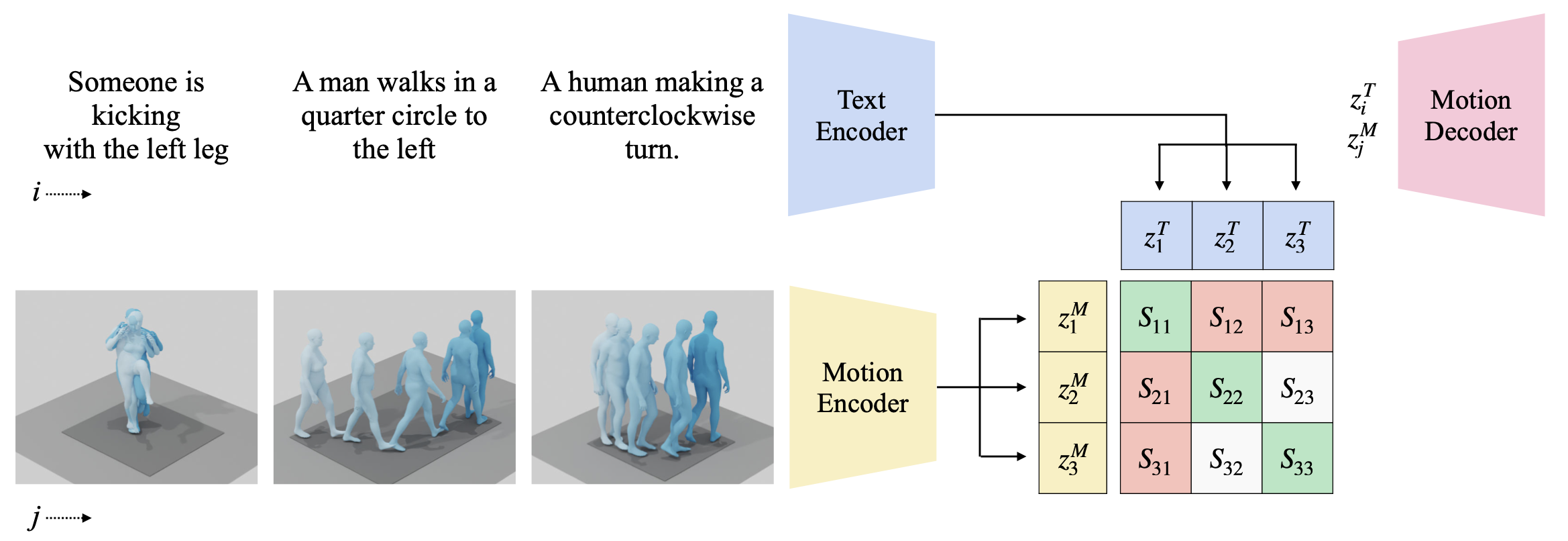}
\end{center}
\caption{\label{fig:tmr} 
The architecture of the TMR \cite{petrovich2023tmrtexttomotionretrievalusing} framework shows the use of dual encoders for text and motion, and a joint embedding space for similarity-based retrieval. Reprinted from \cite{petrovich2023tmrtexttomotionretrievalusing}.
}
\end{figure}

Shifting focus towards improving the alignment between text and motion representations, TMR (Text-to-Motion Retrieval) \cite{petrovich2023tmrtexttomotionretrievalusing} enhances the TEMOS \cite{petrovich22temos} framework primarily for the task of motion retrieval, though still capable of synthesis. Inspired by CLIP \cite{DBLP:conf/icml/RadfordKHRGASAM21}, TMR incorporates a contrastive loss objective during training to explicitly structure the joint latent space, pushing embeddings of corresponding text-motion pairs closer while separating mismatched pairs. Crucially, it introduces a technique (using MPNet \cite{song2020mpnetmaskedpermutedpretraining}) to filter out potentially misleading negative pairs (texts describing similar motions) during contrastive training, leading to significantly improved retrieval performance over prior methods such as TEMOS \cite{petrovich22temos} and T2M \cite{Guo_2022_CVPR}. The main structure of this model is illustrated in Figure \ref{fig:tmr}.

\subsubsection{VQ-VAE-Based Models}\label{motion_vq_vae}  
A notable direction in 3D motion generation involves models based on a VQ-VAE (Vector Quantized Variational Autoencoders) backbone, where motion is represented as a sequence of discrete tokens sampled from a learned codebook. This formulation treats motion synthesis similarly to language modeling, allowing for more structured, controllable generation. The idea was first introduced in T2M-GPT \cite{t2m-gpt}, which uses a transformer with causal masked self-attention to autoregressively predict motion tokens conditioned on text. The motion is encoded into tokens using a pre-trained VQ-VAE, and text prompts are encoded via CLIP \cite{DBLP:conf/icml/RadfordKHRGASAM21}, enabling the model to synthesize motion by generating a sequence of token indices one by one.

Following this foundation, later models explore ways to improve diversity, efficiency, and temporal coherence. DiverseMotion \cite{diversemotion} replaces the autoregressive decoding with a diffusion process, corrupting motion tokens during the forward pass and denoising them in reverse, conditioned on CLIP-based text embeddings. To capture richer semantic context from text, it introduces Hierarchical Semantic Aggregation (HSA), which aggregates token-level embeddings using learnable weights and MLPs. Meanwhile, MoMask \cite{guo2023momaskgenerativemaskedmodeling} adopts a hierarchical codebook structure, combining masked token modeling (inspired by BERT \cite{devlin-etal-2019-bert}) with a residual refinement stage. This allows the model first to generate coarse motion using masked prediction and then incrementally add fine-grained motion detail across quantization layers.

\begin{figure}[t]
\begin{center}
  \includegraphics[width=\textwidth]{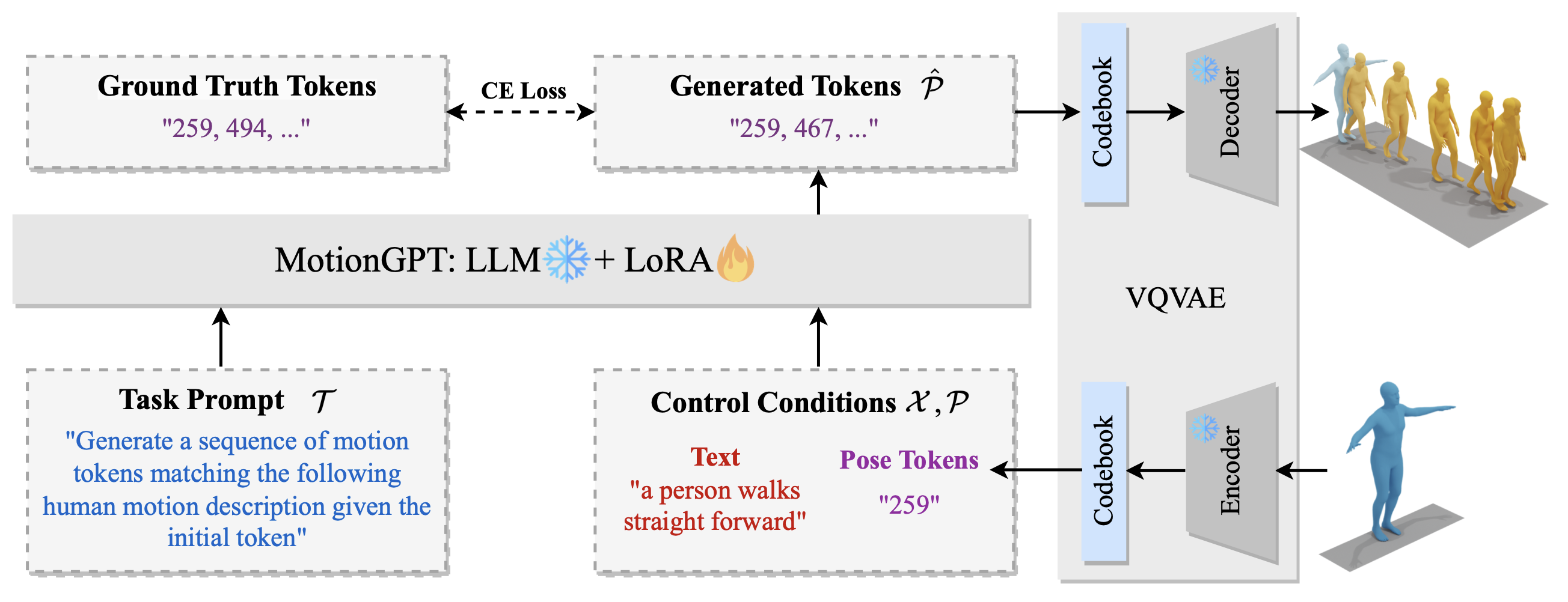}
\end{center}
\caption{\label{fig:motiongpt} 
An overview of MotionGPT \cite{zhang2024motiongptfinetunedllmsgeneralpurpose}, which uses a frozen VQ-VAE and LLM, with LoRA \cite{hu2022lora} applied to fine-tune the LLM for generating motion tokens. Reprinted from \cite{zhang2024motiongptfinetunedllmsgeneralpurpose}. 
}

\end{figure}

Longer and more complex motion sequences are addressed in T2LM \cite{lee2024t2lmlongterm3dhuman}, which maps multi-sentence text into motion using a 1D-convolutional VQ-VAE and a transformer-based text encoder, enabling smooth transitions across actions. Taking a different route, MotionGPT \cite{zhang2024motiongptfinetunedllmsgeneralpurpose} combines VQ-VAE with a large language model (LLM), treating motion generation as a sequence-to-sequence task. Using LoRA \cite{hu2022lora} for efficient fine-tuning, the LLM predicts motion tokens directly from text, offering faster training and strong generalization. The main structure of MotionGPT \cite{zhang2024motiongptfinetunedllmsgeneralpurpose} is illustrated in Figure \ref{fig:motiongpt}. Together, these models demonstrate the flexibility of the VQ-VAE formulation and open new directions for combining motion modeling with advances in language and generative modeling.

\subsubsection{Diffusion-Based Models} \label{motion_diffusion}
Following the success of diffusion models in text-to-image generation \cite{rombach-high-res, pmlr-v139-ramesh21a}, the field of text-to-human motion generation has increasingly adopted the Denoising Diffusion Probabilistic Model (DDPM) framework. These approaches begin by corrupting motion sequences with noise and learn to reverse this process through a denoising network, conditioned on natural language descriptions. Given the sequential nature of motion, transformers are often used within the denoising process to model temporal dependencies, though their integration varies significantly across models.

Early adopters such as Flame \cite{flame}, MotionDiffuse \cite{motiondiffuse}, and HMDM \cite{hmdm} all incorporate transformers for denoising, but differ in their conditioning strategies, temporal encoding, and loss functions. Flame \cite{flame} introduces a transformer-based motion decoder in place of the standard U-Net \cite{DBLP:journals/corr/RonnebergerFB15}, using cross-attention to incorporate text features extracted with RoBERTa \cite{roberta}. It introduces two special tokens for encoding motion length and diffusion timestep, both used during cross-attention to guide generation. MotionDiffuse \cite{motiondiffuse}, although similar in structure, handles the timestep differently by sampling it from a uniform distribution, and introduces variable-length motion generation by dividing sequences into sub-intervals. Each segment is paired with a corresponding text description, enabling part-wise and body-part-specific conditioning. It also incorporates Efficient Attention \cite{efficientattention} to reduce computational cost and uses a classical transformer \cite{vaswani2023attentionneed} for text encoding. While both Flame \cite{flame} and MotionDiffuse \cite{motiondiffuse} rely on noise-based reconstruction, Flame adds a variational lower bound. In contrast, MotionDiffuse \cite{motiondiffuse} optimizes only a mean squared error loss on the predicted noise.

In contrast, HMDM \cite{hmdm} takes a different approach by applying its primary reconstruction loss on the denoised signal rather than the noise. It encodes text using CLIP \cite{DBLP:conf/icml/RadfordKHRGASAM21} and feeds the diffusion timestep into the transformer as a dedicated token, similar to Flame. However, HMDM \cite{hmdm} fixes the motion length and introduces a set of auxiliary loss functions designed to improve physical realism: a positional loss in joint space, a velocity loss to enforce temporal consistency, and a foot contact loss defined using forward kinematics. These losses are combined with a standard reconstruction loss, forming a comprehensive objective that better preserves both spatial accuracy and motion smoothness.

Departing from sequential generation, MakeAnAnimation \cite{azadi2023makeananimationlargescaletextconditional3d} proposes a two-stage framework that first pre-trains on a large static 3D pose dataset, created from pose detection applied to image collections, to learn pose-text associations. Using a U-Net architecture \cite{DBLP:journals/corr/RonnebergerFB15} for the denoising network and a pre-trained T5 encoder \cite{raffel2023exploringlimitstransferlearning} for text, the model generates full motion sequences concurrently. Unlike transformer-based models such as HMDM \cite{hmdm} and Flame \cite{flame}, which enforce temporal consistency through specific loss functions, MakeAnAnimation \cite{azadi2023makeananimationlargescaletextconditional3d} avoids such constraints and relies solely on standard diffusion loss. Despite this, it maintains motion continuity through its concurrent sampling and large-scale pre-training strategy.

Recent works have also expanded the diffusion framework to support spatial and semantic constraints. GMD \cite{gmd} highlights that prior models primarily focus on language conditioning while neglecting spatial grounding—specifically, trajectory-level control derived from body-ground contact. To address this, GMD \cite{gmd} proposes a two-stage pipeline. In the first stage, pose representations are re-normalized with respect to ground location, and guidance signals (which are sparse in time) are made more effective by diffusing their gradients across neighboring frames. These enriched spatial signals are then used as conditioning in the generation stage, enhancing trajectory control during motion synthesis.

Using GMD’s idea, OmniControl \cite{omnicontrol} introduces improved spatial guidance and a new realism constraint. It models global body location by cumulatively summing relative pelvis positions and uses pose keyframes as motion anchors. This cumulative structure enables gradient flow to influence all prior frames of the guiding keyframe, while the relative joint representation ensures local joints affect the pelvis in return. However, since pelvis control alone does not sufficiently influence all joints, the model adds a realism guidance component to propagate motion cues more effectively, resulting in more coherent and controllable generated sequences.     

\subsection{Application}
Text-to-motion generation is revolutionizing various sectors by automating the creation of 3D human motion from natural language descriptions. Such automation has enabled the synthesis of realistic and contextually relevant animations, transforming industries such as gaming, film, virtual reality, and healthcare. These systems significantly reduce manual effort, enhance creative flexibility, and provide efficient solutions for generating diverse motion patterns \cite{ahn2018text2action}.

In gaming and film, models like TEMOS \cite{petrovich22temos} and T2M-GPT \cite{t2m-gpt} generate animations for NPCs and pre-visualization. VR and metaverse applications benefit from systems like MotionCLIP \cite{motionclip}, which enhances human character interactions with stylized actions aligned with semantic embeddings. Models trained on biomechanical constraints can visualize complex procedures like joint replacement surgeries or physical therapy regimens. However, current limitations in muscle activation modeling require hybrid AI-expert validation pipelines \cite{T2M-X}, which may be utilized in the healthcare sector. On the other hand, education benefits from instructional animations for sports training and dance lessons. For example, TEACH \cite{athanasiou2022teachtemporalactioncomposition} extends text-to-motion capabilities by handling sequential instructions (e.g., "perform a squat followed by a jump"), making it ideal for creating step-by-step tutorials. 

Despite the transformative potential, challenges remain in generating extended sequences, handling complex spatial and temporal dependencies, and ensuring ethical use. Future research focuses on integrating spatial constraints, improving trajectory control, enhancing realism, and expanding datasets to include diverse motion types \cite{Zhu2024MotionSurvey}. Addressing these limitations will further unlock the potential of text-to-motion generation, driving innovation and setting new benchmarks for efficiency and creativity across industries.

\section{Object}\label{sec_object}
In generative AI, objects serve as fundamental components of digital environments. An object is any distinct, visually representable entity, ranging from everyday items like furniture and vehicles to intricate structures such as architectural elements and animated characters. These objects are essential in animation, gaming, and virtual reality, where they enhance realism and contribute to narrative engagement \cite{silva2021narrative}.
The ability to generate objects dynamically streamlines creative workflows by enabling rapid content creation from minimal input while ensuring stylistic consistency. Whether defining settings, interacting with characters, or enriching immersive storytelling, procedural techniques synthesizing objects at scale are crucial in advancing generative AI applications \cite{ShapeNet, Pix3D}.

\subsection{Dataset}
Translating textual descriptions into 3D representations presents multiple challenges: shape completion, geometry generation, texture synthesis, and quantitative evaluation. Addressing these challenges requires diverse datasets, each tailored to different stages of model development, from training to validation, to ensure accurate shape priors and realistic outputs. Many text-to-3D methods utilize 2D datasets for shape completion, where paired image-text data, often in the form of multi-view images or depth maps, provide essential structural cues for inferring three-dimensional forms. Additionally, text-image datasets are crucial in evaluating semantic alignment, helping researchers determine how well-generated 3D models correspond to their descriptive prompts.

Researchers rely on specialized 3D datasets for geometry and texture synthesis, which contain detailed object representations in formats such as meshes, point clouds, and implicit surfaces. These datasets offer explicit geometric and material properties, enabling models to learn high-fidelity generation of both shape and surface details. In recent years, synthetic datasets that integrate 3D models with rendered 2D views and textual annotations have become instrumental in training multimodal systems, supporting end-to-end learning of shape, texture, and appearance.

While synthetic datasets facilitate large-scale training, real-world datasets such as ScanNet \cite{ScanNet} and Matterport3D \cite{Matterport3D} are essential for evaluating model robustness in practical scenarios. These datasets capture complex indoor environments with real-world occlusions and varied lighting conditions, offering a rigorous benchmark for assessing generalization. Moreover, custom datasets are often developed for specialized tasks, such as shape-guided generation, appearance control, or fine-grained voxel editing, to ensure models are optimized for specific applications. The continuous evolution in the composition and quality of these datasets plays a pivotal role in advancing text-to-3D generation, fostering the development of more reliable and generalizable models. Table \ref{table:object} provides a comprehensive overview of major datasets for this area.

\begin{table*}[t]
\centering
\renewcommand{\arraystretch}{1.4}
\resizebox{\textwidth}{!}{%
    \begin{tabular}{|>{\centering\arraybackslash}p{7cm}|p{7cm}|P{4cm}|c|}
\hline
\multicolumn{1}{|c|}{\textbf{Name}} & \multicolumn{1}{c|}{\textbf{Statistics}} & \multicolumn{1}{c|}{\textbf{Modalities}} & \multicolumn{1}{c|}{\textbf{Link}} \\ \hline

ShapeNet\cite{ShapeNet} & 3D models in categories like furniture and vehicles. & 3D/Point Cloud Data, Text & \href{http://shapenet.org/about}{ShapeNet} \\ \hline

BuildingNet\cite{BuildingNet} & Architectural structures for shape completion tasks. & 3D/Point Cloud Data, Text & \href{https://github.com/BuildingNet/BuildingNet}{BuildingNet} \\ \hline

Text2Shape\cite{Text2Shape} & Textual descriptions linked to ShapeNet categories. & Text, 3D/Point Cloud Data & \href{https://text2shape.github.io/}{Text2Shape} \\ \hline

ShapeGlot\cite{ShapeGlot} & Textual utterances describing differences between shapes. & Text, 3D/Point Cloud Data & \href{https://github.com/alters-mit/ShapeGlot}{ShapeGlot} \\ \hline

Pix3D\cite{Pix3D} & 3D models aligned with real-world images for evaluation. & Images, 3D/Point Cloud Data & \href{https://github.com/xingyuansun/pix3d}{Pix3D} \\ \hline

LAION-5B\cite{schuhmann2022laion} & Large-scale dataset with 5 billion image-text pairs. & Images, Text & \href{https://laion.ai/}{LAION-5B} \\ \hline

COCO-Stuff\cite{COCOStuff} & Annotated images for real-world 3D synthesis. & Images, Text & \href{https://github.com/nightrome/cocostuff}{COCO-Stuff} \\ \hline

Flickr30K\cite{Flickr30K} & Image dataset with diverse textual descriptions. & Images, Text & \href{https://github.com/ubmdmg/Flickr30kEntities}{Flickr30K} \\ \hline

ModelNet40\cite{ModelNet} & 3D CAD models across 40 object categories. & 3D/Point Cloud Data & \href{https://modelnet.cs.princeton.edu/}{ModelNet40} \\ \hline

ShapeNetCore\cite{ShapeNet} & Subset of ShapeNet with detailed object models. & 3D/Point Cloud Data, Text & \href{http://shapenet.org/about}{ShapeNetCore} \\ \hline

BlendSwap\cite{BlendSwap} & Realistic 3D models with physically based rendering (PBR). & 3D/Point Cloud Data, Images & \href{https://www.blendswap.com/}{BlendSwap} \\ \hline

InstructPix2Pix\cite{InstructPix2Pix} & Dataset for instruction-driven image modifications. & Images, Text & \href{https://github.com/timothybrooks/instruct-pix2pix}{InstructPix2Pix} \\ \hline

MagicBrush\cite{MagicBrush} & Dataset for refining texture and appearance in 3D. & Images & \href{https://github.com/OSU-NLP-Group/MagicBrush}{MagicBrush} \\ \hline

NeRF-Synthetic\cite{NeRFs} & 2D images rendered from synthetic 3D scenes. & Images & \href{https://github.com/bmild/nerf}{NeRF-Synthetic} \\ \hline

ScanNet\cite{ScanNet} & 2.5M RGB-D views with semantic segmentations and camera poses. & Images, Text & \href{http://www.scan-net.org/}{ScanNet} \\ \hline

Matterport3D\cite{Matterport3D} & 10,800 panoramic views from 90 building-scale scenes. & Images, 3D/Point Cloud Data, Text & \href{https://github.com/niessner/Matterport}{Matterport3D} \\ \hline
    \end{tabular}
}
\caption{A collection of datasets most frequently used for 3D Object Generation.}
\label{table:object}
\end{table*}

\subsection{Evaluation}
Evaluating generative object models requires both qualitative and quantitative metrics to assess their performance across visual, structural, and computational dimensions. Geometric fidelity ensures that synthesized objects exhibit accurate and plausible forms, often measured using \textit{Chamfer Distance} \cite{cheng2023sdfusion} or \textit{Earth Mover’s Distance (EMD)} \cite{fan2017point}. Models such as DreamFusion \cite{poole2022dreamfusion} and SDFusion \cite{cheng2023sdfusion} demonstrate strong geometric accuracy by leveraging advanced shape synthesis techniques. In contrast, Magic3D \cite{lin2023magic3d} slightly compromises geometric precision in favor of high-resolution texture details, enhancing the generated surfaces' realism.

Another critical aspect of evaluation is semantic alignment, which assesses the correspondence between generated objects and their textual descriptions. This is commonly measured using \textit{CLIP Similarity} \cite{DBLP:conf/icml/RadfordKHRGASAM21}. DreamFusion \cite{poole2022dreamfusion} and Fantasia3D \cite{chen2023fantasia3d} excel in this area, effectively embedding semantic information to produce objects that closely align with the input prompts. Magic3D \cite{lin2023magic3d}, while maintaining a degree of semantic accuracy, places greater emphasis on surface realism and texture refinement rather than strict alignment with textual input.

Textural quality is crucial in generative object evaluation, particularly for models that enhance visual realism. Metrics such as \textit{Structural Similarity Index (SSIM)} \cite{wangzhou2004image} and \textit{Peak Signal-to-Noise Ratio (PSNR)} \cite{wang2004image} are frequently employed to assess texture fidelity. Magic3D \cite{lin2023magic3d} outperforms other models in this regard, achieving exceptionally detailed textures at the cost of increased computational demands. In comparison, models like DreamFusion \cite{poole2022dreamfusion} and Fantasia3D \cite{chen2023fantasia3d} prioritize a balance between semantic accuracy and textural detail, producing visually coherent outputs with moderate computational overhead.

Computational efficiency is another critical factor, especially for real-world applications. Models such as SDFusion \cite{cheng2023sdfusion} adopt efficient processing strategies, achieving a balance between resource demands and synthesis speed. In contrast, Magic3D \cite{lin2023magic3d} delivers superior high-resolution results but at a significantly higher computational cost, making it less scalable for large-scale applications. Meanwhile, IPDreamer \cite{zeng2023ipdreamer} introduces instruction-driven editing, offering greater flexibility but occasionally sacrificing geometric fidelity in the process.

Finally, \textit{Fréchet Inception Distance (FID)} is a widely used metric for evaluating the realism of generated outputs, particularly when comparing generative models that produce 3D objects from 2D views. Fantasia3D \cite{chen2023fantasia3d} and DreamFusion \cite{poole2022dreamfusion} achieve strong FID scores, indicating their ability to synthesize realistic and visually coherent objects that closely match real-world counterparts.

\subsection{Models}
Text-to-3D generation has rapidly evolved through three primary paradigms: diffusion-guided optimization, amortized inference, and high-fidelity geometry-aware synthesis. These paradigms reflect a growing ambition to balance flexibility, speed, and realism in synthesizing 3D content from text. This section traces the key innovations across each, highlighting how models build upon one another and navigate these trade-offs.

\subsubsection{Diffusion-Guided Score Distillation}\label{object_diffusion_guided_score_distillation}

Score Distillation Sampling (SDS) emerged as a turning point in text-to-3D generation. Rather than relying on hand-crafted priors or CLIP similarity, this technique taps into the rich, implicit knowledge of pretrained 2D diffusion models. By minimizing the KL divergence between the forward process of Gaussian noise and the learned score function, SDS introduced a powerful image-space supervisory signal for optimizing 3D representations.

\begin{figure}[t]
\begin{center}
  \includegraphics[width=\textwidth]{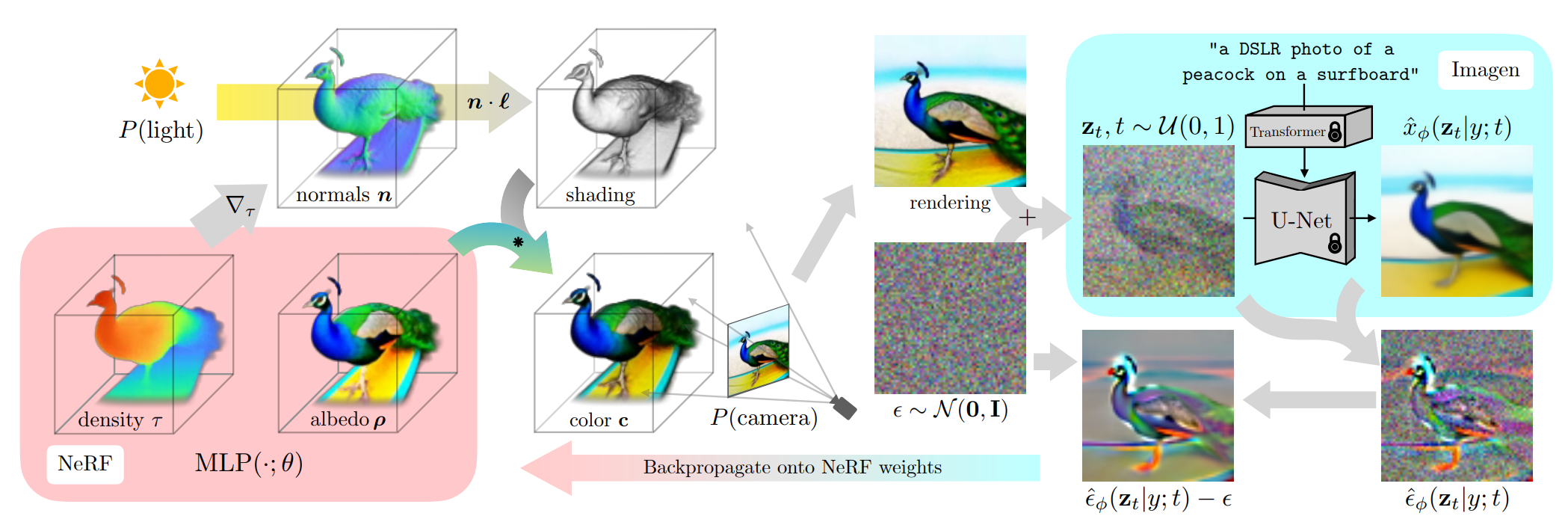}
\end{center}
\caption{
\label{fig:dreamfusion} DreamFusion \cite{poole2022dreamfusion} generates 3D objects from text prompts like \textit{“a DSLR photo of a peacock on a surfboard."} It trains a Neural Radiance Field (NeRF) from scratch for each caption, using shading from normals and a frozen Imagen model \cite{saharia2022photorealistic} to guide updates for improved geometry and appearance. Reprinted from \cite{poole2022dreamfusion}.
}
\end{figure}

This idea first crystallized in DreamFusion \cite{poole2022dreamfusion}, where the authors showed that a NeRF could be steered purely by text via SDS, yielding coherent geometry and textures without any 3D supervision. It has been one of the pioneer models in 3D object generation. Figure \ref{fig:dreamfusion} demonstrates an overview of this approach. Yet, while DreamFusion was groundbreaking, its outputs were often coarse, and the optimization process was long and unstable. Building upon this, subsequent models introduced architectural and procedural refinements. For instance, Magic3D \cite{lin2023magic3d} adopted a two-stage pipeline: a low-resolution NeRF is optimized first, then upsampled and fine-tuned as a textured mesh in a mesh-space latent diffusion framework. This division significantly improved both speed and visual fidelity, enabling 8 times higher resolution supervision and much sharper geometry.

Still, the optimization process remained fragile. One subtle, yet impactful, insight came from DreamTime \cite{huang2024dreamtimeimprovedoptimizationstrategy}, which revisited how timesteps were sampled in SDS. It revealed that uniform timestep sampling, common in early methods, led to inefficient gradients. Instead, aligning the timestep selection with the DDPM sampling schedule resulted in faster convergence and better quality. Another leap came from incorporating multi-scale supervision. HD-Fusion \cite{wu2023hdfusiondetailedtextto3dgeneration} proposed a hierarchical noise estimation and timestep fusion strategy, producing assets with superior resolution and consistency. Combined with structure-preserving priors from ControlNet, SDS-based generation was closer to photorealistic fidelity.

To address structural realism explicitly, Dream3D \cite{xu2023dream3d} incorporated shape priors into the diffusion process by generating rendered-style images and conditioning on shape embeddings. This allowed the optimization to be guided by visual realism and explicit structural cues, leading to significantly more plausible geometries. SDFusion\cite{cheng2023sdfusion}, though often grouped with optimization methods, took a more geometric perspective. Rather than NeRFs or meshes, it represented objects using latent Signed Distance Functions (SDFs). Its diffusion model could condition on multiple modalities, text, images, and partial shapes, while learning to balance these inputs via attention-like weightings. This enabled a wide range of tasks beyond generation, including completion and reconstruction, all while maintaining fine-grained geometric detail.

\subsubsection{Optimization-Amortized Models}\label{object_optimization_amortized_models}

While diffusion-guided optimization yields impressive results, its per-prompt nature makes it ill-suited for large-scale or interactive applications. To overcome this, a new class of models has emerged that amortize the optimization process, learning a direct mapping from text to 3D representation. ATT3D \cite{lorraine2023att3d} was among the first to demonstrate this shift. Instead of optimizing for each prompt individually, it trained on batches of prompts simultaneously, learning a shared model capable of generating diverse 3D assets in a single forward pass. This enabled real-time inference and operations like prompt interpolation and attribute mixing, capabilities out of reach for purely optimization-based methods.

Scaling this idea further, LATTE3D \cite{xie2024latte3d} introduced a larger architecture with 3D-aware priors and enhanced regularization. Leveraging pretrained reconstruction networks and introducing stronger shape constraints produced detailed and robust 3D meshes in less than a second. Its ability to generalize across open-world prompts while preserving visual and structural integrity marked a significant leap in scalability. Interestingly, not all methods fall neatly into either optimization or amortization. IT3D \cite{chen2024it3d} occupies a hybrid space: it retains SDS-based optimization but accelerates and enhances it via a learned refinement module. Using an adversarial discriminator trained on pose-conditioned renderings, this module sharpens geometry and textures, acting as a learned prior over plausible 3D shapes. In doing so, it offers a flexible middle ground and fast, high-quality inference without fully abandoning the benefits of optimization.

\subsubsection{High-Fidelity and Geometry-Aware Models}\label{object_high_fidelity_and_geometry_aware_models}

Beyond generation speed and prompt alignment, another line of research focuses on high-fidelity geometry and realism, emphasizing accurate surface modeling, relightable textures, and physical plausibility. Fantasia3D \cite{chen2023fantasia3d} exemplifies this trend by disentangling geometry and appearance through DMTET-based mesh representations and BRDF-driven material modeling. Instead of directly optimizing RGB outputs, it models surface normals and reflectance, enabling outputs that can be relit or edited realistically. Such separation also supports more controllable generation, where geometry and texture can be manipulated independently.

Editing, too, has emerged as a critical area of focus in recent advancements. Vox-E \cite{sella2023vox} approaches the problem from a 3D volumetric perspective, incorporating a segmentation mechanism driven by cross-attention. This allows users to apply local edits, like changing the color or shape of an object part, while maintaining coherence and structure through 3D-aware regularization. On the frontier of resolution and detail, the Large Gaussian Model (LGM) \cite{10.1007/978-3-031-73235-5_1} brings volumetric Gaussian representations into play. Trained with an asymmetric U-Net and capable of producing high-resolution geometry from minimal inputs (even single images), it supports both image-to-3D and text-to-3D tasks with impressive photorealism. Its differentiable structure also makes mesh extraction straightforward, opening the door to downstream applications in animation and simulation.

Another advanced approach, Meta 3D AssetGen \cite{siddiqui2024meta3dassetgentexttomesh}, combines text-to-image synthesis with physically-based rendering (PBR). It generates shaded and albedo maps for multi-view grids, then reconstructs detailed meshes using signed distance functions (SDF) for geometry and a transformer-based texture refinement module. This produces realistic materials with adjustable lighting properties for high-quality, customizable 3D outputs. Finally, IPDreamer \cite{zeng2023ipdreamer} enhances control for complex image prompts. It extracts detailed appearance features using Image Prompt Score Distillation Sampling (IPSDS). A mask-guided compositional alignment strategy ensures precise geometry-texture coherence, stabilizing outputs even in ambiguous scenarios. This enables robust, high-quality 3D synthesis from challenging prompts.

\subsection{Application}
The rapid progress in text-to-3D object generation is revolutionizing multiple industries by enabling efficient and innovative workflows. In entertainment and gaming, tools such as DreamFusion \cite{poole2022dreamfusion} and Fantasia3D \cite{chen2023fantasia3d} are used to generate detailed 3D assets from text prompts, thereby supporting immersive storytelling and procedural world-building in VR and AR environments. In parallel, the healthcare sector has begun to harness these advances, platforms like Dream3D \cite{xu2023dream3d} yield anatomically accurate 3D models from textual descriptions, which can significantly aid surgical planning, medical training, and personalized patient care.

Meanwhile, in product design and prototyping, approaches like SDFusion \cite{cheng2023sdfusion} and Magic3D (e.g., \cite{lin2023magic3d}) streamline the creation of items ranging from furniture to vehicles by generating designs directly from textual inputs. Educational initiatives also benefit from such technologies; for example, Vox-E \cite{sella2023vox} enhances learning by providing interactive 3D models of historical artifacts and scientific concepts. Furthermore, the ability to generate on-demand 3D renderings from text also offers promising advantages in e-commerce, where improved customer experience and reduced production costs are key considerations. These capabilities are underpinned by foundational datasets and repositories such as ShapeNet \cite{ShapeNet}, BuildingNet \cite{BuildingNet}, and Pix3D \cite{Pix3D}, highlighting the transformative impact of generative 3D modeling across diverse sectors.

\section{Texture}\label{sec_texture}
In generative AI for character animation, texture generation synthesizes detailed surface properties for animated entities, including skin, clothing, and accessories. These textures are essential for defining the visual appearance of characters and conveying realism, emotional expression, and stylized identity. High-quality textures significantly enhance the immersive quality of animated scenes, particularly when aligned with dynamic elements such as motion, lighting, and facial deformation \cite{gao2024genesistexadaptingimagedenoising, bensadoun2024meta3dtexturegenfast}. 

To achieve photorealism and temporal coherence, texture generation must support multi-view consistency and semantic alignment with natural-language descriptions \cite{wang2024consistency2consistentfast3d, cao2023texfusionsynthesizing3dtextures}. Recent models have demonstrated improved texture fidelity and reduced artifacts across complex geometries and animated sequences \cite{chen2023text2textextdriventexturesynthesis, youwang2024paintittexttotexturesynthesisdeep, liu2024vcdtexturevariancealignmentbased}. Stylization and artistic control are further enabled through text-conditioned diffusion frameworks that allow creative re-texturing based on high-level prompts \cite{richardson2023texturetextguidedtexturing3d, gao2024genesistexadaptingimagedenoising}. Traditional methods such as Generative Adversarial Networks (GANs) \cite{goodfellow2014generative} and neural rendering pipelines laid the foundation for modern approaches, which now integrate spatial priors, latent diffusion, and cross-view fusion to produce semantically consistent and expressive textures for animated characters.

Recent advancements have enabled \textit{text-guided texture synthesis}, allowing designers to specify visual features of animated characters using natural language prompts. For instance, systems can generate prompts like ``a velvet jacket with golden embroidery'' or ``robotic armor with worn metallic panels'' and produce coherent textures across the mesh. Methods like TEXTure \cite{richardson2023texturetextguidedtexturing3d}, Text2Tex \cite{chen2023text2textextdriventexturesynthesis}, and TexFusion \cite{cao2023texfusionsynthesizing3dtextures} leverage pre-trained diffusion models conditioned on depth or multi-view geometry to guide the generation process. These models align vision and language through joint embeddings, enabling iterative updates from multiple viewpoints to maintain global consistency. By reducing manual effort and supporting semantic control, such systems transform the way high-fidelity digital characters are textured for animation, games, and interactive media.

\subsection{Dataset}

Effective texture generation relies heavily on the availability of well-annotated, high-quality 3D datasets. In the context of character animation, datasets must capture not only geometry and texture maps but also support semantic alignment through language descriptions, diverse categories (e.g., humans, animals, furniture), and consistent multi-view representations. Attributes such as UV unwrapping quality, mesh resolution, and the presence of paired text prompts are critical for training and evaluating generative models. Moreover, datasets that include complex human scans or stylized assets are especially valuable for validating performance in character-specific applications.

The advancement of generative texture synthesis has been strongly supported by the availability of large-scale, high-quality datasets offering diverse modalities. These typically include \textit{3D geometry}, \textit{surface textures}, and \textit{camera-calibrated multi-view renderings}, complemented by additional data such as \textit{natural language descriptions} or \textit{2D reference images}. Combining these modalities enables models to learn fine-grained spatial representations, semantic consistency, and cross-view alignment, essential for generating expressive and realistic textures for animated characters.

Among the most widely adopted datasets, Objaverse~\cite{deitke2023objaversexluniverse10m3d} provides over 800K textured 3D objects annotated with text descriptions. It supports various applications in generative modeling and serves as a foundational dataset for many state-of-the-art models. ShapeNet~\cite{chen2023text2textextdriventexturesynthesis}, a large-scale structured dataset with category-specific models, offers 3D meshes across various domains, including a dedicated car subset used for benchmarking texture generation. For human-centric tasks, RenderPeople~\cite{youwang2024paintittexttotexturesynthesisdeep} and similar datasets supply high-fidelity scans with detailed anatomical structures and surface properties, enabling rigorous evaluation of textures on realistic geometry.
Public benchmarks such as these play a central role in assessing model performance regarding realism, text alignment, and cross-view consistency. An overview of the most prominent datasets used in texture synthesis research is presented in Table \ref{table:texture_datasets}.

\begin{table*}[t]
\centering
\renewcommand{\arraystretch}{1.4}
\resizebox{\textwidth}{!}{%
\begin{tabular}{|p{5.5cm}|p{7.5cm}|p{3.5cm}|c|}
\hline
\textbf{Name} & \textbf{Statistics} & \textbf{Modalities} & \textbf{Link} \\ \hline

3D-FUTURE~\cite{fu20203dfuture3dfurnitureshape} &
9,992 detailed 3D furniture models with high-resolution textures; 20,240 photorealistic synthetic images across 5,000 diverse scenes. &
3D Geometry, Texture, 2D Images &
\href{https://tianchi.aliyun.com/specials/promotion/alibaba-3d-future}{3D-FUTURE} \\ \hline

Objaverse~\cite{deitke2023objaversexluniverse10m3d} &
Over 800K textured 3D models with natural language descriptions across diverse categories; widely used for training and evaluation in generative tasks. &
3D Geometry, Texture, Language &
\href{https://objaverse.allenai.org/}{Objaverse} \\ \hline

ShapeNet~\cite{chen2023text2textextdriventexturesynthesis} &
Large-scale dataset with category-specific 3D models, including 300 car meshes for texture benchmarking. &
3D Geometry, Texture &
\href{https://shapenet.org/}{ShapeNet} \\ \hline

ShapeNetSem~\cite{yu2023texturegeneration3dmeshes} &
Semantic extension of ShapeNet with 445 diverse annotated 3D meshes for structure-aware evaluation. &
3D Geometry, Texture &
\href{https://shapenet.org/}{ShapeNetSem} \\ \hline

ModelNet40~\cite{richardson2023texturetextguidedtexturing3d} &
Benchmark with 40 categories of 3D CAD models for generalization testing in geometry-aware texture generation. &
3D Geometry &
\href{https://modelnet.cs.princeton.edu/}{ModelNet40} \\ \hline

Sketchfab~\cite{gao2024genesistexadaptingimagedenoising} &
Repository of commercial and scanned 3D models used for qualitative texture evaluation and visualization. &
3D Geometry, Texture &
\href{https://sketchfab.com/}{Sketchfab} \\ \hline

CGTrader~\cite{bensadoun2024meta3dtexturegenfast} &
High-resolution 3D models for qualitative analysis and mesh diversity in text-driven texture synthesis. &
3D Geometry, Texture &
\href{https://www.cgtrader.com/}{CGTrader} \\ \hline

TurboSquid~\cite{gao2024genesistexadaptingimagedenoising} &
The commercial dataset for detailed assets and fine-surface textures is used in high-fidelity mesh evaluations. &
3D Geometry, Texture &
\href{https://www.turbosquid.com/}{TurboSquid} \\ \hline

RenderPeople~\cite{youwang2024paintittexttotexturesynthesisdeep} &
High-quality 3D human scans are commonly used to test text-to-texture models on anatomically realistic meshes. &
3D Scans, Human Meshes &
\href{https://renderpeople.com/}{RenderPeople} \\ \hline

Tripleganger~\cite{youwang2024paintittexttotexturesynthesisdeep} &
Scanned high-fidelity 3D human models are used to evaluate facial and clothing texture realism. &
3D Scans, Human Meshes &
\href{https://triplegangers.com/}{Tripleganger} \\ \hline

Stanford 3D Scans~\cite{youwang2024paintittexttotexturesynthesisdeep} &
High-resolution 3D object scans are used to evaluate generalizations of real-world geometries. &
3D Scans &
\href{http://graphics.stanford.edu/data/3Dscanrep/}{Stanford 3D Scans} \\ \hline

ElBa~\cite{godi2019texelattrepresentingclassifyingelementbased} &
30K synthetic texture images with 3M texel annotations; designed for fine-grained element-based texture analysis. &
2D Texture, Attributes, Spatial Layout &
\href{https://github.com/godimarcovr/Texel-Att}{ElBa} \\ \hline

\end{tabular}
}
\caption{Overview of prominent datasets used in texture generation research, including main statistics, modalities, and reference links.}
\label{table:texture_datasets}
\end{table*}

\subsection{Evaluation}
Evaluating generative models for texture synthesis in 3D character animation necessitates a comprehensive understanding of both visual quality and semantic correspondence with input prompts. Given that textures play a central role in conveying realism, personality, and narrative context in animated characters, evaluation criteria must encompass fidelity, diversity, multi-view consistency, and computational efficiency.

Standard image-based metrics such as Fréchet Inception Distance (FID) and Kernel Inception Distance (KID) are frequently employed to quantify global realism by measuring distributional similarity between generated textures and ground truth images. These metrics have been used extensively in early diffusion-based approaches such as Text2Tex \cite{chen2023text2textextdriventexturesynthesis}, demonstrating significant improvements over GAN-based baselines in both FID and KID scores. As texture realism alone is insufficient in character-focused settings, perceptual diversity is further evaluated using Learned Perceptual Image Patch Similarity (LPIPS), particularly in models like Point-UV Diffusion \cite{yu2023texturegeneration3dmeshes}, which aim to produce varied surface details across character geometry.

User studies are another common evaluation protocol that captures subjective aspects such as visual appeal, coherence, and perceived alignment with textual prompts. These were especially emphasized in models such as TEXTure \cite{richardson2023texturetextguidedtexturing3d}, which balanced fidelity with runtime efficiency and visual consistency across complex character meshes. Paint-it \cite{youwang2024paintittexttotexturesynthesisdeep} introduced a multi-view aware optimization strategy and received high user ratings for realism and material fidelity, while TexPainter \cite{zhang2024texpaintergenerativemeshtexturing} improved visual coherence across camera views through color-space optimization, albeit with a higher runtime cost.

Subsequent models focused more explicitly on runtime and scalability, crucial factors for integration in animation pipelines. For instance, TexFusion \cite{cao2023texfusionsynthesizing3dtextures} demonstrated high-quality textures under significantly reduced inference time (approximately 3 minutes per mesh) while outperforming baseline methods in both FID and user preference. GenesisTex \cite{gao2024genesistexadaptingimagedenoising} and Consistency$^2$ \cite{wang2024consistency2consistentfast3d} employed novel view-fusion and latent-space optimization strategies to achieve superior consistency, outperforming prior baselines in FID, KID, and user preference metrics.

Recent models have adopted text-guided metrics such as ClipFID and ClipScore to evaluate further alignment with prompts, which compute similarity in a joint vision-language embedding space. VCD-Texture \cite{liu2024vcdtexturevariancealignmentbased} utilized these measures alongside traditional metrics to better reflect semantic accuracy and stylistic fidelity in texture generation. It demonstrated superior alignment in text-conditioned scenarios, especially for stylized and expressive characters.

Among recent models, several stand out for their strong all-around performance. VCD-Texture \cite{liu2024vcdtexturevariancealignmentbased} achieves leading scores across realism (FID metric), perceptual alignment (ClipFID and ClipScore metrics), and runtime. Meta 3D TextureGen \cite{bensadoun2024meta3dtexturegenfast} combines fast inference and high resolution output with seamless view consistency, while GenesisTex \cite{gao2024genesistexadaptingimagedenoising} and Consistency$^2$ \cite{wang2024consistency2consistentfast3d} offer strong trade-offs between visual quality and efficiency. These models represent the current state-of-the-art in text-to-texture generation for animated characters, balancing fidelity, semantic control, and deployment readiness.

\subsection{Models}
Text-guided texture generation has rapidly evolved through improvements in model architectures, consistency strategies, and text alignment mechanisms. To provide a clearer overview, we categorize existing models into three major groups: (i) inpainting-based diffusion pipelines, (ii) hierarchical and latent generation models, and (iii) UV-space and variance-aware architectures. This categorization reflects both architectural distinctions and the evolution of key challenges such as multi-view coherence, semantic fidelity, and runtime efficiency.

\subsubsection{Inpainting-Based Diffusion Pipelines}\label{texture_inpainting_based_diffusion_pipelines}

Early efforts focus on masked inpainting techniques that leverage partial geometry and language prompts to synthesize textures in a viewpoint-dependent manner. One of the pioneering approaches~\cite{10203911} uses pseudo-captioning through CLIP similarity to bridge 2D renderings and 3D geometry, enabling semantically aligned texture generation without requiring paired 3D-text data.

\begin{figure}[t]
\begin{center}
  \includegraphics[width=\textwidth]{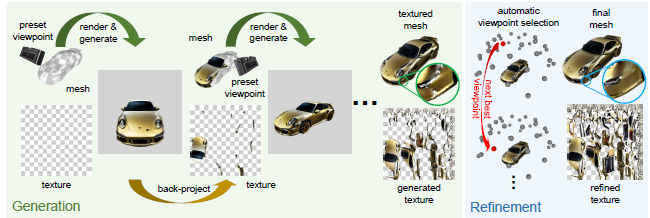}
\end{center}
\caption{\label{fig:Text2Tex}
Pipeline of Text2Tex~\cite{chen2023text2textextdriventexturesynthesis}, featuring a two-stage process for generating high-quality, multi-view consistent textures. Stage I generates textures via depth-to-image diffusion from predefined viewpoints. Stage II refines the results by automatically selecting additional views to correct distortions and artifacts. This approach balances realism, consistency, and efficiency. Reprinted from \cite{chen2023text2textextdriventexturesynthesis}.
}
\end{figure}

Text2Tex~\cite{chen2023text2textextdriventexturesynthesis} introduces a two-stage framework that combines denoising diffusion with depth-to-image translation. In the first stage, initial textures are generated from fixed camera viewpoints, then back-projected into UV space. A second refinement stage iteratively adds new viewpoints to address incomplete coverage and reduce stretching artifacts. This pipeline, shown in Figure~\ref{fig:Text2Tex}, balances quality and efficiency through dynamic view adaptation and geometric alignment.
TEXTure~\cite{richardson2023texturetextguidedtexturing3d} follows a similar paradigm but incorporates a trip-based segmentation strategy. By dividing the surface into regions to generate, refine, or preserve, the model ensures smoother texture transitions and avoids redundant computations during successive denoising passes. Both models establish foundational principles for semantic guidance and partial generation in texture synthesis.

\subsubsection{Hierarchical and Latent Generation Models}\label{texture_hierarchical_and_latent_generation_models}

Subsequent models adopt hierarchical and latent-space generation strategies to overcome the limitations of view-dependency and inference cost. These frameworks prioritize efficiency while maintaining realism and diversity.
Paint-it \cite{youwang2024paintittexttotexturesynthesisdeep} integrates physically-based rendering with U-Net-based reparameterization. A key feature is its use of Score Distillation Sampling (SDS), which aligns texture generation with semantic guidance from CLIP. While it achieves high realism and strong material fidelity, the method demands a long optimization time per mesh, posing scalability challenges.

TexPainter~\cite{zhang2024texpaintergenerativemeshtexturing} advances latent diffusion techniques by operating in color-space embeddings rather than pixel space. It employs a depth-conditioned DDIM~\cite{song2020denoising}, a fast and deterministic sampling method for diffusion models, to denoise latent features across a static set of views. These perspectives are then aggregated into a single unified texture. However, its fixed view configuration limits adaptability in dynamic settings.
TexFusion~\cite{cao2023texfusionsynthesizing3dtextures} further optimizes latent-space generation by introducing the Sequential Interlaced Multiview Sampler (SIMS), a module that interleaves and fuses multi-view features during diffusion. The final texture is reconstructed using a neural color field decoder, significantly reducing runtime while preserving cross-view coherence.

\subsubsection{UV-Space and Variance-Aware Architectures}\label{texture_UV_space_and_variance_aware_architectures}

More recent approaches shift toward operating directly in UV texture space, aiming to resolve inconsistencies caused by view-dependent sampling and rasterization variance. These models fuse multi-view features early in the pipeline to ensure consistent, high-fidelity texture generation across complex geometry.
GenesisTex \cite{gao2024genesistexadaptingimagedenoising} performs cross-view fusion during diffusion by employing a cross-attention mechanism that aligns latent features across different camera perspectives. After texture generation, it applies Img2Img post-processing, a method where a pre-trained image-to-image model refines outputs, to reduce seams and enhance surface detail, particularly in challenging regions.
Building on the theme of efficiency, Consistency$^2$ \cite{wang2024consistency2consistentfast3d} leverages Latent Consistency Models (LCMs) to achieve fast generation with only four denoising steps per view. Disentangling noise and color components enables flexible resolution control while preserving multi-view coherence.

\begin{figure}[t]
\begin{center}
  \includegraphics[width=\textwidth]{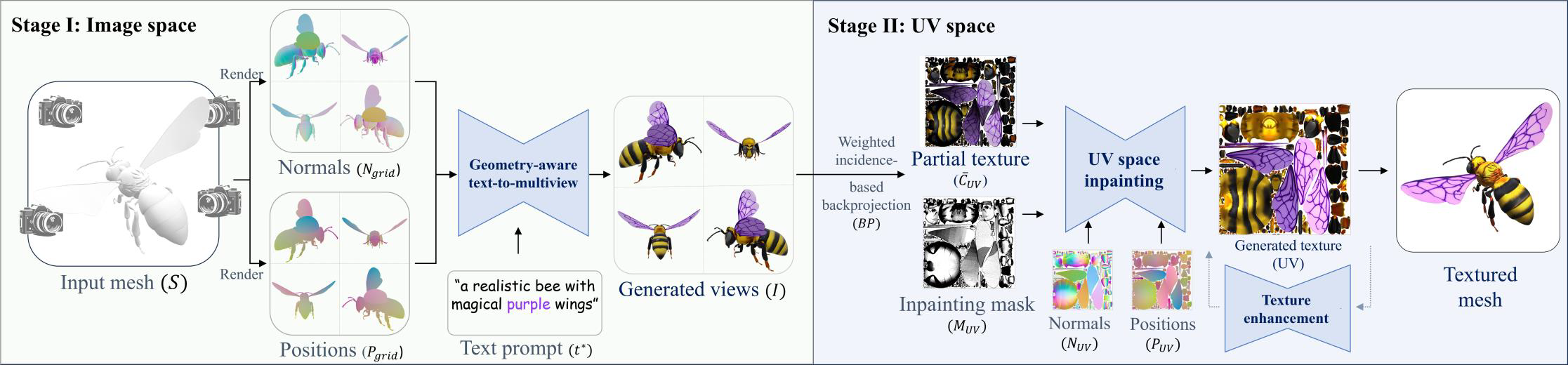}
\end{center}
\caption{\label{fig:Meta3D}
Meta 3D TextureGen~\cite{bensadoun2024meta3dtexturegenfast} employs a two-stage architecture: a geometry-aware diffusion process generates multi-view images, followed by UV-space inpainting using incidence-aware blending. This design enables seamless, high-resolution (up to 4K) textures with minimal artifacts. Reprinted from \cite{bensadoun2024meta3dtexturegenfast}.
}
\end{figure}

Extending these ideas, Meta 3D TextureGen \cite{bensadoun2024meta3dtexturegenfast} introduces a two-stage architecture. A geometry-aware diffusion model generates multi-view renderings conditioned on shape cues in the first phase. In the second, these images are fused and refined in UV space using incidence-aware blending. A patch-based upscaling module allows the system to produce seamless textures at up to 4K resolution. As depicted in Figure \ref{fig:Meta3D}, this pipeline proves robust even in occluded or texture-deficient areas.
In a complementary direction, VCD-Texture \cite{liu2024vcdtexturevariancealignmentbased} focuses on modeling statistical variance across views to mitigate rendering inconsistencies. It introduces Variance Alignment to maintain feature integrity during rasterization, alongside Joint Noise Prediction and Multi-View Aggregation modules that enhance texture fidelity. These strategies make the model particularly effective for stylized characters and geometrically complex assets.

Point-UV Diffusion~\cite{yu2023texturegeneration3dmeshes} employs a coarse-to-fine mechanism where textures are first generated directly on mesh surfaces and then enhanced through 2D diffusion in UV space. This approach separates global structure from fine detail, effectively addressing view misalignment and UV distortion.
Together, these models represent the evolving landscape of text-to-texture generation. From early inpainting pipelines to advanced UV-aware systems, the field has matured toward higher semantic control, greater cross-view realism, and practical efficiency. The advancements mentioned above are critical requirements in today's animated character workflows.

\subsection{Application}
AI-based texture generation has become a key enabler across various creative and industrial domains, significantly impacting animated character design, virtual environments, and digital asset production. In character animation and film, generative models are employed to create semantically aligned textures for skin, clothing, and accessories, which enhances expressiveness and believability. The use of high-fidelity and consistent textures across multiple views plays a critical role in conveying emotion and personality through animated surfaces~\cite{chen2023text2textextdriventexturesynthesis, richardson2023texturetextguidedtexturing3d, xie2021neural}.

In real-time applications such as gaming and virtual or augmented reality (VR/AR), efficiency and multi-view coherence are essential. Recent models support scalable generation of texture assets, significantly reducing manual effort while ensuring stylistic alignment and temporal consistency~\cite{wang2024consistency2consistentfast3d, cao2023texfusionsynthesizing3dtextures, pieters2020survey}. These properties allow rapid customization of 3D characters and environments, improving immersion in interactive platforms.
In digital art and creative prototyping, text-to-texture pipelines allow artists to explore stylistic variations and rapidly iterate on design concepts~\cite{youwang2024paintittexttotexturesynthesisdeep, gao2024genesistexadaptingimagedenoising}. Such workflows enable concept art generation, character ideation, and asset reuse through stylized re-texturing and refinement. These flexible pipelines are especially aligned with the needs of modern creative industries, where rapid ideation and diversity are key~\cite{qi2018deeptexturesynthesis}.

In fields requiring ultra-high-resolution or real-time deployment, models such as Meta 3D TextureGen~\cite{bensadoun2024meta3dtexturegenfast} and VCD-Texture~\cite{liu2024vcdtexturevariancealignmentbased} have been adopted to support immersive AR/VR experiences and realistic simulations. Their ability to reduce artifacts and maintain visual quality under tight computational constraints makes them suitable for deployment in demanding environments.
Beyond entertainment, generative texture techniques are being extended to architecture, medical visualization, and education. In these domains, textured 3D models enable more accurate design prototyping, enhance anatomical detail in training scenarios, and support engaging, interactive storytelling~\cite{gatys2016imagestyletransfer}. Across all these applications, integrating natural-language guidance and fast, high-quality synthesis empowers technical users and artists to produce rich, expressive content with greater efficiency~\cite{lagae2010surveyproceduralnoise}.

\section{Open Problems \& Research Directions}

Despite the rapid advancements in generative AI for character animation, several open challenges remain. Addressing these challenges is crucial for further improving AI-driven animation systems' realism, controllability, and efficiency. This section outlines key unresolved issues and promising research directions in the field.

\subsection{Data Limitations and Ethical Considerations}
The effectiveness of generative models depends on the availability of large-scale, high-quality datasets. However, existing datasets often suffer from biases, limited diversity, and insufficient annotations \cite{gebru2021datasheetsdatasets, Buolamwini2018GenderSI}. Many publicly available datasets primarily focus on specific demographics or animation styles, leading to a lack of generalizability in trained models. 
Moreover, there is a notable lack of comprehensive datasets that cover multiple subfields of character animation, such as gestures, motions, and expressions, which limits the ability to develop models that generalize across different aspects of animation. 
Additionally, privacy concerns and ethical considerations regarding data collection, especially for human facial expressions, gestures, and motion, present significant obstacles \cite{rocher2019reidentification, mahendran2023ethicalmotion}. Future research should explore data augmentation techniques \cite{yang2023imagedataaugmentationdeep},  and federated learning approaches \cite{mcmahan2017federated} to enhance data diversity while respecting privacy constraints.

\subsection{Real-Time Performance and Computational Efficiency}
Current state-of-the-art models, such as diffusion-based architectures and large-scale transformer networks, often require substantial computational resources, limiting their applicability in real-time animation workflows. Optimizing inference speed without compromising animation quality is an essential research direction. Techniques such as model compression \cite{cheng2020surveymodelcompressionacceleration}, quantization \cite{gholami2021surveyquantizationmethodsefficient}, and knowledge distillation \cite{hinton2015distillingknowledgeneuralnetwork} could help reduce computational overhead. Additionally, lightweight generative models designed for real-time applications, particularly in gaming, virtual reality (VR), and augmented reality (AR), remain an open area for further investigation.

\subsection{Controllability and User-Guided Generation}  
A persistent challenge in generative AI for animation is achieving precise, user-controllable outputs. Many current models generate animations in a stochastic manner, making it difficult to fine-tune specific aspects of character movement, facial expressions, or gestures \cite{tang2025generativeaicelanimationsurvey, Canet_Sola_2024}. Future research should focus on integrating more effective control mechanisms, such as prompt-conditioned generation, reinforcement learning with human feedback, and interactive editing interfaces that allow animators to guide AI-generated outputs with minimal manual adjustments \cite{yu2022reinforcementlearningbaseduserguided}.

\subsection{Multimodal and Cross-Domain Integration}
Character animation often involves the integration of multiple modalities, such as speech, motion, and environmental context. While recent advancements in models like CLIP \cite{DBLP:conf/icml/RadfordKHRGASAM21} and ControlNet \cite{zhang2023adding} and also multimodal LLMs \cite{geminiteam2024gemini15unlockingmultimodal, openai2024gpt4technicalreport} have enabled improved multimodal synthesis, achieving seamless synchronization between different data streams in this domain remains a challenge. Future research could explore novel architectures that effectively unify text, speech, motion, and visual inputs, leading to more coherent and context-aware character animations.

\subsection{Robustness and Generalization Across Styles and Domains}
Most generative AI models are trained on specific datasets and struggle with generalizing to unseen styles, artistic directions, or cultural variations in expression. A key research direction is developing models that can adapt across different animation styles, ranging from hyper-realistic rendering to stylized 2D animations, without requiring extensive retraining. Domain adaptation techniques, few-shot learning, and transfer learning approaches may help improve model flexibility across diverse artistic styles and applications \cite{zhang2021surveyunsuperviseddomainadaptation, song2022comprehensivesurveyfewshotlearning}.

\subsection{Evaluation Metrics for Character Animation}

Assessing the quality of AI-generated character animation remains a complex challenge. Common evaluation metrics focus on computational efficiency, perceptual realism, and alignment with ground-truth human motion. Objective metrics include Fréchet Inception Distance (FID) \cite{heusel2018ganstrainedtimescaleupdate} for visual quality, Structural Similarity Index (SSIM) \cite{wangzhou2004image} for texture consistency, and motion-based metrics. However, these metrics do not fully capture the subjective experience of human viewers. Future research should emphasize user studies and perceptual metrics that assess naturalness, emotional expressiveness, and engagement in animation. Additionally, benchmark datasets and standardized evaluation frameworks tailored to generative animation are needed to ensure fair and comprehensive model comparisons.

\subsection{Realism, Identity Preservation, Naturalness, and Interpretability}
While generative AI continues to push the boundaries of realism in character animation, maintaining identity preservation and achieving naturalness remain significant challenges. Facial animations, in particular, need to retain individual identity features while adapting expressions and movements fluidly. Many current models introduce artifacts or fail to maintain personality consistency across sequences. Techniques such as identity-preserving loss functions \cite{xiao2022identity} and adversarial training with perceptual constraints \cite{laidlaw2021perceptual} can help ensure realism without distorting identity characteristics. Furthermore, improving biomechanical accuracy in generated human motion remains an open problem. Models should incorporate physiological constraints and real-world physics-based priors to enhance the believability of synthesized movements.

Additionally, many generative models operate as black boxes, making interpreting or debugging their outputs difficult. This lack of transparency can lead to unpredictable artifacts in animation, limiting trust and adoption by professionals. Future work should focus on explainable AI (XAI) approaches to provide better insights into how models generate animations, allowing animators to fine-tune outputs more effectively \cite{rai2023realisticgenerative3dface, islam2021explainableartificialintelligenceapproaches}.

\subsection{Ethical and Societal Implications of AI-Generated Characters}  
The widespread use of AI-generated characters in entertainment, social media, and virtual environments raises ethical concerns regarding deepfake misuse \cite{Citron2018DeepfakesAT}, identity representation, and cultural sensitivity \cite{articlkke}. The potential for AI to generate hyper-realistic yet fabricated animations calls for developing robust detection mechanisms and ethical guidelines. Research should focus on creating watermarking techniques for AI-generated animations, transparent disclosure practices, and frameworks for mitigating biases in character portrayal.

\subsection{Future Outlook}
The field of generative AI for character animation is set for transformative growth, with revolutionary potential in gaming, film production, virtual assistants, and beyond. These technologies can significantly accelerate the creation of animations, movies, and video games, reducing production times while maintaining high-quality outputs. Additionally, they open up new possibilities for creative expression by enabling artists to prototype and refine ideas more efficiently. Addressing these open challenges will require interdisciplinary collaboration across computer vision, machine learning, human-computer interaction, and digital arts. Future research can unlock new frontiers in AI-driven animation by improving data quality, optimizing computational efficiency, enhancing controllability, and ensuring ethical deployment, making it more accessible, adaptable, and responsible.

\section{Conclusion}
Generative AI is reshaping animation, making it possible to produce lifelike characters and dynamic scenes with unprecedented efficiency and realism. This survey provides a comprehensive overview of generative AI in animation, covering key aspects such as facial expressions, gestures, motion dynamics, avatars, objects, textures, and image synthesis. We present a unified perspective highlighting these technologies' extensive impact by integrating traditionally fragmented domains.

Each section outlines advances in its respective subfield, demonstrating how architectures such as generative adversarial networks (GANs), variational autoencoders (VAEs), transformers, and diffusion models address critical challenges like temporal coherence and multimodal consistency. In addition, we review the datasets used in these animation subfields, the evaluation metrics applied to assess model performance, and the real-world applications of these approaches. 
Beyond surveying recent progress, we provide a foundational background on models, frameworks, and evaluation metrics, equipping researchers with the necessary knowledge to build upon existing work. One central focus of this survey is the role of foundational models, such as CLIP, large language models (LLMs), and diffusion-based techniques, in advancing generative AI for animation. These innovations have enabled more realistic, diverse, and controllable character animations, from identity-preserving facial expression synthesis to complex gesture generation.

Although these innovations have advanced the field, significant challenges remain, including limited robustness and cross-domain generalizations, ethical and societal implications, data constraints, and issues with the synchronization and realism of different parts of the animation. We discuss these open problems and outline promising future research directions to further enhance the capabilities and accessibility of generative AI in animation. This survey is a comprehensive resource for researchers and practitioners, offering insight into current techniques, practical applications, essential background knowledge, and emerging trends. By consolidating knowledge across subfields, we aim to catalyze future innovations in AI-driven character animation, ushering in more advanced, efficient, and creative production pipelines.


\bibliographystyle{IEEEtran}
\bibliography{main}
\newpage
\appendix

\section{Background} \label{sec:app_background}
This section establishes the fundamental concepts, models, and evaluation metrics essential to character animation in generative AI. Since generative approaches to character animation draw from multiple subfields of machine learning and AI, a comprehensive understanding of the underlying principles is crucial. We provide a detailed overview of the foundational models that support various aspects of AI-driven character animation, creating a technical framework for the specialized topics discussed in subsequent sections.
\subsection{Models}

\subsubsection{Language}
Language processing and generation have become central areas of focus in AI research, fueled by the growing demand for natural language interfaces and content generation systems, such as chatbots and virtual assistants. Over the years, language model architectures have advanced significantly, progressing from N-Grams or traditional Recurrent Neural Networks to Transformer-based models and, most recently, to Large Language Models (LLMs). Each generation of models has addressed the limitations of its predecessors while introducing new capabilities, allowing for more accurate and context-aware language understanding and generation \cite{raeniLM}.

\paragraph{Recurrent Neural Networks (RNNs) \cite{rumelhart1985learning}} RNNs are a class of neural networks designed to model sequential data by maintaining a dynamic hidden state that evolves over time. Given an input sequence $\{x_1, x_2, \dots, x_T\}$, the hidden state $h_t$ at each time step $t$ is updated as:
\begin{equation}
    h_t = \sigma(W_h h_{t-1} + W_x x_t + b_h)
\end{equation}
where $W_h$ and $W_x$ are weight matrices, $b_h$ is a bias term, and $\sigma$ is a non-linear activation function, typically $\tanh$. This hidden state $h_t$ encodes information from the current input $x_t$ and the previous hidden state $h_{t-1}$, capturing temporal dependencies in the sequence. The output $y_t$ at each time step is computed using the hidden state:
\begin{equation}
y_t = \phi(W_y h_t + b_y)
\end{equation}
Where $W_y$ is a weight matrix and $\phi$ is an activation function, often softmax, depending on the task.

RNNs are often extended with additional layers (multi-layer RNNs) to capture more complex patterns. Additionally, a popular variant known as bidirectional RNNs (BRNNs) \cite{schuster1997bidirectional} processes the input sequence in both forward and backward directions, maintaining two hidden states, $h_t^{\text{forward}}$ and $h_t^{\text{backward}}$, which are combined to produce the output:
\begin{equation}
y_t = \phi(W_y [h_t^{\text{forward}}; h_t^{\text{backward}}] + b_y)
\end{equation}
This structure leverages both past and future context, significantly improving performance in applications such as speech recognition and machine translation \cite{schuster1997bidirectional}. 

Another widely adopted extension of RNNs is the encoder-decoder framework, which is extensively used in sequence-to-sequence tasks such as machine translation \cite{sutskever2014sequence}. In this architecture, an encoder RNN compresses the input sequence into a fixed-length context vector that is passed to a decoder RNN to generate the output sequence. To mitigate the limitations of encoding all information into a single context vector, attention mechanisms were introduced \cite{vaswani2023attentionneed}. Attention enables the model to dynamically focus on different parts of the input sequence during decoding, thereby enhancing performance on long-sequence tasks.

Despite their strengths, RNNs encounter issues like the vanishing and exploding gradient problem when trained using backpropagation through time (BPTT), limiting their ability to capture long-range dependencies \cite{bengio1994learning}.

\paragraph{Long Short-Term Memory (LSTM) \cite{hochreiter1997long}} LSTM networks are a type of RNN specifically designed to capture long-term dependencies \cite{hochreiter1997long}. LSTMs introduce memory cells $c_t$ that retain information across long periods, with three regulating gates: the input gate $i_t$, the forget gate $f_t$, and the output gate $o_t$. These gates govern the flow of information into and out of the memory cell. The updated equations for an LSTM are as follows:
\begin{equation}
i_t = \sigma(W_i x_t + U_i h_{t-1} + b_i), \quad f_t = \sigma(W_f x_t + U_f h_{t-1} + b_f)
\end{equation}
\begin{equation}
c_t = f_t \odot c_{t-1} + i_t \odot \tanh(W_c x_t + U_c h_{t-1} + b_c)
\end{equation}
\begin{equation}
o_t = \sigma(W_o x_t + U_o h_{t-1} + b_o), \quad h_t = o_t \odot \tanh(c_t)
\end{equation}
where $\sigma$ is the sigmoid activation, and $\odot$ denotes element-wise multiplication. The forget gate $f_t$ controls which information to discard from the cell state, while the input gate $i_t$ manages new information added to the memory cell. The output gate $o_t$ determines which part of the cell state is passed to the hidden state $h_t$. These gating mechanisms allow LSTMs to address the vanishing gradient issue effectively, enabling them to retain and propagate information over long sequences \cite{gers1999learning}.

\paragraph{Gated Recurrent Unit (GRU) \cite{cho2014learning}} GRU is a streamlined variant of the LSTM, providing similar capabilities with a more efficient parameter set \cite{cho2014learning}. GRU’s design merges the input and forget gates into a single update gate $z_t$, and replaces the memory cell with a direct link to the hidden state, simplifying its structure while retaining temporal modeling performance. The GRU update equations are as follows:
\begin{equation}
z_t = \sigma(W_z x_t + U_z h_{t-1} + b_z), \quad r_t = \sigma(W_r x_t + U_r h_{t-1} + b_r)
\end{equation}
\begin{equation}
\tilde{h}_t = \tanh(W_h x_t + r_t \odot U_h h_{t-1} + b_h), \quad h_t = z_t \odot h_{t-1} + (1 - z_t) \odot \tilde{h}_t
\end{equation}
Where $z_t$ serves as an update gate that determines how much of the previous hidden state should be retained, and $r_t$ is the reset gate that manages the influence of prior inputs on the candidate hidden state $\tilde{h}_t$. GRUs are widely used in sequence-based tasks, especially where computational efficiency is a priority \cite{chung2014empirical}.

\paragraph{Attention \cite{vaswani2023attentionneed}}
The attention mechanism has become foundational in the development of advanced machine learning models, especially generative AI systems that synthesize language, motion, and visual data. Its ability to dynamically prioritize relevant portions of the input while generating output has made attention highly effective in tasks like animation generation, gesture synthesis, and avatar modeling, where capturing detailed temporal and spatial relationships is essential.

The core principle behind attention mechanisms is inspired by how humans selectively focus on certain aspects of their surroundings while ignoring others. In machine learning, this mechanism allows a model to weigh the importance of different input elements at various stages of processing. This selective focus has proven especially useful in generating sequential data by capturing dependencies and patterns across time or space.

The attention mechanism was first proposed in the context of neural machine translation (NMT) to address the limitations of the traditional encoder-decoder architecture, where a fixed-length vector representation of the source sentence constrained the ability of the decoder to handle long sequences effectively \cite{bahdanau2016neuralmachinetranslationjointly}.
This mechanism computes a weighted combination of encoder states, enabling the decoder to attend to specific parts of the input sentence during translation. 

Mathematically, attention is modeled by first computing the alignment scores between an encoder hidden state $h_i$ and a decoder hidden state $s_j$. These scores are normalized using a softmax function to produce attention weights, which are then used to calculate a context vector, a weighted sum of the encoder states:
\begin{equation}
\alpha_{ij} = \frac{\exp(e_{ij})}{\sum_{k} \exp(e_{ik})}, \quad c_j = \sum_{i} \alpha_{ij} h_i
\label{eq:attention1}
\end{equation}

where $\alpha_{ij}$ represents the attention weight between the $i$-th input and $j$-th output position, and $c_j$ is the context vector used by the decoder at step $j$. This mechanism allowed models to selectively focus on specific parts of the input, thus improving translation quality.

\paragraph{Self-Attention and Scaled Dot-Product Attention}
A significant advancement in attention is self-attention, which relates different positions within a single sequence to create a richer data representation. In self-attention, the model simultaneously attends to all positions in the input sequence, enhancing its ability to capture long-range dependencies. This method, popularized by the Transformer architecture, removes the need for recurrent or convolutional layers in sequence transduction tasks \cite{vaswani2023attentionneed}. The Transformer’s core component, scaled dot-product attention, computes attention scores by taking the dot product of query ($Q$) and key ($K$) vectors, scaled by the dimension of the key vector $d_k$ to control gradient magnitudes:
\begin{equation}
\text{Attention}(Q, K, V) = \text{softmax}\left(\frac{QK^T}{\sqrt{d_k}}\right) V
\label{eq:attention2}
\end{equation}

Here, $Q$, $K$, and $V$ are the query, key, and value matrices, respectively. The softmax function ensures that the attention weights sum to 1, and the scaling factor $\sqrt{d_k}$ mitigates gradient instability in large $d_k$ values. This mechanism allows the model to focus on different parts of the sequence at each step, enhancing its capacity to capture dependencies regardless of distance within the sequence.

\paragraph{Multi-Head Attention} Multi-head attention extends the attention mechanism by allowing multiple parallel attention heads to capture varied relationships within the input sequence. Each head operates on different projections of the query, key, and value matrices, with outputs concatenated and linearly transformed:
\begin{equation}
\text{MultiHead}(Q, K, V) = \text{Concat}(\text{head}_1, \dots, \text{head}_h) W^O
\label{eq:multihead}
\end{equation}

This allows the model to jointly attend to information from different subspaces at different positions, enabling a more comprehensive representation of the sequence. Multi-head attention has been shown to improve the expressiveness and performance of models in capturing intricate patterns within the data.

\paragraph{Attention in Generative AI} Attention mechanisms enhance the realism and coherence of animated sequences by modeling temporal and spatial dependencies for fluid, contextually accurate movements. Generative models like Variational Autoencoders (VAEs) \cite{kingma2022autoencodingvariationalbayes} and Generative Adversarial Networks (GANs) \cite{goodfellow2014generativeadversarialnetworks} increasingly incorporate attention to higher-quality outputs. For example, in avatar motion synthesis, attention allows the model to focus on key joints or body parts when predicting future movements, leading to more lifelike animations. This is especially important in tasks where avatars need to replicate subtle gestures or emotional expressions, as attention helps the model understand which parts of the body or face are most important in conveying a particular emotion or action.

Attention can be used in a multimodal setting, where multiple streams of data (such as audio, visual, and motion data) are processed in parallel. Multimodal attention mechanisms allow generative models to consider the relationships between different types of input. In a typical multimodal attention setup, each modality (e.g., audio and motion) is first encoded into a separate representation space, and attention is applied across these modalities to capture relevant dependencies. This enables the model to generate animations that are tightly coupled with input signals like voice or facial expression data.

\paragraph{Transformer \cite{vaswani2023attentionneed}} The Transformer model has revolutionized the field of deep learning by significantly improving the efficiency and effectiveness of processing sequential data. Its core innovation is the use of the \textit{self-attention} mechanism. The Transformer consists of an encoder-decoder structure, though variants often use only the encoder (e.g., BERT) \cite{devlin-etal-2019-bert} or decoder (e.g., GPT) \cite{Radford2018ImprovingLU}, depending on the task. In its original design, the encoder processes the input sequence while the decoder generates an output sequence, attending to both the input and previously generated outputs.

Each encoder and decoder layer contains two main components: \textit{multi-head self-attention} and a \textit{feed-forward neural network}, both enhanced by residual connections and layer normalization to stabilize training. Without recurrence or convolution, the Transformer has no inherent sequence order; thus, positional encoding is added to the input embeddings to convey token order. The positional encoding is often implemented using sine and cosine functions of different frequencies, defined as:
\begin{equation}
\text{PE}_{(pos, 2i)} = \sin\left(\frac{pos}{10000^{2i/d_{model}}}\right)
\label{eq:pos_encoding_sin}
\end{equation}
\begin{equation}
\text{PE}_{(pos, 2i+1)} = \cos\left(\frac{pos}{10000^{2i/d_{model}}}\right)
\label{eq:pos_encoding_cos}
\end{equation}

where \(pos\) is the position in the sequence and \(i\) is the dimension. This encoding ensures that the model is sensitive to the order of tokens in the sequence, while also generalizing to input sequences longer than those seen during training.

\paragraph{Transformers in Generative AI}
While Transformers were initially designed for NLP tasks such as machine translation and language modeling, they have been successfully adapted to many other domains. For instance, in computer vision, Vision Transformers (ViTs) have been shown to perform competitively with CNNs for image classification tasks by treating an image as a sequence of patches \cite{dosovitskiy2021imageworth16x16words}.
In the realm of generative AI, the Transformer architecture has also been applied to tasks such as image generation \cite{esser2021tamingtransformershighresolutionimage} \cite{weissenborn2020scalingautoregressivevideomodels}, video generation \cite{weissenborn2020scalingautoregressivevideomodels}, and motion synthesis \cite{petrovich2021actionconditioned3dhumanmotion}. For example, in the generation of animated sequences, Transformer-based models can generate sequences of frames by learning the temporal dependencies between keyframes, thus allowing for the creation of smooth animations.

One particularly exciting application of Transformers is in the domain of avatar generation and motion synthesis for animation. By applying attention mechanisms to sequences of human poses or gestures, Transformers can generate realistic avatar motions that follow specific stylistic or emotional patterns. These models can attend to previous movements to predict future motion, making them particularly suited for tasks where continuity and coherence are essential, such as animating a character’s gestures in response to speech or actions in virtual environments.

\paragraph{BERT \cite{devlin-etal-2019-bert}}
Bidirectional Encoder Representations from Transformers (BERT) represent a major breakthrough in natural language processing. Unlike traditional models that process language either left-to-right (as in GPT) or right-to-left, BERT uses a bidirectional approach, allowing it to consider the context from both directions when making predictions. This has resulted in state-of-the-art performance on a range of NLP tasks such as question answering, named entity recognition, and sentence classification.

The architecture of BERT is based on the Transformer model, specifically employing the encoder portion of the Transformer, with several layers of multi-head self-attention and feed-forward neural networks. BERT is trained using two tasks: \textit{masked language modeling (MLM)} and \textit{next sentence prediction (NSP)}. In MLM, $15\%$ of the input tokens are masked, and BERT is tasked with predicting these missing words based on the surrounding context, making it effective at understanding relationships in a text sequence. NSP is designed to help the model understand relationships between sentences, further improving its ability to handle tasks like document classification and summarization.

The pretraining-finetuning paradigm introduced by BERT has since become a standard in NLP, allowing the model to be fine-tuned on a specific task with relatively little task-specific data. Its success has led to the development of numerous variations, including DistilBERT \cite{sanh2020distilbertdistilledversionbert}, which aims to retain much of BERT’s capabilities while being more resource-efficient.
BERT has become a fundamental tool in modern NLP due to its ability to generalize across a wide range of tasks with minimal adjustments. However, it is computationally intensive and struggles with very long sequences due to its limited input window. Nevertheless, its impact on the field is profound, leading to substantial improvements in performance across many benchmarks.

\paragraph{GPT-1 \cite{Radford2018ImprovingLU}}
Following the advent of the Transformer architecture, the GPT (Generative Pretrained Transformer) family of models, developed by OpenAI, represents some of the most influential advancements in the AI community. The journey began with GPT-1, a 117M-parameter model trained by combining the decoder part of Transformers with unsupervised pre-training \cite{dai2015semisupervisedsequencelearning}. GPT-1 was trained in two stages: first, a Transformer model was trained on a large corpus of data in an unsupervised manner using language modeling as the training objective; then, the model was fine-tuned on smaller, supervised datasets to perform specific tasks.

\textbf{GPT-2 \cite{Radford2019LanguageMA}} GPT-2 is a direct scale-up of GPT-1, featuring over ten times the number of parameters and trained on a dataset more than ten times larger. Specifically, GPT-2 is a 1.5 billion parameter Transformer-based language model trained on 8 million web pages. The increase in both model size and training data led to emergent behaviors such as improved text coherence, enhanced contextual understanding, and better generalization across various domains. GPT-2 maintains the same fundamental objective as GPT-1, predicting the next word in a sequence given all previous words. However, the expanded architecture, with more and wider layers, allows GPT-2 to process information more thoroughly and capture more complex patterns in the data.

\textbf{GPT-3 \cite{NEURIPS2020_1457c0d6} } This marked a transformative moment for AI, boasting 175 billion parameters, making it the largest language model of its time. GPT-3 demonstrated remarkable capabilities in zero-shot, one-shot, and few-shot learning, enabling it to perform tasks such as translation, question answering, and even code generation with little to no task-specific training. GPT-3 laid the groundwork for numerous AI-powered applications, including chatbots and content creation tools. Its vast size allowed it to leverage contextual information more effectively, which became one of its defining strengths. However, GPT-3 also exhibited limitations, including biases and incoherent outputs when handling complex or ambiguous prompts.

The GPT-3.5 model, often referred to as ChatGPT, serves as an intermediary between GPT-3 and GPT-4, designed to enhance conversational capabilities. GPT-3.5 introduced optimized training techniques focused on improving conversational responses and increasing robustness in generating long-form dialogue. It became the core of OpenAI's ChatGPT product, which gained widespread recognition for its application in chatbot systems.

\textbf{InstructGPT \cite{NEURIPS2022_b1efde53}} InstructGPT marked a significant shift in OpenAI’s approach to training language models by incorporating reinforcement learning from human feedback (RLHF) \cite{10.5555/3294996.3295184}. The InstructGPT models are trained to better align their outputs with human instructions, significantly improving the usability and safety of AI systems. Mathematically, the RLHF process combines three critical objectives: (i) a Reward Model (RM) that learns to predict human preferences by comparing pairs of responses to the same prompt. The reward model's loss function is:
\begin{equation}    
{\mathcal {L}}(\theta) = -\mathbb{E}_{(x, y_w, y_l)}\left[\log(\sigma(r_{\theta}(x, y_w) - r_{\theta}(x, y_l)))\right]
\end{equation}
Where \(r_{\theta}(x, y_w)\) and \(r_{\theta}(x, y_l)\) represent the reward model scores for the "winning" (preferred) and "losing" responses, respectively. This objective encourages the model to assign higher scores to preferred responses, aligning it with human feedback. (ii) a KL divergence constraint to maintain proximity to the original supervised fine-tuned model; and (iii) a pretraining objective to preserve general language capabilities. These objectives are combined in the final optimization function:
\begin{equation} 
\text{objective}(\phi) = E_{x \sim D_{RL}}[RM(x, y) - \beta KL(LLM^{RL}_\phi || LLM^{SFT})] + \gamma E_{x \sim D_{pretrain}}[\log LLM^{RL}_\phi(x)]
\end{equation}
where $\beta$ and $\gamma$ are hyperparameters controlling the balance between reward maximization, model stability, and language modeling capabilities. This approach addressed critical issues present in earlier GPT versions, such as hallucinations and the generation of misleading or harmful content. It improved the model's ability to follow instructions and generate more human-aligned responses. The major innovation of InstructGPT lies in its training methodology: instead of solely optimizing for next-word prediction, the model learns to prioritize human-preferred outputs, making it far more reliable for real-world applications where generating helpful and safe content is paramount.


\textbf{GPT-4 \cite{openai2024gpt4technicalreport}} GPT-4 is the latest in the GPT series, representing a significant advancement over its predecessors. GPT-4 is a multimodal model capable of processing both text and images, thereby extending its applicability to a wider range of tasks. It can understand and generate text across various contexts, from simple queries to complex instructions in multiple domains. This multimodal capability enhances GPT-4's utility in areas such as image analysis, document interpretation, and generative design. Key improvements in GPT-4 include better handling of nuanced instructions, improved context retention, and more accurate responses during extended conversations. Additionally, GPT-4 features an expanded context window and improved efficiency in managing ambiguous or complex tasks, making it ideal for professional and academic applications.

Recent developments in the GPT family include the introduction of GPT-4o. This model represents the next generation of multimodal language models, leveraging more advanced fine-tuning techniques and a larger architecture to improve generalization capabilities even further. Although detailed official documentation on GPT-4o is limited, it is believed to feature significant improvements in handling multi-turn conversations, reducing biases, and generating more context-aware, reliable outputs across diverse tasks.

\paragraph{GPT in Generative AI} From GPT-2 onwards, GPT models have found numerous applications in generative AI beyond mere text generation. They have been adapted for various creative tasks, including generating code, music, and visual content when combined with other multimodal models like CLIP. For example, PoseGPT \cite{10.1007/978-3-031-20068-7_24} focuses on pose estimation tasks within the context of video generation. PoseGPT generates human-like pose data that can animate characters in real time, making it useful in applications such as gaming and virtual avatars. Similarly, GestureGPT \cite{zeng2023gesturegpt} extends the GPT framework to generate realistic human gestures based on text or audio input. MotionGPT \cite{NEURIPS2023_3fbf0c1e} is designed for generating motion sequences and has been successfully applied in fields like animation and robotics.

GPT models demonstrate a continuous trajectory of scaling, fine-tuning, and expanding capabilities, from text-only generation to multimodal and context-aware models. They have become indispensable in many domains, and their impact on generative AI continues to grow.

\subsubsection{Vision}

\paragraph{Convolutional Neural Networks (CNNs)}  
CNNs are a specialized form of artificial neural networks designed specifically for image-related tasks. Unlike traditional fully connected networks, CNNs exploit spatial hierarchies in data by leveraging layers structured in three dimensions: height, width, and depth (where depth corresponds to the number of feature maps or channels). Each neuron in a CNN is connected only to a small receptive field of the previous layer, allowing for local pattern recognition and hierarchical feature extraction. This architecture makes CNNs highly effective for tasks such as image recognition, where detecting localized structures like edges and textures is crucial.  
The core operation in CNNs is the convolution, mathematically defined as:  

\begin{equation}
(f * g)(x, y) = \sum_{i=0}^{m-1} \sum_{j=0}^{n-1} f(x-i, y-j) \cdot g(i, j)
\end{equation}  

where \( f \) represents the input image, \( g \) is the convolutional kernel (filter), and \( (x,y) \) are spatial coordinates. Here, \( m \) and \( n \) denote the height and width of the kernel, respectively.  

CNNs consist of three primary types of layers: (i) convolutional layers, (ii) pooling layers, and (iii) fully connected layers. (i) Convolutional layers apply filters over the input to extract important visual features, producing activation maps. (ii) Pooling layers, such as max pooling or average pooling, reduce the spatial dimensions while retaining essential information, thereby improving computational efficiency and reducing overfitting. Finally, (iii) fully connected layers aggregate the learned features for final classification or regression. Through this structured feature transformation, CNNs achieve remarkable performance in tasks such as object recognition \cite{gandarias2019cnn}, segmentation \cite{ajmal2018convolutional}, and image generation \cite{van2016conditional}.

\paragraph{3D CNNs}  
Three-Dimensional Convolutional Neural Networks (3D CNNs) extend the capabilities of traditional CNNs to process volumetric data, such as medical imaging (e.g., MRI scans), volumetric object recognition, and spatiotemporal tasks like action recognition in videos. Unlike standard 2D CNNs, which operate on height and width, 3D CNNs incorporate an additional depth dimension, allowing them to capture spatiotemporal relationships within the data. This is achieved by using 3D convolutional kernels that slide across the height, width, and depth of the input volume.  
The 3D convolution operation is defined as:  

\begin{equation}
(f * g)(x, y, z) = \sum_{i=0}^{m-1} \sum_{j=0}^{n-1} \sum_{k=0}^{p-1} f(x-i, y-j, z-k) \cdot g(i, j, k)
\end{equation}  

where \( f \) is the 3D input volume, \( g \) is the 3D kernel, and \( (x,y,z) \) represent the spatial coordinates of the output feature volume. The dimensions \( m, n, p \) correspond to the kernel size along the three axes.  

3D CNNs are particularly effective for applications requiring an understanding of complex structures and temporal dependencies. Since training 3D models often requires large datasets, techniques such as transfer learning \cite{9533302}, data augmentation \cite{8821676}, and pretraining on related tasks \cite{sun2024contrastive} can help improve generalization. Overall, 3D CNNs provide a powerful approach to processing volumetric and sequential data, making them essential for applications in video analysis, medical imaging, and 3D object detection.

\paragraph{U-Net}  
U-Net is a convolutional neural network architecture originally designed for biomedical image segmentation. It follows a symmetric encoder-decoder structure, where the encoder progressively reduces spatial dimensions while extracting hierarchical features, and the decoder restores spatial resolution to generate detailed segmentation maps. The architecture employs skip connections, which directly link corresponding encoder and decoder layers to preserve fine-grained spatial information, improving the accuracy of the segmentation.  
The skip connections in U-Net can be represented as:  

\begin{equation}
F_l = \text{Concat}(H_l(F_{l-1}), F_{l-1})
\end{equation}  

where \( F_l \) is the feature map at layer \( l \), \( H_l \) represents the transformation applied at layer \( l \), and 'Concat' denotes feature concatenation rather than element-wise addition (as seen in residual networks).  

Due to its ability to efficiently learn from small datasets and capture multi-scale features, U-Net has been widely adopted in various fields, including medical image analysis \cite{neha2024u}, satellite imagery \cite{ulmas2020segmentation}, and autonomous driving \cite{tran2019robust}. More recently, it has become a core component in diffusion models, where it acts as a denoiser by leveraging its skip connections and multi-scale feature extraction to generate high-quality images from noisy inputs.  

\paragraph{Inception \cite{szegedy2015going}} Inception focuses on improving the efficiency and performance of deep convolutional neural networks for tasks like image classification and detection. The key method involves the development of the Inception module, a novel building block that processes multiple spatial scales of information within the same network layer. This module incorporates convolutions of varying sizes (1x1, 3x3, and 5x5) and pooling operations in parallel, allowing the network to capture features at different resolutions simultaneously. One of the major innovations is the use of 1x1 convolutions for dimensionality reduction before applying the larger convolutions, which drastically reduces the computational cost without sacrificing the depth or accuracy of the model.
The architecture allows for stacking multiple Inception modules to create deeper networks while maintaining efficient resource usage. Another feature of the model is the balance between increasing both the depth and width of the network, but controlling computational complexity. The paper applied the Inception architecture in GoogLeNet, a 22-layer network, which uses far fewer parameters than previous models yet achieved superior performance in the ImageNet competition. Additionally, by introducing auxiliary classifiers at intermediate layers, they ensured better gradient propagation during training, which helped prevent vanishing gradients in the deeper parts of the network. This method combines the strengths of traditional convolutional layers with modern techniques like multi-scale processing, making it both scalable and suitable for practical applications \cite{cao2021application}.

\paragraph{VGG \cite{DBLP:journals/corr/SimonyanZ14a}} This primarily focuses on evaluating the impact of increasing the depth of convolutional networks (ConvNets) on their performance for image recognition tasks. To conduct this evaluation, the authors designed a series of ConvNet configurations, varying primarily in the number of convolutional layers, from 11 to 19 weight layers. Each layer uses very small 3x3 convolution filters, which are the smallest receptive fields that can capture directional patterns such as up/down and left/right. By stacking multiple such layers, the networks achieve effective receptive fields equivalent to larger filters (like 7x7) while introducing additional non-linearities through rectification layers (ReLU) and using fewer parameters. This makes the network both more computationally efficient and more expressive in its ability to capture complex visual features.
For training, the networks use a fixed image input size of 224x224 pixels, and the input images are augmented by random crops, horizontal flips, and RGB color shifts. The networks are trained using stochastic gradient descent with momentum, mini-batches of size 256, weight decay regularization, and dropout in the fully connected layers to prevent overfitting. To further improve performance, it describes a strategy of initializing deeper networks using weights from shallower ones, which stabilizes training by preventing gradient instability. At test time, they adopt a fully convolutional approach where fully connected layers are converted to convolutional ones, enabling dense evaluation of images without requiring multiple cropped samples, which speeds up testing. The authors also experiment with multi-scale training and testing, showing that scale jittering (training on images resized to different scales) improves the network's ability to generalize across varying object sizes. 

\paragraph{ResNet \cite{7780459}} ResNet introduces a new residual learning framework designed to address the difficulties of training very deep neural networks. The core method proposed is based on reformulating the learning process so that the network layers learn residual functions rather than directly approximating the desired output. Specifically, the residual learning framework uses shortcut connections, which skip one or more layers, allowing the network to learn the residual mapping $F(x)=H(x)-x$, where  $H(x)$ is the original mapping. The addition of the shortcut connection simplifies the optimization process, as it helps preserve gradient flow during backpropagation. This method addresses the degradation problem, where deeper networks often result in increased training error. It is shown that residual networks (ResNets) not only prevent this issue but also enable the training of extremely deep networks with up to 152 layers. The paper validates their approach through comprehensive experiments on the ImageNet \cite{5206848} and CIFAR-10 \cite{krizhevsky2009learning} datasets. The shortcut connections, implemented as identity mappings, do not add extra parameters or computational complexity, making the method both efficient and scalable. They compare residual networks to plain networks, demonstrating that ResNets are easier to optimize and consistently yield better performance as network depth increases. 

\paragraph{Vision Transformers \cite{dosovitskiy2021imageworth16x16words}} Vision Transformers (ViTs) represent a significant breakthrough in computer vision by leveraging the self-attention mechanism and transformer architecture. While Transformers have become the state-of-the-art architecture for natural language processing tasks, their application to computer vision has been more limited. ViTs apply the attention mechanism to images, previously handled by Convolutional Neural Networks (CNNs), by interpreting an image as a sequence of patches and processing it using a standard Transformer encoder. Specifically, an input image is divided into fixed-size patches, each of which is linearly projected into a feature vector before being passed through the Transformer. This process can be formally expressed as:

\begin{equation}
Z_0 = [x_1 E; x_2 E; ...; x_N E] + E_{\text{pos}}
\end{equation}

where \( x_i \) represents the flattened image patches, \( E \) is the learnable linear projection matrix, and \( E_{\text{pos}} \) denotes the positional encoding. The resulting sequence of patch embeddings is then fed into a Transformer encoder, enabling global context modeling across the entire image. This approach proves effective, especially when followed by pre-training on large datasets. ViTs either match or exceed the state-of-the-art performance on many image classification benchmarks while being relatively efficient to pre-train \cite{han2022survey}.

\subsubsection{Speech}
There are two main tasks in Speech processing: ASR (Automatic Speech Recognition, Speech to Text) and TTS (Text to Speech).
Converting spoken language into written text is called Automatic Speech Recognition. ASR is used in virtual assistants and transcription services \cite{autoSpeechRecognitionSurvey}. Some challenges in ASR are noisy environments, accents, and diverse languages. Text-to-speech (TTS) converts text to spoken output. Naturalness, clarity, and emotions are important factors for the quality of a TTS system.

Transformers have become foundational in speech processing due to their ability to capture long-range dependencies and parallelize computations. In TTS systems like FastSpeech \cite{ren2019fastspeech}, self-attention ensures that each phoneme is matched correctly to its corresponding segment of speech.
FastSpeech \cite{ren2019fastspeech} achieves real-time speech synthesis by replacing autoregressive decoding with non-autoregressive transformers that reduce latency. Multi-head attention makes robustness better by allowing the model to attend to multiple parts of the sequence simultaneously. Whisper \cite{radford2022robust} uses that for multilingual transcription and translation.
Self-supervised models like Wave2Vec \cite{schneider2019wav2vec} and HuBERT \cite{hsu2021hubert} learn from huge amounts of unlabeled audio data, making them more effective for low-resource languages and noisy environments, producing generalizable embeddings that improve tasks like ASR. In multimodal speech systems, models such as Seamless \cite{seamless2023} utilize advanced attention mechanisms, such as Efficient Monotonic Multihead Attention (EMMA), to align spoken words across languages while preserving vocal styles and prosody during translation. Similarly, Tacotron \cite{wang2017tacotron} employs attention-based sequence-to-sequence architectures to ensure robust alignment between text and audio, effectively mitigating issues like skipped or repeated words in synthesized speech.

In the following discussion, we first examine prominent TTS models, including WaveNet, Tacotron, and FastSpeech, which have significantly advanced the synthesis of natural and expressive speech. Next, we delve into key ASR models such as Wave2Vec, HuBERT, and Whisper, which have revolutionized speech recognition through self-supervised learning and multilingual transcription. Finally, we explore Seamless, a versatile model capable of handling multiple speech tasks, including translation and generation.

\paragraph{WaveNet \cite{oord2016wavenet}}
WaveNet introduced a paradigm shift in audio synthesis by directly modeling raw waveforms. Instead of relying on traditional parametric or concatenative methods, WaveNet employed a deep autoregressive approach that generated audio one sample at a time, with each prediction conditioned on all previous samples. As an autoregressive model designed for raw audio waveform generation, it employs dilated causal convolutions to model the temporal dependencies in audio signals, ensuring that predictions at each time step tt
t depend only on past time steps, thereby avoiding any leakage of future information:
\begin{equation}
y(t) = f(x(t-1), x(t-2), \ldots, x(t-d))
\end{equation}
where 
d is the dilation factor, enabling the receptive field of the model to grow exponentially with depth.

The architecture leveraged these dilated causal convolutions to dramatically expand the receptive field while maintaining computational efficiency. This allowed the model to capture long-range dependencies in audio signals, which was crucial for producing natural-sounding speech. WaveNet also modeled audio as a probability distribution rather than deterministic values, allowing it to capture the natural variations present in human speech.
WaveNet can generate realistic and natural-sounding speech better than parametric and concatenative synthesis methods, and the use of dilated convolutions allows the model to capture long-range dependencies efficiently. Wavenet was a paradigm shift from feature-based parametric approaches to raw waveform synthesis and enabled higher fidelity and expressiveness in speech. Its flexibility meant it could be conditioned on different inputs, making it adaptable for various applications, including text-to-speech, music generation, and voice conversion. However, high latency, resource requirements, and computational inefficiency remained problems in WaveNet that would be addressed in later iterations.

\paragraph{Tacotron \cite{wang2017tacotron} and Tacotron 2 \cite{shen2018natural}}
Tacotron represented the first fully end-to-end neural text-to-speech system that went directly from text to spectrograms without requiring handcrafted linguistic features. It employed a sequence-to-sequence architecture with attention mechanisms to align text and audio features, effectively learning the complex mappings between written language and spoken sounds. The system consisted of an encoder-decoder framework with a text encoder processing character-level input and a spectrogram decoder generating the audio representation. This approach eliminated the need for sophisticated linguistic front-ends that were previously required in TTS systems. Tacotron's ability to work directly with characters rather than phonemes or other linguistic units simplified the pipeline and reduced the expertise needed to build speech synthesis systems, democratizing access to TTS technology.

Tacotron 2 enhances the original Tacotron approach by generating mel-scale spectrograms and using WaveNet \cite{oord2016wavenet} to produce the final waveform. This combined Tacotron's text-to-spectrogram model with a modified WaveNet for waveform generation, creating a fully neural end-to-end system that significantly raised the bar for synthetic speech quality. The model featured an improved attention mechanism that reduced alignment errors and produced more consistent speech. This integration resulted in substantially higher audio fidelity that often approached human-level naturalness in controlled settings. Tacotron 2 excelled at modeling prosodic elements such as intonation, stress, and rhythm, capturing the nuanced characteristics that make speech sound authentic. Its more streamlined architecture was easier to train and resulted in fewer artifacts and errors compared to its predecessor. However, Tacotron 2 introduces computational inefficiencies and higher resource demands due to its reliance on WaveNet. The success of Tacotron 2 demonstrated that neural approaches could handle the entire TTS pipeline effectively, setting a new standard for speech synthesis that has influenced virtually all subsequent research in the field.

\paragraph{FastSpeech \cite{ren2019fastspeech} and FastSpeech 2 \cite{ren2021fastspeech}}
FastSpeech uses a non-autoregressive approach for text-to-speech synthesis, overcoming the inference latency issues of autoregressive models like WaveNet \cite{oord2016wavenet}. It replaces RNNs with transformers for parallel processing to achieve speedup during inference. FastSpeech 2 improves upon its predecessor by introducing variance predictors for pitch, energy, and duration. FastSpeech enables real-time synthesis by using parallel generation of speech spectrograms. Also robust alignment mechanism reduces skipped or mispronounced words. If we want to say a disadvantage in FastSpeech is that speech quality does not fully match autoregressive approaches like WaveNet \cite{oord2016wavenet}.

\paragraph{Wave2Vec \cite{schneider2019wav2vec} and Wave2Vec 2 \cite{baevski2020wav2vec}}
Wave2Vec \cite{schneider2019wav2vec} is a self-supervised learning framework designed to extract speech representations directly from raw audio, reducing the need for labeled data. By learning from large-scale unlabeled audio, it produces generalizable representations that enhance ASR across different languages and accents. Wave2Vec 2 builds on this approach by integrating quantization and contextualized embeddings, further improving ASR performance. Additionally, it helps bridge the gap between supervised and unsupervised ASR, making high-quality speech recognition more accessible, particularly for low-resource languages.
The key innovation of Wave2Vec lies in its ability to learn useful speech representations from raw audio in a self-supervised manner. The original Wave2Vec architecture consists of a multi-layer convolutional encoder that processes raw waveform inputs, followed by a context network that captures broader contextual information. The model is trained using a contrastive predictive coding objective, where it learns to identify the correct future audio sample among a set of negative examples.

Wave2vec 2 \cite{baevski2020wav2vec} expands on this foundation with several architectural improvements. It introduces a quantization module that discretizes the latent representations, creating a more robust feature set. The model incorporates a Transformer-based context network, replacing the convolutional approach of the original, which allows for capturing longer-range dependencies in the speech signal. Wave2vec 2 also employs a more sophisticated contrastive learning objective with masked prediction, where portions of the latent speech representations are hidden, and the model must predict these masked regions from the surrounding context.
These innovations enable Wave2Vec 2 to achieve state-of-the-art results with minimal labeled data, democratizing access to high-quality speech recognition systems across diverse languages and reducing the resource gap in speech technology development.

\paragraph{HuBERT \cite{hsu2021hubert} }
HuBERT extends Wave2Vec \cite{schneider2019wav2vec} by incorporating clustering-based pseudo-labeling and BERT-style masked prediction. Speech features are initially clustered using techniques like K-means, and these clusters serve as training targets for self-supervised learning. The model then generates discrete representations of audio signals based on these clusters. HuBERT improves its understanding of speech by randomly masking portions of the input and training itself to predict the masked segments using contextual cues. Through self-supervised pretraining on raw audio, it learns rich speech representations.
HuBERT outperforms Wave2Vec 2 in ASR benchmarks, demonstrating its effectiveness in speech recognition tasks. One of its key advantages is its reduced dependency on labeled data, making it particularly useful for low-resource languages. However, this comes at the cost of high computational requirements, which can be a limitation in resource-constrained environments.

\paragraph{Whisper \cite{radford2022robust}}
Developed by OpenAI, Whisper \cite{radford2022robust} is a transformer-based model designed for multitask speech processing, including ASR, translation, and transcription. Its main innovation lies in its weakly-supervised training approach, utilizing 680,000 hours of web-collected audio paired with transcripts, eliminating the need for human-labeled datasets. Architecturally, Whisper follows an encoder-decoder transformer design where the encoder processes log-mel spectrograms through a convolutional layer and transformer blocks, while the decoder generates text tokens autoregressively with task-specific prompts. Trained on a large-scale multilingual corpus covering 96 languages and 117,000 hours of non-English audio, Whisper demonstrates robust performance across various domains without requiring language-specific fine-tuning. Its zero-shot transfer learning capability allows it to perform well on unseen languages and tasks, while effectively handling noisy environments and diverse speech patterns. 

Available in multiple sizes from the compact "tiny" model (39M parameters) to the comprehensive "large" variant (1.55B parameters), Whisper incorporates special tokens for timestamps, punctuation, and language identification, enhancing its utility for practical applications. However, despite being one of the most versatile speech processing models available, its large model size and computational requirements may pose challenges for deployment on resource-limited devices.

\paragraph{Seamless \cite{seamless2023}}
Seamless has the ability of universal speech translation and generation and aims to create a comprehensive, end-to-end speech translation system that enables real-time multilingual communication. Seamless integrates several key components: a speech encoder that converts audio into a universal representation, a text encoder for processing written language, and a shared decoder capable of generating both text and speech outputs. The model incorporates SeamlessM4T, a transformer-based neural architecture that handles translation across multiple modalities \textit{(speech-to-speech, speech-to-text, text-to-speech, text-to-text)}, enabling a single unified model rather than separate cascaded systems. 

Seamless introduces the SeamlessExpressive module for emotional preservation, capturing the speaker's original emotional context through prosody modeling and specialized training on expressive speech data. A key innovation of Seamless is its UnitY framework, which employs discrete speech units as an intermediate representation between speech and text, allowing for more efficient training and inference while maintaining high fidelity. The model architecture also enables joint training across 101 languages with a shared vocabulary of speech units and text tokens, facilitating zero-shot translation between language pairs never encountered during training.

Seamless employs attention mechanisms within its Transformer-based architecture to determine which parts of the input are most critical for accurate translation and generation, while its advanced SONAR acoustic encoder \cite{duquenne2023sonar} provides robust feature extraction across diverse acoustic conditions. These innovations lead to reduced word error rates compared to other models across various benchmarks, as well as significantly lower latency. SeamlessStreaming, for instance, delivers near real-time translation with latencies ranging from under a second to just a few seconds, depending on the language pair and configuration.

\subsubsection{Temporal Sequence Modeling}

\paragraph{Temporal Convolutional Networks (TCNs) \cite{TCN}}
TCNs is a neural network architecture designed specifically for modeling sequential data. TCNs employ causal convolutions, ensuring that predictions at each time step $t$ rely solely on inputs from preceding time steps, avoiding any leakage of future information. This property makes TCNs particularly effective for time series tasks, sequence modeling, and autoregressive forecasting.
\begin{equation} 
y(t) = \sum_{i=0}^{k-1} w(i) \cdot x(t - i)
\end{equation}

To capture long-range dependencies, TCNs leverage dilated convolutions, allowing the receptive field of the network to expand exponentially with network depth while maintaining computational efficiency:
\begin{equation} 
y(t) = \sum_{i=0}^{k-1} w(i) \cdot x(t - d \cdot i)
\end{equation}

Where $d$ is the dilation factor that controls the spacing of input gaps.

In addition to preserving sequence structure, TCNs offer advantages over RNN-based models, including the ability to parallelize training and a reduced risk of vanishing or exploding gradients, which are common issues in deep recurrent networks. This efficient structure allows TCNs to perform well in long-sequence tasks, making them a robust choice for temporal sequence modeling. TCNs have also gained attention in generative AI applications, where sequential dependencies are crucial. By leveraging the ability to handle long sequences effectively, TCNs are increasingly used for generating time-dependent outputs, such as audio synthesis, text generation, and video frame prediction.

\paragraph{Transformer-XL \cite{dai2019transformerxlattentivelanguagemodels}} Transformer-XL is an extension of the Transformer model that addresses its limitations in handling long sequences. Standard Transformers are limited by fixed-length context windows, which often truncate long dependencies. Transformer-XL overcomes this by introducing a segment-level recurrence mechanism that allows it to reuse hidden states from previous segments, enabling long-range dependency modeling over extended contexts. This recurrent mechanism makes Transformer-XL particularly effective in natural language processing tasks like text generation and language modeling, where it excels in capturing semantic dependencies across distant words. Moreover, Transformer-XL achieves significant improvements in both memory efficiency and performance, allowing it to handle much longer sequences than traditional Transformers.

\paragraph{ConvLSTM \cite{shi2015convolutionallstmnetworkmachine} } ConvLSTM combines convolutional neural networks (CNNs) with Long Short-Term Memory (LSTM) units to capture both spatial and temporal dependencies, making it well-suited for data with spatiotemporal structures, such as video or radar sequences. Unlike traditional LSTMs, which process sequences without considering spatial dimensions, ConvLSTM applies convolutions within the LSTM cell itself. This enables the model to effectively capture spatial patterns in data while also learning temporal dynamics. ConvLSTM has been successfully applied in applications like precipitation nowcasting, where both time-dependent and location-specific features are crucial for accurate predictions. By leveraging the strengths of both CNNs and LSTMs, ConvLSTM enhances the modeling of complex temporal sequences in spatial data.

\subsubsection{Computer Graphics}

\paragraph{SMPL \cite{SMPL}} Skinned Multi-Person Linear model (SMPL) is a popular parametric model that represents the 3D geometry of the human body. It provides a compact, low-dimensional representation of body shape and pose, making it highly efficient for tasks like 3D body reconstruction, animation, and human motion capture. The SMPL model is based on a combination of linear blend skinning (LBS) and principal component analysis (PCA), which allows it to model the surface of the human body with a small number of parameters, namely: 
shape parameters \( \boldsymbol{\beta} \in \mathbb{R}^{|\boldsymbol{\beta}|} \) that define variations in body shape, capturing different body types; and pose parameters \( \boldsymbol{\theta} \in \mathbb{R}^{|\boldsymbol{\theta}|} \) which encode the rotations of the body’s joints to represent the pose.

The body surface is then defined as:
\begin{equation}
\mathbf{T}_{\text{pose}}(\boldsymbol{\beta}, \boldsymbol{\theta}) = \mathbf{T}_{\text{shape}}(\boldsymbol{\beta}) + \mathbf{T}_{\text{pose}}(\boldsymbol{\theta})
\end{equation}
where \( \mathbf{T}_{\text{shape}} \) models the neutral body shape and \( \mathbf{T}_{\text{pose}} \) defines the deformations due to pose.

\paragraph{SMPL+H \cite{MANO:SIGGRAPHASIA:2017}} This is an extension of the SMPL model that incorporates hand modeling, allowing for a more detailed representation of human hands in addition to the body. It introduces additional parameters for the hand joints, \( \boldsymbol{\theta}_{\text{hands}} \in \mathbb{R}^{|\boldsymbol{\theta}_{\text{hands}}|} \), which encodes the pose of the hand joints.
This extension makes SMPL+H suitable for scenarios where detailed hand movements and gestures are important, such as gesture recognition and human-object interaction tasks.

\begin{figure}
    \centering
    \includegraphics[width=0.75\linewidth]{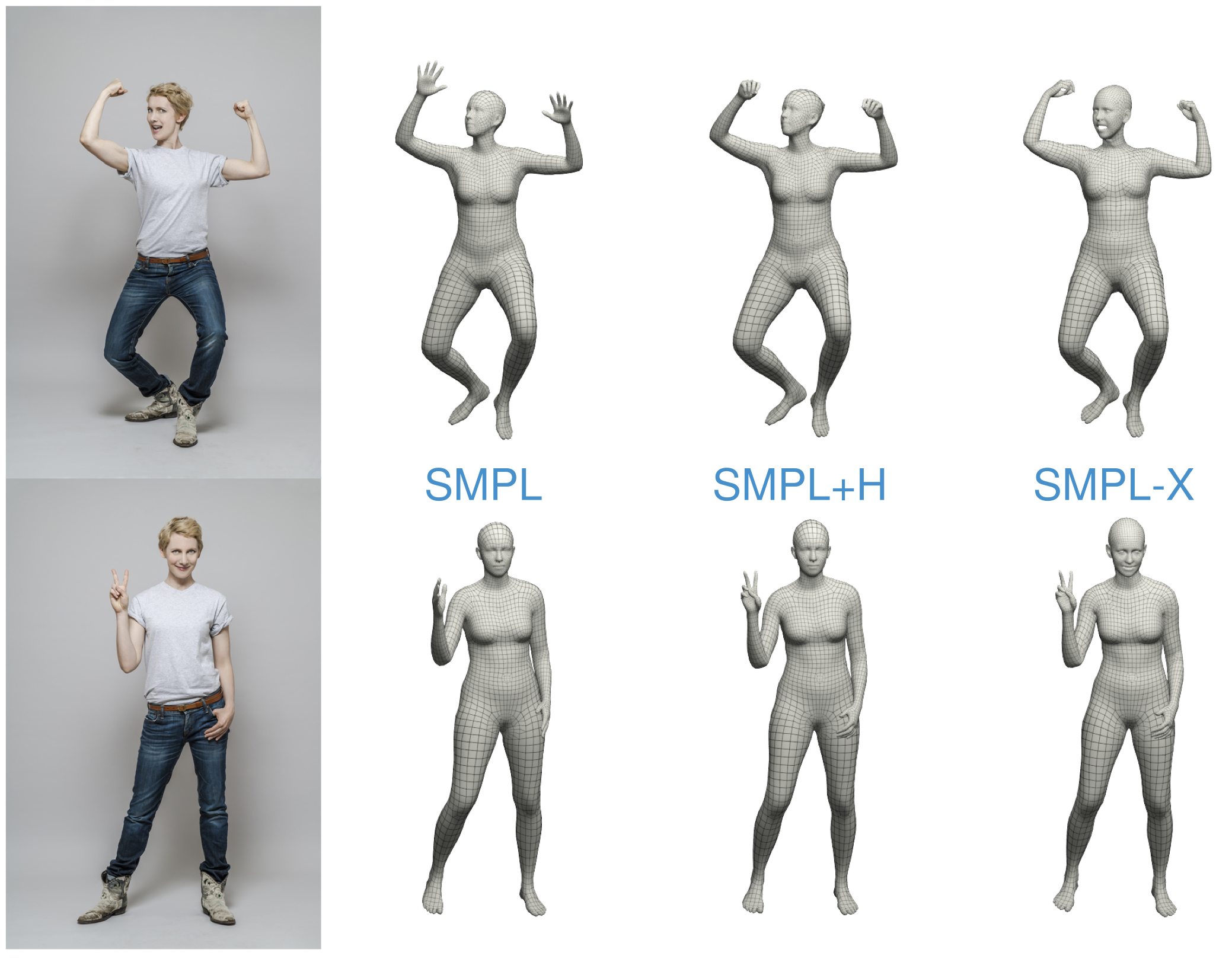}
\caption{Comparison of SMPL \cite{SMPL}, SMPL+H \cite{MANO:SIGGRAPHASIA:2017}, and SMPL-X \cite{SMPL-X:2019} models: SMPL captures basic body shapes, SMPL+H adds detailed hand poses, and SMPL-X includes body, hands, and facial expressions for a more complete and realistic representation. Reprinted from \cite{SMPL-X:2019}.}
    \label{fig:smpl}
\end{figure}

\paragraph{SMPL-X \cite{SMPL-X:2019}} SMPL-X further extends SMPL+H by incorporating facial expressions and more detailed hand modeling, allowing for full-body human modeling, including facial expressions and hand gestures. It introduces additional parameters for the face and hands. First, the expression parameters \( \boldsymbol{\psi} \in \mathbb{R}^{|\boldsymbol{\psi}|} \) that model facial expressions. Second, the hand pose parameters \( \boldsymbol{\theta}_{\text{hands}} \in \mathbb{R}^{|\boldsymbol{\theta}_{\text{hands}}|} \) which encode the pose of the hand joints (same as in SMPL+H).
SMPL-X is ideal for tasks that require detailed human motion and interaction, such as virtual avatars, human-object interaction modeling, and human-computer interaction. Its ability to model both the body and expressive movements of the face and hands makes it a versatile tool in 3D animation and reconstruction. Figure \ref{fig:smpl} illustrates the SMPL, SMPL+H, and SMPL-X models generated from the same input images, highlighting the differences in body, hand, and facial detail representation.

SMIL (Skinned Multi-Infant Linear Model) \cite{hesse2018learningtracking3dbody} and SMAL (Skinned Multi-Animal Linear Model) \cite{Zuffi:CVPR:2017} are advanced 3D body models designed to overcome challenges in capturing and analyzing non-cooperative subjects. To begin with, SMIL is the first 3D body model developed specifically for infants, aimed at addressing the lack of high-quality scans by learning from the low-quality incomplete RGB-D data of freely moving infants. It provides a basic tool for the General Movement Assessment and other applications for early detection of neurodevelopmental disorders. In contrast, SMAL focuses on animals, representing a new approach to learning 3D articulated models from a few scans of toy figurines in various poses. SMAL is a novelty for aligning scans of varied shapes and sizes to a common template, hence allowing the creation of a shape space for animals such as lions, horses, and dogs. Both SMIL and SMAL are examples of breakthroughs in modeling challenging subjects that have opened up detailed analysis and animation in their respective domains.

Such models can be applied to creating realistic virtual avatars, character animation, simulation of human interactions, and visualization of human poses. For instance, AI models trained to learn human pose could directly use detailed data to improve performance. Generative AI systems can create highly detailed representations of humans, including intricate hand and face details, and find application in VR and gaming. The low dimensionality makes them ideal for seamless integration and allows for efficient generation and manipulation of 3D human data. The SMPL model helps these systems create and understand full-body motion and makes it possible to visualize motions properly by generating animations. When AI applications need hand interactions, SMPL+H helps them train models for creating realistic hand interactions and handling gesture tasks. Moreover, SMPL-X brings face, hand, and body parts together and makes it possible for applications to use these factors to generate a human body model realistically.

\subsubsection{Multimodal}

\paragraph{CLIP \cite{DBLP:conf/icml/RadfordKHRGASAM21}}
The introduction of the CLIP model represents a pivotal moment in the evolution of multimodal foundation models. CLIP, which stands for Contrastive Language-Image Pre-training, is a joint image-text embedding model trained on 400 million image-text pairs. It encodes images and text into the same embedding space by leveraging the power of contrastive learning. For instance, an image of a cat and the caption "This is a cat" will have highly similar embeddings after being processed by CLIP’s respective image and text encoders. The model consists of two main components: a text encoder, which utilizes a Transformer \cite{vaswani2023attentionneed}, and an image encoder, which employs either the Vision Transformer (ViT) \cite{dosovitskiy2021imageworth16x16words} or ResNet-50 \cite{7780459}. These pre-trained encoders enable CLIP to perform well without relying on a complex architecture.

The primary innovation of CLIP lies in its contrastive loss algorithm \cite{nakada2023understandingmultimodalcontrastivelearning}. Given a batch of $N$ (image, text) pairs, CLIP computes a dense cosine similarity matrix between all $N \times N$ possible (image, text) candidates within the batch. The text and image encoders are jointly trained to maximize the similarity between $N$ correct pairs of (image, text) associations while minimizing the similarity for $N \times (N-1)$ incorrect pairs via a symmetric cross-entropy loss. The image embeddings $\mathbf{I}_e$ and text embeddings $\mathbf{T}_e$ are obtained by projecting the output of the image encoder $\mathbf{I}_f$ and text encoder $\mathbf{T}_f$ into a shared multimodal space using learned projection matrices and normalizing them to unit length:
\begin{equation}
\mathbf{I}_e = \frac{\mathbf{I}_f \mathbf{W}_i}{\|\mathbf{I}_f \mathbf{W}_i\|_2}, \quad
\mathbf{T}_e = \frac{\mathbf{T}_f \mathbf{W}_t}{\|\mathbf{T}_f \mathbf{W}_t\|_2}
\end{equation}

Here, $\mathbf{I}_f$ and $\mathbf{T}_f$ are the features extracted by the image and text encoders, respectively, and $\mathbf{W}_i$ and $\mathbf{W}_t$ are learned projection matrices that map the features to a shared embedding space. The normalization ensures that the embeddings $\mathbf{I}_e$ and $\mathbf{T}_e$ lie on a unit embedding space, allowing the model to compare them using cosine similarity. Next, the similarity matrix $\mathbf{S}$ is calculated by taking the dot product of the normalized image and text embeddings and scaling it by a learnable temperature parameter $t$:
\begin{equation}
\mathbf{S} = \exp(t) \cdot (\mathbf{I}_e \mathbf{T}_e^\top)
\end{equation}
The temperature parameter $t$ controls the sharpness of the distribution of similarities. When higher, it reduces sensitivity to small differences, while a lower temperature sharpens that sensitivity.

Finally, the symmetric contrastive loss is computed by applying cross-entropy loss on both the image-to-text and text-to-image directions. The loss, applied symmetrically, is defined as:
\begin{equation}
\mathcal{L} = \frac{1}{2} \left( \frac{1}{n} \sum_{i=1}^{n} \text{CrossEntropy}(\mathbf{S}_{i,:}, \text{labels}) + \frac{1}{n} \sum_{j=1}^{n} \text{CrossEntropy}(\mathbf{S}_{:,j}, \text{labels}) \right)
\end{equation}
Where $\mathbf{S}_{i,:}$ refers to the similarities between the $i$-th image and all text embeddings and $\mathbf{S}_{:,j}$ refers to the similarities between the $j$-th text and all image embeddings. The $\text{labels}$ vector represents the correct image-text pairings within the batch.

\paragraph{CLIP in Generative AI} CLIP’s multimodal capabilities have been transformative in generative AI, enabling models to generate content that integrates textual and visual information in meaningful ways \cite{10.1145/3503161.3547910}. For example, in avatar generation, CLIP can ensure that the visual output aligns with textual descriptions of gestures, expressions, or even personality traits, leading to more personalized and context-aware avatars \cite{10.1145/3528223.3530094}. 
Moreover, CLIP can cross-reference text, images, and even videos in its extended versions \cite{Wang_Liu_Jiao_Wang_Hao_Li_Li_Chen_Liu_2024}. This capability guarantees that generated content remains semantically consistent with the provided instructions. It encompasses various elements, including visual gestures, motion patterns, and detailed facial expressions \cite{10310257}. As a result, user experiences are significantly enhanced in interactive environments such as gaming, virtual meetings, and animation.

One of CLIP’s most remarkable features is its ability to perform zero-shot learning, allowing it to make predictions for unseen categories or domains without explicit training \cite{DBLP:conf/icml/RadfordKHRGASAM21}. This zero-shot capability has been demonstrated across various domains, such as image classification, where CLIP can classify images solely based on textual descriptions. This generalization power is critical in generative AI systems where models must synthesize content for a broad range of inputs and tasks. Notably, CLIP has also inspired subsequent models like DALL-E \cite{pmlr-v139-ramesh21a}, which extends CLIP's framework to generate high-quality images directly from textual descriptions, marking a significant development in the field of multimodal generative AI.

\paragraph{GAN \cite{goodfellow2020generative}}
Generative Adversarial Networks (GANs) consist of two neural networks: a \textit{generator (G)} and a \textit{discriminator (D)}, which engage in an adversarial, game-theoretic process. The generator \( G(z; \theta_g) \) takes noise \( z \) sampled from a prior distribution \( p_z(z) \) and maps it to a synthetic data point in the target space, aiming to mimic the real data distribution \( p_{\text{data}}(x) \). Meanwhile, the discriminator \( D(x; \theta_d) \) outputs the probability that a given sample is real rather than generated. The objective is framed as a \textit{minimax optimization problem}, where \( G \) tries to generate realistic data to fool \( D \), and \( D \) aims to correctly classify real and fake data. The optimization problem can be expressed as: \begin{equation} \min_G \max_D V(D, G) = \mathbb{E}_{x \sim p_{\text{data}}(x)}[\log D(x)] + \mathbb{E}_{z \sim p_z(z)}[\log (1 - D(G(z)))] \end{equation} 
The goal for the generator is to improve until the discriminator cannot distinguish between real and fake data, ideally reaching a \textit{Nash equilibrium}, where \( D(x) = 0.5 \) for all \( x \). In practice, the original minimax objective can lead to vanishing gradients for the generator in early training. To address this, a more stable alternative called the \textit{non-saturating loss} is often used. Instead of minimizing \( \log(1 - D(G(z))) \), the generator maximizes \( \log D(G(z)) \), resulting in the following objective: \begin{equation} J_G = \mathbb{E}_{z \sim p_z(z)}[-\log D(G(z))] \end{equation} GANs are powerful because they model the data distribution implicitly without explicitly estimating the probability density. However, challenges such as \textit{training instability} and \textit{mode collapse} remain. 

\paragraph{Conditional GANs (cGANs) \cite{mirza2014conditionalgenerativeadversarialnets}} Conditional GANs extend traditional GANs by incorporating extra information, like class labels or attributes, to guide the generation process. In this setup, both the generator and discriminator receive this additional input to improve control over the generated outputs. The generator produces samples \( G(z|y) \) based on the input condition \( y \), while the discriminator checks whether the generated sample aligns with both the data distribution and the given condition. This design enables a wide range of tasks, such as text-to-image generation \cite{zhang2017stackgan}, face synthesis \cite{wav2pix2019icassp}, and image translation \cite{isola2017image}. The cGAN objective adjusts the standard GAN loss to ensure the outputs are not only realistic but also consistent with the provided conditions.

\paragraph{CycleGAN \cite{zhu2017unpaired}} CycleGAN is a framework designed for unpaired image-to-image translation, enabling transformation between two different domains (e.g., photos and paintings) without needing aligned image pairs for training. The key innovation of CycleGAN is the introduction of \textit{cycle consistency}, which ensures that if an image is converted from one domain to another and then back again, it closely matches the original input. The model uses two generators: \(G\), which maps images from domain \(X\) to domain \(Y\), and \(F\), which performs the reverse transformation from \(Y\) to \(X\). Additionally, two discriminators, \(D_X\) and \(D_Y\), are trained to distinguish between real and generated images in each domain, ensuring that the generated outputs are indistinguishable from real samples. The objective function consists of two components. First is the \textit{adversarial loss}, which encourages the generated images to appear realistic. For the generator \(G\), the adversarial loss is defined as: \begin{equation} \mathcal{L}_{GAN}(G, D_Y, X, Y) = \mathbb{E}_{y \sim p_{data}(y)}[\log D_Y(y)] + \mathbb{E}_{x \sim p_{data}(x)}[\log (1 - D_Y(G(x)))] \end{equation} This ensures that \(G(X)\) produces images that fool the discriminator \(D_Y\). Similarly, a parallel adversarial loss applies for the reverse generator \(F\) and discriminator \(D_X\). The second component is the \textit{cycle consistency loss}, which ensures that a round-trip translation returns the original input image. It is defined as: \begin{equation} \mathcal{L}_{cyc}(G, F) = \mathbb{E}_{x \sim p_{data}(x)}[||F(G(x)) - x||_1] + \mathbb{E}_{y \sim p_{data}(y)}[||G(F(y)) - y||_1] \end{equation} This loss penalizes discrepancies between the input image and the reconstructed image after forward and reverse transformations. The final objective function combines these losses: \begin{equation} \mathcal{L}(G, F, D_X, D_Y) = \mathcal{L}_{GAN}(G, D_Y, X, Y) + \mathcal{L}_{GAN}(F, D_X, Y, X) + \lambda \mathcal{L}_{cyc}(G, F) \end{equation} where \(\lambda\) is a hyperparameter that balances the importance of the cycle consistency loss relative to the adversarial loss. CycleGAN has been successfully applied to tasks such as style transfer, where it translates images between two visual styles (e.g., photos to paintings) without needing paired datasets. 

\paragraph{Autoencoders \cite{rumelhart1986learning}} Autoencoders are a special type of neural networks that try to reconstruct an input from the output. They consist of two main components: an encoder that compresses input data into a lower-dimensional representation and a decoder that reconstructs the original data from this compressed form. Indeed, the basic idea of their architecture is to extract the most important features of the input and then reconstruct the input from those features. Autoencoders are primarily used for unsupervised learning tasks, particularly in dimensionality reduction and data compression. This notion was introduced in the mid-1980s and has evolved significantly. \cite{HintonSalakhutdinov2006b} demonstrated their use for deep learning-based dimensionality reduction and subsequent innovations like denoising autoencoders \cite{vincent2008extracting}. Furthermore, they play a foundational role in generative AI, as they can model the underlying data structure by learning efficient representations in the latent space. This capability enables the generation of new data samples by decoding points from the latent space, forming the basis for techniques such as Variational Autoencoders (VAEs), which are widely used for tasks like image synthesis and generation.

More recently, \textit{supervised auto-encoders} have emerged, incorporating label information into the encoding process to improve feature learning. By aligning the learned representations with specific task objectives, these models have shown improved performance in classification and regression tasks compared to traditional auto-encoders \cite{zhao2015stacked}. This approach helps integrate both supervised and unsupervised learning for more robust representations.

\paragraph{Variational Autoencoders (VAEs) \cite{kingma2022autoencodingvariationalbayes}} VAEs are powerful extensions of traditional ones, particularly in generation tasks. The primary issue of autoencoders is that they focus solely on reducing the reconstruction error without ensuring that the latent space follows any structured or interpretable distribution. As a result, sampling directly from this latent space often leads to poor or unrealistic generations. VAEs address these issues by mapping the input to a probability distribution, typically a Gaussian, instead of mapping the input to a single point in the latent space. The key idea is to force the latent space to follow a predefined prior distribution. So, the loss function of VAE has one more term than the reconstruction error in autoencoders. The second term is the Kullback-Leibler (KL) divergence, which measures how much the learned distribution deviates from the prior. So, By adding this term and choosing a standard normal distribution as a prior distribution, we can ensure that the latent space is smooth and allows for sampling realistic data. The VAE optimizes the following objective called evidence lower bound (ELBO) \cite{jordan1999introduction}: 
\begin{equation} \mathcal{L} = \mathbb{E}_{q_{\phi}(z|x)} \left[ \log p_{\theta}(x|z) \right] - D_{\text{KL}} \left( q_{\phi}(z|x) \, || \, p(z) \right) \end{equation}
In the above $q_{\phi}(z|x)$ is the encoder network(approximate posterior), which maps the input $x$ to the latent variable, $p_{\theta}(x|z)$ is the decoder network(approximate likelihood), which reconstructs the input data $x$ from $z$ and $p(z)$ is the prior distribution over the latent variables.

\paragraph{Vector Quantized Variational Autoencoders (VQ-VAEs) \cite{oord2018neuraldiscreterepresentationlearning}} VQ-VAEs enhance the traditional VAE framework by using discrete latent variables instead of continuous ones. As the latent space is continuous in standard VAEs, it is challenging to model discrete data, such as categories, or tokens. VQ-VAEs address this by mapping the encoder output to a discrete codebook of learned embeddings. This approach allows the encoder to produce a latent vector quantized by selecting the closest codebook vector, creating a more structured and interpretable latent space. This makes VQ-VAEs particularly suitable for tasks such as image generation, speech synthesis, and natural language processing, where the data inherently has discrete structures.

Mathematically, given an input \( x \), the encoder produces a latent vector \( z_e(x) \), which is then mapped to the nearest vector \( e_k \) from a set of codebook vectors \( e_1, e_2, \dots, e_K \). The decoder reconstructs the original input from the quantized vector \( e_k \). The loss function of VQ-VAEs combines three terms: the reconstruction loss \( \mathcal{L}_{rec} \), the codebook loss \( \mathcal{L}_{vq} \) to minimize the distance between the encoder’s output and the closest codebook vector, and the commitment loss \( \mathcal{L}_{com} \) to ensure the encoder commits to the chosen codebook entry. The overall loss respectively is: \begin {equation}
\mathcal{L} =  \log p(x|z_q(x)) + || \text{sg}[z_e(x)] - e_k ||_2^2 + \beta || z_e(x) - \text{sg}[e_k] ||_2^2 \end{equation}
Here, sg represents the stop-gradient operation, preventing the codebook vectors from being updated by the gradients directly. This structure encourages efficient use of the codebook and ensures that the model produces high-quality reconstructions while maintaining a robust and interpretable latent space.

\paragraph{NeRF \cite{NERF}}
\begin{figure}
    \centering
    \includegraphics[width=0.5\linewidth]{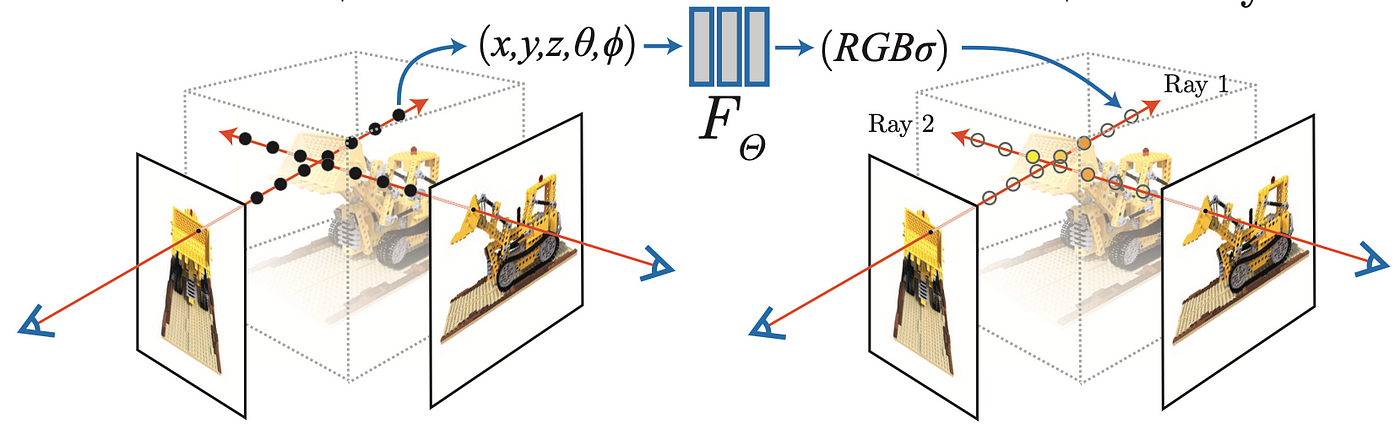}
    \caption{An overview of volumetric rendering in NeRF. Reprinted from \cite{NERF}.}
    \label{fig:nerf}
\end{figure}

Traditionally, there have been two broad categories of methods to represent 3D objects: explicit and implicit representations. Explicit methods such as voxel grids, point clouds, and structurally meshed surfaces model objects (i.e., you can see the structure through the stored data). However, these approaches are memory-intensive. In contrast, implicit representations offer a parametric description of 3D scenes, requiring significantly less memory. For a more comprehensive review of traditional methods, readers are encouraged to refer to \cite{3dsurvey}. 

NeRFs (Neural Radiance Fields) \cite{NERF} belong to the class of implicit representations but take a step further by using a neural network to learn the volumetric properties of a scene or an object. In other words, NeRF encodes the color and density at every point in 5D representations. In inference time, a frame of rays is cast to evaluate the color of each pixel in the final rendered image.   For each pixel, a ray is cast through the scene, and the color contribution along the ray is calculated as:
\begin{equation}C(\mathbf{r})=\int_{t_n}^{t_f} T(t) \sigma(\mathbf{r}(t)) \mathbf{c}(\mathbf{r}(t), \mathbf{d}) dt
\end{equation}
Where $T(t)=\exp \left(-\int_{t_n}^t \sigma(\mathbf{r}(s)) d s\right)$ is the accumulated transmittance along the ray, $\sigma(\mathbf{r}(t))$ represents the density function, which indicates the opacity at point $\mathbf{r}(t)$, and $\mathbf{c}(\mathbf{r}(t), \mathbf{d})$ gives the color of the point for a ray direction $\mathbf{d}$. The final pixel color, $C(r)$, is the accumulated contribution of all the densities and colors along the ray. Figure \ref{fig:nerf} provides an overview of this process.
A benefit of volumetric rendering is that it models the interaction between light and the medium through which it passes, which can make the product of renders more natural.
However, one key limitation of NeRF is that it requires retraining for each new scene, meaning that the trained weights of a network only capture the information for an individual object or environment.

\paragraph{NeRFs in Generative AI}
While NeRFs were initially developed for novel view synthesis of static scenes, they have been successfully adapted to various generative tasks in computer graphics and computer vision. For instance, in the domain of dynamic scene generation, Dynamic NeRFs \cite{dnerf} have been shown to perform competitively with traditional computer graphics techniques by treating time as an additional dimension in the neural representation, allowing for the synthesis of novel viewpoints of moving objects and scenes.

In the realm of generative AI, NeRF architecture has also been applied to tasks such as avatar creation, scene editing, and object manipulation. For example, in the generation of photorealistic human avatars, NeRF-based models can generate novel views of a person by learning a continuous volumetric representation of their appearance and geometry, thus allowing for the creation of highly detailed and consistent 3D human models.
One particularly exciting application of NeRFs is in the domain of content creation and virtual production. By applying volumetric rendering to learned scene representations, NeRFs can generate photorealistic views of objects or environments that can be seamlessly integrated into virtual productions. These models can capture complex lighting and material properties, making them particularly suited for tasks where physical accuracy and visual fidelity are essential, such as virtual film production or architectural visualization.

\paragraph{3D Gaussian Splatting \cite{3DGS}}
3D Gaussian Splatting (3DGS) is an alternative method for representing 3D scenes that builds on the advantages of NeRFs while addressing some of their limitations. It represents a 3D scene as a collection of Gaussian functions, where each Gaussian defines a local distribution of points in space. Instead of using a dense grid or neural representation to encode the scene, the scene is defined through a set of spatially distributed Gaussians that approximate the shape, color, and opacity of the object or scene.

Unlike NeRF \cite{NERF}, which requires a neural network to infer the color and density at every point along the ray, Gaussian Splatting takes advantage of more direct and computationally efficient rendering techniques by leveraging splats, small, volumetric, Gaussian-shaped primitives. These splats are rendered by blending their contributions based on their distance from the camera view. The splatting process computes the weighted contribution of each Gaussian along a ray-like volumetric rendering in NeRF but typically with fewer computational resources and memory requirements.

Each 3D Gaussian splat is parameterized by:
\textit{Mean position} (\( \mu_i \in \mathbb{R}^3 \)),
\textit{Covariance matrix} (\( \Sigma_i \in \mathbb{R}^{3 \times 3}) \) (defines the size, shape, and orientation of the Gaussian),
\textit{Color and opacity} (\( \mathcal{}{c}_i \in \mathbb{R}^3 \) and \( \alpha_i \in [0, 1] \))
The Gaussian density function for a point \( \mathbf{x} \) in 3D space is defined as:
\begin{equation}
G_i(\mathbf{x}) = \exp\left( -\frac{1}{2} (\mathbf{x} - \mu_i)^T \Sigma_i^{-1} (\mathbf{x} - \mu_i) \right)
\end{equation}
This equation gives the contribution of the \(i\)-th Gaussian splat to any point \( \mathbf{x} \) in space.
The rendering of a scene using 3D Gaussian Splatting requires blending the contributions of multiple Gaussians along the ray. For each ray \( \mathbf{r}(t) \), the color contribution at a specific point is:
\begin{equation}
C(\mathbf{r}) = \int_{t_n}^{t_f} \sum_{i} \alpha_i G_i(\mathbf{r}(t)) \, \mathbf{c}_i \, dt
\end{equation}
Here, \( G_i(\mathbf{r}(t)) \) is the Gaussian function evaluated along the ray at time \( t \), while \( \alpha_i \) controls the opacity of the Gaussian. The contributions from multiple Gaussians are blended along the ray, producing the final pixel color.
To render the splats correctly, a weighted accumulation of the Gaussians is needed, taking into account their opacity:
\begin{equation}
C(\mathbf{r}) = \sum_{i} \left( \alpha_i \prod_{j < i} (1 - \alpha_j) \right) \mathbf{c}_i\label{eq:splataccum}
\end{equation}
Equation~\ref{eq:splataccum} ensures how much opacity is contributed by subsequent Gaussians, which practically simulates depth and occlusion effects along the ray.


\paragraph{Denoising Diffusion Probabilistic Models (DDPMs) \cite{ho-denoising}} DDPMs represent a unique class of deep generative models that can generate high-quality and diverse outputs. They operate through two phases: a forward diffusion process, which progressively adds noise to data until it resembles white noise, and the reverse process called denoising. The essential idea is to systematically and slowly destroy the structure in a data distribution through an iterative forward diffusion process. It then learns a reverse diffusion process that restores structure and data, yielding a highly flexible and tractable generative model of the data. The forward process is modeled as a Markov process, adding Gaussian noise at each step using the following equation:

\begin{equation}
    q(x_t | x_{t-1}) = \mathcal{N}(x_t; \sqrt{1-\beta_t} x_{t-1}, \beta_t)
\end{equation}

Where $\beta_t$ controls the variance. The forward diffusion process uses a schedule (usually a Linear schedule) that scales the mean and the variance. At the end of this process, diffused data will have just a standard normal distribution. The reverse phase, or denoising, involves learning to reconstruct the original data from the noise using a neural network, essentially reversing the forward process. In this process, we can approximate $q(x_{t-1} |x_t )$ by trying to train a model to this denoising distribution. This model's structure explicitly uses properties of a Gaussian distribution, such as using the reparameterization trick in the case of sampling, and is defined by a noise schedule that controls the diffusion process.
DDPMs make use of a reverse process, starting from a base distribution, typically a standard normal distribution, and iteratively creating less noisy data through a parametric denoising model. This model is typically a normal distribution with parameters predicted by a neural network, making the true denoising distribution generally intractable. The training utilizes a similar variational upper bound as in VAE \cite{kingma2022autoencodingvariationalbayes} and minimizes the bound by minimizing the Kullback-Leibler divergence between the true and parametric distributions.
\begin{equation}
KL\left(q(x_{t-1} | x_t, x_0) \parallel p_{\theta}(x_{t-1} | x_t)\right) = \frac{1}{2\sigma_t^2} \|\mu_{\text{posterior}} - \mu_{\theta}(x_t)\|^2
\end{equation}
where $\mu_\theta$ is the output of the neural network for a given $x_t$, and $\sigma_t^2$ is the variance parameter. This approach enhances sample generation quality by leveraging the reparameterization trick and optimizing the noise prediction network.
The neural network architectures for such models generally include a U-Net \cite{DBLP:journals/corr/RonnebergerFB15} structure with self-attention and residual blocks conditioned on time via embeddings, including sinusoidal positional embeddings or random Fourier features. Training also explores hyperparameters such as the variance schedule ($\beta_t$) and noise variance ($\sigma_t^2$), with some approaches allowing these to be learned adaptively. This process underlines the importance of the diffusion parameters and network design for high-quality outputs in diffusion models.

This has been one of the pivotal approaches in generative AI, with applications ranging from high-quality image synthesis to text-to-image synthesis, video generation, and audio applications. The DDPM gives a more powerful capture of complex data distribution, hence a very solid basis for state-of-the-art generative systems in many different modalities. Their iterative nature and scalability further introduce them as a versatile tool for achieving high-fidelity outputs within practical AI applications.

\paragraph{ControlNet \cite{zhang2023adding}} ControlNet is an advanced model that enhances control within image diffusion processes by conditioning the generative model with auxiliary input images, such as Canny edges, sketches, human poses, and depth maps. This approach enables more precise guidance of the diffusion model, reducing the need for exhaustive prompt experimentation while increasing the specificity of generated images.

\begin{figure}[t]\centering
\includegraphics[height=5cm]{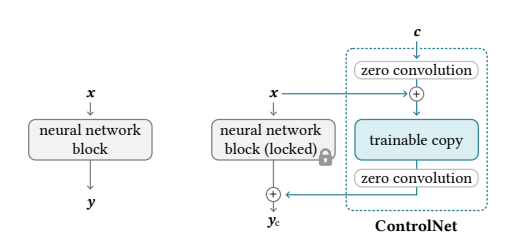}
\caption{Before ControlNet (left): A neural network block processes input \( x \) to output \( y \).  
After ControlNet (right): The original block is locked as \( y_c \), while a trainable copy integrates external input \( c \) via zero convolutions, enhancing functionality. Reprinted from \cite{zhang2023adding}.}
\label{fig:controlnet}
\end{figure}

ControlNet employs a dual-copy architecture for neural network blocks, incorporating both a "locked" copy that preserves the original model and a "trainable" copy that adapts to new conditions. This design allows for efficient fine-tuning with small image-pair datasets without compromising the integrity of production-ready models. A notable feature, "zero convolutions" (1x1 convolutions initialized with zero weights), ensures that new layers initially produce zero outputs, thereby preserving the original model's function. This mechanism allows ControlNet to integrate seamlessly with pre-trained models, providing an adaptable framework for training on small-scale or personal devices. The architecture also supports the merging, replacement, or offsetting of model components, offering artists and designers greater flexibility. Figure \ref{fig:controlnet} illustrates the difference that ControlNet has made to the architecture of neural network models for image generation.

Given an input image \( z_0 \), image diffusion algorithms iteratively add noise to produce a noisy image \( z_t \), where \( t \) represents the noise addition steps. With a set of conditions, including time step \( t \), text prompts \( c_t \), and a task-specific condition \( c_f \), these algorithms learn a network \( \epsilon_\theta \) to predict the noise added to \( z_t \), using the following objective:

\begin{equation}
L = \mathbb{E}_{z_0, t, c_t, c_f, \epsilon \sim \mathcal{N}(0,1)} \left[ \| \epsilon - \epsilon_\theta(z_t, t, c_t, c_f) \|_2^2 \right]
\end{equation}

Where \( L \) is the learning objective for the diffusion model, directly used to fine-tune models with ControlNet. During training, 50\% of text prompts \( c_t \) are replaced with empty strings to strengthen ControlNet’s capacity to recognize and generate outputs based solely on conditioning images (e.g., edges, poses, or depth maps), enhancing its ability to capture image-based semantics without text prompts.

\paragraph{ControlNet in Generative AI} ControlNet is an invaluable asset in generative AI, as it offers refined control over the image creation process. It allows users to specify intricate structural details, ensuring that the resulting visuals closely match the intended attributes. This capability promotes consistency, particularly in the depiction of complex figures like human poses, effectively reducing common distortions. Moreover, by accommodating a wide range of input types such as sketches and depth maps, ControlNet enables even basic outlines to inform and direct the production of high-quality images. Its robustness is further enhanced through training methods that intermittently remove prompts, shifting the focus to image-based conditions rather than relying solely on text. Overall, ControlNet significantly boosts the precision, versatility, and adaptability of generative AI applications, making it especially valuable for tasks that demand both creativity and structural accuracy.

\paragraph{DALL-E \cite{pmlr-v139-ramesh21a}} DALL-E is an AI model developed by OpenAI that generates images from textual descriptions. This model utilizes advanced neural networks and deep learning techniques to analyze the input text and associate it with visual concepts, creating new and creative images. 
DALL-E is a simple decoder-only transformer that receives both the text and the image as a single stream of 1280 tokens (256 for the text and 1024 for the image) and models all of them autoregressively. The attention mask at each of its 64 self-attention layers allows each image token to attend to all text tokens. DALL-E uses the standard causal mask for the text tokens and sparse attention for the image tokens with either a row, column, or convolutional attention pattern, depending on the layer.

Traditionally, text-to-image generation has focused on improving modeling assumptions for training on fixed datasets, which may involve complex architectures, auxiliary losses, or side information like object part labels or segmentation masks provided during training.
A simpler approach to this task is based on a transformer that autoregressively models text and image tokens as a single data stream. With sufficient data and scale, this approach proves competitive with domain-specific models when evaluated in a zero-shot fashion \cite{pmlr-v139-ramesh21a}.

The objective is to train a transformer to autoregressively model the text and image tokens as a unified data stream. However, directly using pixels as image tokens would require excessive memory for high-resolution images. Likelihood objectives tend to prioritize modeling short-range dependencies between pixels, meaning much of the model’s capacity would be used to capture high-frequency details instead of the low-frequency structures that make objects visually recognizable.
These challenges are addressed through a two-stage training procedure \cite{pmlr-v139-ramesh21a}:
(i) A discrete variational autoencoder (dVAE) is trained to compress each 256x256 RGB image into a 32x32 grid of image tokens, where each element can take on 8192 possible values. This reduces the context size for the transformer by a factor of 192 without significant degradation in visual quality. (ii) Up to 256 BPE-encoded text tokens are concatenated with the 32x32 (1024) image tokens and an autoregressive transformer is trained to model the joint distribution over the text and image tokens.
Then, model optimization techniques were used to achieve better results and optimal resource utilization. After the optimization is performed, the samples drawn from the transformer are reranked using a pre-trained contrastive model. For each caption and its candidate image, the contrastive model assigns a score based on how well the image aligns with the caption, and the top k images are selected \cite{pmlr-v139-ramesh21a}.

In the end, they were able to investigate a simple approach for text-to-image generation based on an autoregressive transformer when it is executed at scale. It is found that scale can lead to improved generalization, both in terms of zero-shot performance relative to previous domain-specific approaches and in terms of the range of capabilities that emerge from a single generative model. The findings suggest that improving generalization as a function of scale may be a useful driver for progress on this task \cite{pmlr-v139-ramesh21a}.

\subsection{Metrics}
Evaluating generative models requires carefully designed metrics to quantify different aspects of the generated content. These metrics serve as essential tools for assessing key attributes such as visual fidelity, realism, physical plausibility, and computational efficiency. Given the complexity and multi-faceted nature of generative outputs, evaluation measures are typically categorized based on distinct aspects of performance. A well-structured taxonomy of these metrics offers valuable insights into the capabilities and limitations of generative models, facilitating both research advancements and practical applications. 

\subsubsection{Quality and Realism of Generated Output}

This category encompasses evaluation metrics specifically designed to assess the perceptual quality, realism, and naturalness of generated content. Such evaluation often involves quantifying visual, auditory, or perceptual fidelity by comparing generated outputs to real-world data or human-annotated references. These metrics are particularly crucial in applications such as image synthesis, motion generation, and facial animation, where maintaining high perceptual quality and close resemblance to real-world counterparts is essential.

To evaluate the perceptual quality and realism of generated content, we present a set of metrics arranged from low-level pixel-based comparisons to high-level distributional and semantic assessments. We begin with basic pixel-wise metrics such as \textit{Mean Squared Error (MSE)} and \textit{Peak Signal-to-Noise Ratio (PSNR)}, which directly compare image intensities. We then include \textit{Structural Similarity Index (SSIM)}, which incorporates structural information and better aligns with human perception. Moving toward perceptual metrics, \textit{Learned Perceptual Image Patch Similarity (LPIPS)} compares deep feature representations to quantify visual similarity more holistically.

Beyond direct comparisons, we explore distributional similarity metrics such as \textit{Fréchet Inception Distance (FID)} and its modality-specific extensions, \textit{Fréchet Gesture Distance (FGD)}, \textit{Fréchet Video Distance (FVD)}, and \textit{Fréchet Audio Distance (FAD)}, which compare the statistics of deep embeddings extracted from real and generated content. Additionally, we include \textit{CLIP Score} and \textit{CLIP-MMD}, which measure alignment in a multimodal embedding space. Finally, we consider \textit{Identity Consistency}, which evaluates whether the generated output preserves identity-related features, particularly in tasks involving facial or gesture synthesis. Together, these metrics provide a layered and comprehensive view of how realistic, perceptually convincing, and semantically faithful the generated content appears across various modalities.

\paragraph{Mean Squared Error (MSE)} 
Mean Squared Error (MSE) is a widely used metric for measuring pixel-wise differences between a generated image and its corresponding real counterpart. It calculates the average of the squared differences between pixel values, providing an estimate of how closely the generated image approximates the ground truth. The MSE is computed as follows:
\begin{center}
\begin{equation}
\text{MSE} = \frac{1}{n}\sum_{i=1}^{n} (x_i - y_i)^2
\end{equation}
\end{center}
Here, \( n \) is the total number of pixels in the image, \( x_i \) represents the pixel value of the generated image at position \( i \), and \( y_i \) denotes the corresponding pixel value in the real image.

Although MSE does not originate from a single source, it has been extensively adopted in the evaluation of generative models across image synthesis, video-based motion reconstruction, and animation. For example, it has been employed to assess frame-level reconstruction in video-to-pose translation \cite{wang2018video2pose}, visual consistency in human animation pipelines \cite{chai2020everybodydance}, and pixel fidelity in transformer-based image synthesis \cite{esser2021taming}.

Lower MSE values indicate a higher degree of similarity between the generated and real images, suggesting better reconstruction quality. However, since MSE evaluates differences at the pixel level, it fails to capture perceptual differences or high-level structural variations, making it less suitable for assessing visual realism in generative models.

\paragraph{Peak Signal-to-Noise Ratio (PSNR) \cite{wang2004image}}  
The Peak Signal-to-Noise Ratio (PSNR) is a widely adopted metric for assessing the reconstruction quality of images or videos, especially after compression or generation. It quantifies the ratio between the maximum possible signal power and the power of the noise introduced by compression or generative artifacts, and is expressed in decibels (dB). The PSNR score is computed as follows:

\begin{center}
\begin{equation}
\mathrm{PSNR} = 10 \cdot \log_{10} \left( \frac{L^2}{\mathrm{MSE}} \right)
\end{equation}
\end{center}

Here, \( L \) represents the dynamic range of pixel values (e.g., 255 for 8-bit images), and \( \mathrm{MSE} \) is the Mean Squared Error between the original and generated image. Higher PSNR values indicate better reconstruction fidelity. However, PSNR primarily measures low-level pixel accuracy and may not align well with human perceptual judgments, especially in complex visual scenes. It should be complemented with other metrics that better capture perceptual quality.

\paragraph{Structural Similarity Index (SSIM) \cite{wangzhou2004image}}  
The Structural Similarity Index (SSIM) is a perceptual metric that evaluates the similarity between two images based on structural information, luminance, and contrast rather than raw pixel-wise differences. SSIM is designed to be more consistent with human visual perception, making it especially useful in generative tasks involving visual realism. The SSIM score is computed as follows:

\begin{center}
\begin{equation}
\mathrm{PSNR} = 10 \cdot \log_{10} \left( \frac{L^2}{\mathrm{MSE}} \right)
\end{equation}
\end{center}

In this equation, \( \mu_x \) and \( \mu_y \) denote the means of the images \( x \) and \( y \); \( \sigma_x^2 \), \( \sigma_y^2 \) represent their variances; and \( \sigma_{xy} \) is the covariance between them. Constants \( C_1 \) and \( C_2 \) stabilize the division when denominators are small. SSIM ranges from -1 to 1, where a score closer to 1 indicates stronger structural similarity. It is especially valuable in applications such as super-resolution, inpainting, and generative image restoration, where perceptual quality matters more than raw pixel accuracy.

\paragraph{Learned Perceptual Image Patch Similarity (LPIPS) \cite{zhang2018unreasonableeffectivenessdeepfeatures}} 
The Learned Perceptual Image Patch Similarity (LPIPS) metric evaluates the perceptual similarity between two images by comparing deep feature representations extracted from a pretrained neural network. Unlike pixel-wise metrics such as MSE, LPIPS is more closely aligned with human visual perception, making it a valuable tool for assessing image quality in generative tasks. The LPIPS score is computed as follows:
\begin{center}
\begin{equation}
\text{LPIPS}(x, y) = \sum_l \frac{1}{H_l W_l} \sum_{h=1}^{H_l} \sum_{w=1}^{W_l} \left\| \hat{w}_l \odot \left( \hat{\phi}_l(x)_{h,w} - \hat{\phi}_l(y)_{h,w} \right) \right\|_2^2
\end{equation}
\end{center}
Here, \( \hat{\phi}_l(x) \) and \( \hat{\phi}_l(y) \) denote the unit-normalized deep feature embeddings extracted from layer \( l \) of a pretrained network for images \( x \) and \( y \), respectively. The terms \( H_l \) and \( W_l \) represent the height and width of the spatial dimensions of the feature maps at layer \( l \). The vector \( \hat{w}_l \) contains learned per-channel weights that modulate the contribution of each channel to the final perceptual distance. The symbol \( \odot \) denotes element-wise multiplication, and \( \| \cdot \|_2^2 \) is the squared Euclidean norm. Lower LPIPS scores indicate higher perceptual similarity between the images, making this metric well-suited for tasks like image generation, super-resolution, and style transfer, where human visual perception is a more appropriate criterion than pixel-level fidelity.

\paragraph{Fréchet Inception Distance (FID) \cite{heusel2018ganstrainedtimescaleupdate}} 
The Fréchet Inception Distance (FID) quantifies the similarity between the feature distributions of real and generated samples. It measures the statistical distance between deep feature representations extracted using a pretrained Inception network. Lower FID values indicate that the generated samples are perceptually closer to real data, both in structure and content. The FID score is computed as:

\begin{center}
\begin{equation}
\text{FID} = \left\| \mu_r - \mu_g \right\|^2 + \text{Tr} \left( \Sigma_r + \Sigma_g - 2\sqrt{\Sigma_r \Sigma_g} \right)
\end{equation}
\end{center}

Here, \( \mu_r \) and \( \mu_g \) denote the means, and \( \Sigma_r \), \( \Sigma_g \) the covariances of real and generated feature distributions, respectively. This metric is widely used in image synthesis and character animation to assess the quality and realism of generated outputs.

Several adaptations of FID have been proposed to better align with domain-specific evaluation needs across different modalities. The Fréchet Gesture Distance (FGD) \cite{Zabala_2021} modifies the original formulation by using motion-specific feature extractors instead of visual encoders, allowing it to evaluate the fidelity of generated gestures in co-speech animation, avatar behavior, and embodied agents. Similarly, the Fréchet Video Distance (FVD) \cite{unterthiner2019accurategenerativemodelsvideo} extends FID to video generation by incorporating temporal dynamics. It utilizes spatiotemporal networks such as I3D to extract features that account for both spatial and temporal coherence, making it suitable for evaluating video synthesis, character motion, and human activity generation. On the audio side, the Fréchet Audio Distance (FAD) \cite{kilgour2019frechetaudiodistancemetric} applies the same statistical formulation to embeddings extracted from real and generated audio using pretrained models like VGGish \cite{10.1109/ICASSP.2017.7952132}. FAD has been used in speech generation, music enhancement, and audio-based animation tasks to measure how perceptually similar the generated audio is to real-world recordings. While all these variants share the same core mathematical structure as FID, their effectiveness depends on the modality-specific embeddings used, enabling them to capture distributional fidelity in gestures, videos, or audio, respectively.

\paragraph{CLIP-MMD \cite{jayasumana2024rethinkingfidbetterevaluation}} 
CLIP-MMD (CLIP-based Maximum Mean Discrepancy) or CMMD is a perceptual metric that evaluates the similarity between the distributions of real and generated samples in a multimodal embedding space. Unlike traditional metrics such as FID that rely on second-order statistics and Gaussian assumptions, CLIP-MMD employs Maximum Mean Discrepancy (MMD) in the CLIP feature space, making it more suitable for assessing perceptual and semantic fidelity, particularly in domains involving text-to-image or image-to-image generation. The CLIP-MMD score is computed as follows:
\begin{center}
\begin{equation}
\text{MMD}^2(X, Y) = \mathbb{E}_{x,x' \sim X} [k(x, x')] + \mathbb{E}_{y,y' \sim Y} [k(y, y')] - 2\mathbb{E}_{x \sim X, y \sim Y} [k(x, y)]
\end{equation}
\end{center}

Here, \( X \) and \( Y \) are the sets of CLIP feature embeddings extracted from real and generated samples, respectively, and \( k(\cdot, \cdot) \) is a kernel function, typically a Gaussian or polynomial kernel, used to compute similarity in the embedding space.

CLIP-MMD offers a non-parametric and distribution-free alternative to FID by capturing higher-order discrepancies without relying on moment-matching assumptions. It has shown improved correlation with human judgment in evaluating semantic alignment and perceptual quality across tasks such as text-to-image synthesis, visual style transfer, and multimodal generative modeling.

\paragraph{CLIP Score \cite{hessel2022clipscorereferencefreeevaluationmetric}} 
The CLIP Score is a metric designed to evaluate the semantic similarity between generated visual content and its corresponding textual description. It leverages the CLIP model \cite{DBLP:conf/icml/RadfordKHRGASAM21}, which maps both images and text into a shared embedding space. The metric quantifies the alignment between a generated image and its textual description by calculating the cosine similarity between their respective feature embeddings. The CLIP Score is defined as:
\begin{equation}
\text{CLIPScore}(t, i) = 2.5 \cdot \max\left( \frac{t \cdot i}{\|t\| \, \|i\|}, \, 0 \right)
\end{equation}
In this formula, \( \mathbf{t} \) represents the text embedding produced by CLIP for a given textual prompt, while \( \mathbf{i} \) denotes the image embedding generated by CLIP for the corresponding image. The cosine similarity between these embeddings is computed by taking the dot product \( t \cdot i \), which measures the degree of alignment between the text and image features in the shared semantic space. The norms \( \|t\| \) and \( \|i\| \) are the Euclidean magnitudes of the text and image embeddings, respectively, and they serve to normalize the similarity score. This normalization ensures that the similarity is independent of the absolute lengths of the embeddings, making it a scale-invariant measure. The \(\max\) function is used to guarantee that the score is never negative, as a negative value would indicate a misalignment between the embeddings. Finally, the constant factor of \(2.5\) is introduced to scale the result to a suitable range, ensuring the CLIP Score provides meaningful and comparable values. Higher CLIP Scores indicate a stronger semantic alignment between the generated image and the textual description, making this metric particularly valuable for evaluating text-to-image generation models and other multimodal AI applications.

\paragraph{Identity Consistency \cite{dong2022protectingcelebritiesdeepfakeidentity}} 
The Identity Consistency metric measures the degree to which the inner and outer regions of a face image represent the same identity. This concept is particularly important in deepfake detection, where identity inconsistencies between manipulated facial regions can indicate tampering. \cite{dong2022protectingcelebritiesdeepfakeidentity} introduced the Identity Consistency Transformer (ICT), which extracts separate identity vectors for the inner and outer facial regions and compares them to assess coherence. The Identity Consistency score is defined as:

\begin{equation}
\text{Identity Consistency}(I) = d(f_{\text{in}}, f_{\text{out}})
\end{equation}

Here, \( I \) denotes the input face image, and \( f_{\text{in}} \) and \( f_{\text{out}} \) represent identity embeddings extracted from the inner and outer facial regions using ICT. The function \( d(\cdot, \cdot) \) is a distance metric such as cosine or Euclidean distance. Lower scores indicate higher identity consistency and suggest the face is likely authentic, while higher scores may reveal manipulated or inconsistent facial features. This metric is especially useful in detecting face swaps in deepfake content and protecting the identity integrity of public figures.

These metrics serve as key indicators of how realistic and visually convincing the generated content appears. They quantify the alignment between real and generated data distributions, ensuring that synthesized outputs maintain high perceptual fidelity. This category of metrics is particularly relevant in applications such as Generative Adversarial Networks (GANs) \cite{goodfellow2014generativeadversarialnetworks} for image synthesis and gesture generation models, where the primary objective is to produce outputs that are both visually plausible and semantically consistent with real-world data.
\\

\subsubsection{Diversity and Multimodality}
This category evaluates a model’s ability to generate diverse and varied outputs across multiple modalities, including text, images, and gestures. Metrics in this domain assess whether the model effectively mitigates mode collapse, a phenomenon where the model produces overly deterministic outputs with limited variation. Instead, these metrics ensure that the generative process retains flexibility, allowing for a rich and diverse set of outputs that align with the natural variability observed in real-world data.

Diversity and multimodality are particularly crucial in applications where creative content generation, user personalization, and multimodal synthesis play a fundamental role. Ensuring high diversity not only enhances the expressiveness and adaptability of generative models but also fosters better engagement and realism in AI-driven content creation. To evaluate diversity in generated outputs, we consider three complementary metrics that capture different aspects of variation. We begin with the \textit{Diversity} metric, which measures global variation across independent samples. Next, we examine the \textit{Multimodality} metric, which focuses on within-class diversity under specific conditioning labels. Finally, we consider the \textit{Average Pairwise Distance (APD)}, which provides a more general statistical perspective by averaging distances between all sample pairs. Together, these metrics offer a comprehensive view of the model's ability to produce varied, non-redundant outputs across multiple modalities.


\paragraph{Diversity \cite{Textguided3H}} 
The Diversity metric quantifies the variation between two independently sampled subsets of the generated outputs. It ensures that the model avoids producing repetitive or overly similar outputs. The Diversity score is computed as follows:
\begin{center}
\begin{equation}
\text{Diversity} = \frac{1}{N} \sum_{i=1}^{N} \left\| x_i - x'_i \right\|_2
\end{equation}
\end{center}
Here, \( \{x_1, ..., x_N\} \) and \( \{x'_1, ..., x'_N\} \) are two independently sampled sets from the model’s output distribution, and \( \| \cdot \|_2 \) denotes the Euclidean distance between corresponding samples. Higher Diversity values indicate a broader range of variation in the generated outputs, which is essential for producing diverse and non-redundant results.

\paragraph{Multimodality \cite{Textguided3H}} 
The Multimodality metric evaluates the model’s ability to generate diverse outputs within a specific action (label) class. Unlike the Diversity metric, which measures global variation, Multimodality focuses on within-class variation under a fixed conditioning label. The score is computed as follows:
\begin{center}
\begin{equation}
\text{Multimodality} = \frac{1}{C \cdot N} \sum_{c=1}^{C} \sum_{n=1}^{N} \left\| x_{c,n} - x'_{c,n} \right\|_2
\end{equation}
\end{center}
Here, \( \{ x_{c,1}, \dots, x_{c,N} \} \) and \( \{ x'_{c,1}, \dots, x'_{c,N} \} \) denote two independently sampled sets of outputs conditioned on action class \( c \), where \( C \) is the total number of classes and \( N \) is the number of samples per class. The term \( \| \cdot \|_2 \) represents the Euclidean distance between corresponding samples. Higher Multimodality scores indicate the model’s capacity to generate rich and diverse outputs within each class, promoting mode variety and preventing collapse.

\paragraph{Average Pairwise Distance (APD) \cite{mix-and-match-perturbation}} 
The Average Pairwise Distance (APD) metric quantifies the average distance between all generated samples, helping to ensure that the model produces a diverse set of outputs. It is particularly useful in multimodal generation tasks, where maintaining variability across outputs is critical. It is used in various related generative works such as Pose-NDF \cite{tiwari22posendf}. The APD score is calculated as follows:
\begin{center}
\begin{equation}
\text{APD} = \frac{1}{N(N-1)} \sum_{i \neq j} \left\| x_i - x_j \right\|_2
\end{equation}
\end{center}
Here, \( N \) denotes the total number of generated samples, while \( x_i \) and \( x_j \) represent individual generated instances. The term \( \| x_i - x_j \|_2 \) measures the Euclidean distance between the two samples. Higher APD values indicate greater diversity among the generated outputs, reducing the likelihood of mode collapse and encouraging the model to explore a broader range of possibilities.

\subsubsection{Relevance and Accuracy}
This category of metrics evaluates how accurately generated content aligns with ground truth data or expected task-specific outcomes. These measures are particularly relevant in applications requiring high precision, such as motion synthesis, speech-aligned gesture generation, and keypoint-based animation, ensuring that generated outputs are semantically and structurally accurate.
By assessing the degree of alignment between generated and real data, these metrics help determine whether the model faithfully replicates desired patterns, reducing errors and enhancing reliability in generative tasks. 

To assess the relevance and accuracy of generated outputs, we present metrics that span spatial, emotional, temporal, and semantic dimensions. We begin with positional metrics such as \textit{Mean Absolute Joint Error (MAJE)} and \textit{Lip Vertex Error (LVE)}, which evaluate the spatial precision of joints and facial features. \textit{Emotional Vertex Error (EVE)} follows, measuring the accuracy of expression-related regions. We then consider temporal synchrony metrics, including \textit{Beat Consistency (BC)} and \textit{Audio-Visual Synchrony Score}, which assess alignment between motion and speech. Finally, we include \textit{Multimodal Distance (MM-Distance)}, which quantifies semantic alignment between different modalities. This progression allows for a layered evaluation of how closely the generated content matches real-world data both structurally and contextually.


\paragraph{Mean Absolute Joint Error (MAJE)} 
The Mean Absolute Joint Error (MAJE) metric quantifies the discrepancy between predicted and ground truth joint positions in motion synthesis tasks. Unlike squared error metrics, MAJE computes the absolute differences, making it less sensitive to large outliers while providing a more intuitive measure of positional accuracy. The MAJE score is calculated as follows:
\begin{center}
\begin{equation}
\text{MAJE} = \frac{1}{n} \sum_{i=1}^{n} |x_i - y_i|
\end{equation}
\end{center}
Here, \( n \) denotes the total number of evaluated joints, \( x_i \) represents the predicted joint position at index \( i \), and \( y_i \) corresponds to the ground truth joint position. The absolute difference \( |x_i - y_i| \) captures the positional error between predicted and actual joint positions, enabling accurate evaluation of the model's performance in motion synthesis and pose estimation.

Although MAJE does not originate from a single foundational paper, it has been widely adopted in gesture generation and pose estimation literature as a core evaluation metric for spatial accuracy. Notably, prior works have employed MAJE to assess the fidelity of synthesized joint trajectories in tasks such as gesture-from-audio generation \cite{li2021audio2gestures,ginosar2019learning}, and expressive motion modeling from multimodal cues \cite{ferstl2021expressive}. 
Lower MAJE values indicate greater positional accuracy, making this metric particularly important for tasks such as motion synthesis, human pose estimation, and gesture generation.


\paragraph{Lip Vertex Error \cite{richard2021meshtalk} and Emotional Vertex Error \cite{Peng_2023_ICCV}}  
These are region-specific perceptual metrics designed to evaluate spatial accuracy in speech-driven and expressive facial animation. Both metrics compute the maximum per-frame vertex-wise error across time, targeting different regions of the face based on application-specific needs. Despite sharing a unified mathematical formulation, their interpretative focus and targeted facial regions differ. The common formulation is given by:

\begin{center}
\begin{equation}
\mathrm{VE} = \frac{1}{T} \sum_{t=1}^{T} \max_{v \in V_R} \left\| \hat{y}_{t,v} - y_{t,v} \right\|_2
\end{equation}
\end{center}

In this formulation, \( T \) denotes the number of frames in the animation sequence, and \( v \in V_R \) refers to the subset of vertices within a specified region of interest \( V_R \). \( \hat{y}_{t,v} \) and \( y_{t,v} \) represent the predicted and ground truth 3D coordinates of vertex \( v \) at time step \( t \), respectively. While this equation remains constant across both metrics, the semantics of \( V_R \) and the source of the vertex data vary. Specifically, Lip Vertex Error (LVE) is computed over the lip region using mesh data generated from audio input, typically driven by a neutral template mesh as in MeshTalk \cite{richard2021meshtalk}. In contrast, Emotional Vertex Error (EVE) targets emotion-relevant areas such as the brows, eyes, and forehead, and is computed over expression-driven blendshape coefficients mapped to 3D vertices using the FLAME model \cite{Peng_2023_ICCV}.

Thus, while LVE quantifies the accuracy of phoneme-synchronized lip motion in speech-driven animation, EVE measures the emotional fidelity and expressiveness captured in the upper face. These metrics complement each other: LVE focuses on lip-sync precision, which is crucial for intelligibility, while EVE emphasizes emotional realism, which is key for expressive and affective character animation.

\paragraph{Beat Consistency (BC) \cite{guichoux20242d2ddoesdimensionality}} 
The Beat Consistency (BC) metric quantifies the degree of temporal alignment between motion and speech rhythms based on beat timing. It is particularly useful in speech-driven gesture generation, dance synthesis, and multimodal AI systems, where synchrony across modalities enhances realism and coherence. The BC score is calculated as follows:
\begin{center}
\begin{equation}
\text{BC} = \frac{1}{n} \sum_{i=1}^{n} \exp\left( -\frac{ \min_{t_j \in B_y} \| t_{x_i} - t_j \|^2 }{2\sigma^2} \right)
\end{equation}
\end{center}
Here, \( n \) denotes the number of detected audio beats \( t_{x_i} \), and \( B_y \) is the set of detected motion beat timestamps. For each audio beat, the squared temporal distance to the nearest motion beat is computed and passed through an exponential decay function, where \( \sigma \) is a smoothing factor (typically set to 0.1). Higher BC values indicate stronger alignment between motion and speech timing, making this metric essential for evaluating gesture animation, speech-gesture synchrony, and expressive avatar generation.

\paragraph{Multimodal Distance \cite{guo2022tm2t}} 
The Multimodal Distance (MM-Distance) metric measures the degree of alignment between generated motion and its corresponding textual description. It evaluates how closely the feature representations of two different modalities, such as text and motion, correspond to each other, ensuring that the generated motion remains semantically relevant to the input description. The MM-Distance score is calculated as follows:
\begin{center}
\begin{equation}
\text{MM-Distance} = \sqrt{\frac{1}{N} \sum_{n=1}^{N} \left\| f_{a,n} - f_{b,n} \right\|_2^2}
\end{equation}
\end{center}
Here, \( N \) represents the total number of evaluated samples, \( f_{a,n} \) denotes the feature embedding of the generated motion for sample \( n \), and \( f_{b,n} \) corresponds to the feature embedding of the associated textual input. The term \( \| \cdot \|_2^2 \) denotes the squared Euclidean (L2) norm.

The set of paired feature representations is defined as:
\[
\{(f_{a,1}, f_{b,1}), \dots, (f_{a,N}, f_{b,N})\}
\]
where each pair \( (f_{a,n}, f_{b,n}) \) consists of feature vectors extracted from two different modalities, \( a \) (e.g., motion) and \( b \) (e.g., text). While MM-Distance can be used to assess alignment between any pair of modalities, it is most commonly applied in text-to-motion generation tasks to evaluate how well the synthesized motion captures the semantics of the input description, ensuring higher fidelity in motion synthesis.

The above metrics assess how accurately the generated content aligns with ground truth data or expected task-specific outcomes, providing a quantitative measure of correctness. They are particularly crucial in applications where semantic accuracy holds greater importance than purely visual fidelity. In tasks such as human motion synthesis and gesture generation, ensuring that the generated movements are contextually and semantically appropriate is essential. Relevance and accuracy metrics help evaluate whether the model faithfully replicates meaningful gestures and motion patterns rather than just producing visually plausible outputs.

\subsubsection{Physical Plausibility and Interaction}
This category evaluates the physical realism of generated content, particularly in scenarios involving motion, object interaction, and environmental constraints. These metrics ensure that synthesized outputs adhere to real-world physical principles, such as smooth motion trajectories, non-colliding movements, and biomechanically plausible gestures. In applications like human motion synthesis, robotics, and animation, maintaining physical plausibility is crucial to prevent unnatural or implausible results.

Although these metrics are less frequently used compared to visual or perceptual measures, they are essential for applications that require physically grounded outputs. Evaluating physical plausibility helps ensure that the generated content is not only visually appealing but also mechanically viable and functionally realistic.
We present the key metrics in this category along with their respective formulations.
To assess the physical plausibility of generated motion, we introduce metrics that capture both overall dynamic consistency and specific physical interactions. We begin with the \textit{Mean Acceleration Difference (MAD)}, which evaluates how closely the model’s motion adheres to realistic acceleration patterns, reflecting the smoothness and naturalness of the generated sequences. We then examine the \textit{Foot Skating (FS)} metric, which focuses on the physical correctness of foot-ground contact by detecting unnatural sliding artifacts. Together, these metrics provide a comprehensive view of physical realism, from internal motion dynamics to external environmental interactions.

\paragraph{Mean Acceleration Difference (MAD) \cite{bhattacharya2021speech2affectivegestures}} 
The Mean Acceleration Difference (MAD) metric quantifies the discrepancy between actual and predicted acceleration in generated motion. It is essential for ensuring that synthetic motion adheres to realistic dynamics and avoids abrupt or unnatural changes in movement. The MAD score is calculated as follows:
\begin{center}
\begin{equation}
\text{MAD} = \frac{1}{n} \sum_{i=1}^{n} \left\| a_i - \hat{a}_i \right\|_2
\end{equation}
\end{center}
Here, \( n \) denotes the total number of evaluated time steps or motion frames, \( a_i \) represents the actual acceleration at time step \( i \), and \( \hat{a}_i \) is the predicted acceleration. The Euclidean norm \( \| \cdot \|_2 \) measures the magnitude of the difference between actual and predicted values.

Lower MAD scores indicate that the generated motion more closely follows expected acceleration patterns, resulting in smoother and more natural transitions. This metric is particularly relevant in human motion synthesis, character animation, and physics-based simulations, where adherence to realistic kinematic behavior is critical for visual and physical plausibility.

\paragraph{Foot Skating (FS)~\cite{mourot2022underpressuredeeplearningfoot}}  
The Foot Skating (FS) metric detects unnatural foot movements in human pose generation, particularly in motion synthesis tasks where physically plausible foot contact is critical. This metric quantifies foot instability by measuring the \textbf{horizontal velocity of the feet during ground contact phases}, where the velocity is theoretically expected to approach zero. The FS score is computed as:
\begin{equation}
\text{FS} = \frac{1}{|C|} \sum_{t \in C} \left\| \text{horizontal\_foot\_velocity}(t) \right\|
\end{equation}
Here, \( C \) denotes the set of time steps during which the foot is in contact with the ground (as determined by foot contact labels), and \( \text{horizontal\_foot\_velocity}(t) \) represents the horizontal component (typically in the XY-plane) of the foot’s velocity at time \( t \). The Euclidean norm \( \| \cdot \| \) is used to measure the magnitude of velocity.

Lower FS values indicate more physically plausible foot contact, minimizing artifacts such as sliding or footskate.

By ensuring adherence to real-world physical principles, these metrics help evaluate whether generated movements exhibit characteristics such as smooth transitions, biomechanical plausibility, and non-colliding interactions. Such evaluations are critical in fields like robotics, animation, and virtual avatar creation, where maintaining physical plausibility is essential for producing realistic and functionally appropriate motion.

\subsubsection{Efficiency and Computational Metrics}
This category focuses on evaluating the computational efficiency of generative models, ensuring that they balance output quality with resource consumption. Metrics in this cluster measure factors such as inference speed, memory usage, and scalability, which are critical for real-time applications and large-scale deployments.
While these metrics are often underreported in generative modeling literature, they play a pivotal role in determining the feasibility of deploying models in time-sensitive and resource-constrained environments. In scenarios such as interactive AI systems, gaming, robotics, and animation pipelines, computational efficiency directly impacts usability.

To evaluate the computational efficiency of generative models, we consider two complementary metrics that capture both processing speed and output quality under resource constraints. We begin with \textit{Execution Time}, a widely used yet informally standardized metric that directly measures the duration required to generate outputs. Although \textit{Execution Time} lacks a single point of origin, it has been employed in various generative works, particularly in animation and character modeling contexts, to assess inference latency and runtime efficiency \cite{alex2020characteranimation, bhattacharya2021speech2affectivegestures, yao2023stylegesture}. We then introduce \textit{Kernel Inception Distance (KID)}, a metric designed to evaluate the quality of generated outputs while ensuring computational efficiency. This approach is especially advantageous in situations with limited data or constrained resources. Together, these metrics provide a balanced perspective on the speed, performance, and deployability of generative systems.


\paragraph{Execution Time} 
The Execution Time metric measures the duration required to generate outputs. It is calculated as:
\begin{center}
\begin{equation}
\text{Execution Time} = \text{End Time} - \text{Start Time}
\end{equation}
\end{center}
Here, Start Time refers to the timestamp when the model begins processing an input, and End Time denotes the moment the output is fully generated. The difference reflects the total time taken for inference or synthesis.
Lower execution times indicate greater computational efficiency, making the model more suitable for real-time applications such as interactive AI, gaming, and robotics. Conversely, higher execution times may reveal performance bottlenecks that require optimization, especially in large-scale or latency-sensitive scenarios.
Execution Time has been widely adopted in various generative modeling contexts. For example, prior works have used execution time to assess the runtime performance of character animation systems \cite{alex2020characteranimation}, speech-driven gesture generation \cite{bhattacharya2021speech2affectivegestures}, and style-aware motion synthesis \cite{yao2023stylegesture}.

\paragraph{Kernel Inception Distance (KID) \cite{bińkowski2021demystifyingmmdgans}} 
The Kernel Inception Distance (KID) metric evaluates the quality of generated outputs while maintaining computational efficiency. It is proposed as an alternative to the Fréchet Inception Distance (FID), offering an unbiased estimator that is particularly reliable when working with small sample sizes. The KID score is computed as follows:
\begin{center}
\begin{equation}
\text{KID} = \frac{1}{n(n-1)} \sum_{i \neq j} k(\phi(x_i), \phi(x_j)) + \frac{1}{m(m-1)} \sum_{i \neq j} k(\phi(y_i), \phi(y_j)) - \frac{2}{nm} \sum_{i,j} k(\phi(x_i), \phi(y_j))
\end{equation}
\end{center}
Here, \( x_i \) and \( y_j \) are feature representations of real and generated samples, respectively, and \( \phi(\cdot) \) denotes the feature embedding extracted using an Inception model. The function \( k(\cdot, \cdot) \) represents a polynomial kernel used to compute similarities in the feature space. The values \( n \) and \( m \) correspond to the number of real and generated samples.

Lower KID values indicate that the distribution of generated samples is closer to that of the real samples, reflecting higher-quality outputs. Unlike FID, KID does not assume Gaussianity in feature distributions, making it more robust in scenarios involving small datasets or non-normal data distributions. KID is widely adopted in the evaluation of generative adversarial networks (GANs), image synthesis, and benchmarking generative models, where both quality and computational efficiency are important.


\subsubsection{Deep Learning-Based Synchronization and Perceptual Metrics}

This category introduces evaluation metrics that leverage deep learning models to evaluate temporal synchrony and perceptual alignment between modalities, particularly audio and visual streams. Unlike conventional metrics that rely on geometric distances, pixel-wise errors, or handcrafted rules, these approaches learn evaluative behavior directly from data, using either self-supervised contrastive learning or supervision from human ratings.

Such metrics are particularly relevant in scenarios where human perception plays a central role, like speech-driven animation, video dubbing, and avatar-based communication, as they better align with how users experience multimodal coherence. By relying on pretrained networks and perceptually grounded feature representations, they offer task-specific, scalable alternatives to handcrafted synchrony measures.
We introduce two representative examples in this category: \textit{SyncNet-based Metrics}, which assess audio-visual alignment using learned contrastive embeddings, and the \textit{PEAVS Score}, a reference-free model trained on large-scale human opinion data to predict perceived synchrony quality.

\paragraph{SyncNet-based Metrics \cite{chung2017out}}  
Many evaluation metrics leverage the SyncNet model to assess temporal alignment between audio and visual modalities in speech-driven facial animation. SyncNet is a two-stream convolutional network trained to detect whether a video frame and an audio segment are synchronized.

The core idea behind SyncNet-based metrics is to compute temporal distances between audio and video embeddings using a sliding window approach. While the original SyncNet method does not explicitly define a scalar synchrony score, later works operationalize it by aggregating distances over time \cite{bozkurt2023personalizedspeechdrivenexpressive3d, prajwal2020lip}. One widely used metric derived from SyncNet is the Lip Sync Error - Distance (LSE-D) \cite{prajwal2020lip}, which is defined as:

\begin{center}
\begin{equation}
\text{LSE-D} = \frac{1}{T} \sum_{t=1}^{T} \left\| a(t) - v(t) \right\|_2
\end{equation}
\end{center}

Here, \( a(t) \) and \( v(t) \) denote feature embeddings extracted from audio and visual streams at time step \( t \), and \( T \) is the total number of analyzed frames. Euclidean distance is used to measure alignment. Lower LSE-D indicates better lip-sync and temporal coherence, which is essential for realistic speech animation, dubbing, and avatar interaction.

\paragraph{PEAVS Score \cite{goncalves2024peavs}}  
The Perceptual Evaluation of Audio-Visual Synchrony (PEAVS) is a reference-free, model-based metric designed to predict human-perceived synchrony between audio and visual streams. Unlike metrics relying on geometric alignment, PEAVS incorporates a deep neural architecture trained on 120,000+ human annotations spanning nine synchronization error types (e.g., temporal shifts, intermittent muting).

The PEAVS model combines motion (I3D \cite{carreira2017quo}) and audio (VGGish \cite{10.1109/ICASSP.2017.7952132}) features, processes them through a Transformer-based fusion network with cross-modal attention, and outputs a discrete interpretable score between 1 and 5 (1 = severe asynchrony, 5 = perfect synchronization). To align predictions with human ratings, the model is trained using the Concordance Correlation Coefficient (CCC), which jointly captures precision and accuracy of the predictions:

\begin{equation}
\mathcal{L}_{\text{PEAVS}} = \frac{2 \rho \sigma_x \sigma_y}{\sigma_x^2 + \sigma_y^2 + (\mu_x - \mu_y)^2}
\end{equation}

Here, \( \rho \) denotes the Pearson correlation between predicted (\( \hat{s}_i \)) and ground-truth (\( s_i \)) scores across a batch; \( \mu_x, \mu_y \) are the respective means, and \( \sigma_x, \sigma_y \) the standard deviations. PEAVS thus produces a continuous synchrony score that aligns closely with perceptual expectations, making it highly suitable for evaluating dubbing, video synthesis, and expressive speech-driven avatars.

\end{document}